\documentclass{article}

     \usepackage[nonatbib,preprint]{neurips_2022}

\usepackage[utf8]{inputenc} % allow utf-8 input
\usepackage[T1]{fontenc}    % use 8-bit T1 fonts
\usepackage{hyperref}       % hyperlinks
\usepackage{url}            % simple URL typesetting
\usepackage{booktabs}       % professional-quality tables
\usepackage{amsfonts}       % blackboard math symbols
\usepackage{nicefrac}       % compact symbols for 1/2, etc.
\usepackage{microtype}      % microtypography
\usepackage{xcolor}         % colors

\usepackage{amsmath}
\usepackage{amssymb}
\usepackage{mathtools}
\usepackage{amsthm}
\usepackage{wrapfig}

\newcommand{\trace}{\mathbf{tr}}

\newcommand{\reals}{{\mbox{$\mathbb{R}$}}}

\newcommand{\diag}{\mathop{\bf diag}}

\newcommand{\eg}{{\it e.g.}}
\newcommand{\ie}{{\it i.e.}}

\newcommand{\vect}[1]{\text{vec}\left(#1\right)}
\newcommand{\inn}[2]{\left\langle#1, #2\right\rangle}

\newcommand{\size}[1]{\left|#1\right|}
\newcommand{\abs}[1]{\size{#1}}

\newcommand{\bigo}[1]{\mathcal{O}\left(#1\right)}

\newcommand{\norm}[1]{\|#1\|}
\newcommand{\Norm}[1]{\left\|#1\right\|}

\newtheorem{theorem}{Theorem}[section]

\newtheorem{lemma}[theorem]{Lemma}

\usepackage{amsmath,amsfonts,bm}

\def\1{\bm{1}}
\def\0{\bm{0}}

\DeclareMathAlphabet{\mathsfit}{\encodingdefault}{\sfdefault}{m}{sl}
\SetMathAlphabet{\mathsfit}{bold}{\encodingdefault}{\sfdefault}{bx}{n}

\newcommand{\E}{\mathbb{E}}

\usepackage{enumitem} % label for enumerate

\title{Gradient Descent Optimizes Infinite-Depth ReLU Implicit Networks with Linear Widths}

\author{%
  Tianxiang Gao
	\\
  Department of Computer Science\\
  Iowa State University\\
  \texttt{gaotx@iastate.edu} \\
   \And
   Hongyang Gao \\
Department of Computer Science\\
Iowa State University\\
\texttt{hygao@iastate.edu} \\
}

\begin{document}

\maketitle

\begin{abstract}
	Implicit deep learning has recently become popular in the machine learning community since these implicit models can achieve competitive performance with state-of-the-art deep networks while using significantly less memory and computational resources. However, our theoretical understanding of when and how first-order methods such as gradient descent (GD) converge on \textit{nonlinear} implicit networks is limited. Although this type of problem has been studied in standard feed-forward networks, the case of implicit models is still intriguing because implicit networks have \textit{infinitely} many layers. The corresponding equilibrium equation probably admits no or multiple solutions during training. This paper studies the convergence of both gradient flow (GF) and gradient descent for nonlinear ReLU activated implicit networks. To deal with the well-posedness problem, we introduce a fixed scalar to scale the weight matrix of the implicit layer and show that there exists a small enough scaling constant, keeping the equilibrium equation well-posed throughout training. As a result, we prove that both GF and GD converge to a global minimum at a linear rate if the width $m$ of the implicit network is \textit{linear} in the sample size $N$, \ie,  $m=\Omega(N)$.
\end{abstract}

\section{Introduction}
Recently, implicit neural networks attracts increasing attention in the machine learning community, which have achieved competitive or dominated performances of traditional neural networks in various domains such as sequence modeling \cite{bai2019deep} with significantly less usage of computational resources \cite{dabre2019recurrent,dehghani2018universal,bai2018trellis}. In implicit neural networks, the feature vectors are not created recursively as traditional neural networks but provided implicitly through a solution of an equilibrium equation. Implicit neural networks generalize the recursive rules of many commonly used neural network architectures such as feed-forward, convolution, residual, and recurrent networks \cite{bai2019deep, el2019implicit,bai2020multiscale}.
% Moreover, these implicit models have also been shown competitive or even dominated performance of traditional neural networks in domains such as sequence modeling \cite{bai2019deep}, whereas use significantly less memory and fewer computational resources \cite{dabre2019recurrent,dehghani2018universal,bai2018trellis}. 
However, the theoretical understanding of when and how a simple first-order method such as gradient descent (GD) works for these implicit models is limited, though this type of convergence problem has been well studied in standard feed-forward networks \cite{du2019gradient,allen2019convergence,zou2020gradient,nguyen2020global,arora2019exact,oymak2020toward,nguyen2021proof}.
Since implicit networks can have infinitely many layers, the equilibrium equation is not necessarily well-posed as it may admit zero or multiple solutions during training. For example,  \cite{chen2018neural,bai2019deep,bai2021stabilizing,kawaguchi2021theory} all observe instability of forward propagations in implicit models. Specifically, the number of iterations required for forward propagation to find equilibrium points grows with training epochs. Forward propagation likely becomes divergent as training goes longer. A line of recent works make efforts to handle this well-posedness challenge. For example, \cite{el2019implicit} reformulates the training problem in a so-called Fenchel divergence formulation and use the projected gradient descent method to solve the relaxed optimization problem; \cite{winston2020monotone} formulates a splitting problem for the forward propagation of implicit neural networks and use a proximal operation to find the fixed point; \cite{bai2021stabilizing} proposes a special regularization to ensure the well-posedness. However, none of these works can theoretically ensure the convergence of a gradient-based method.

Under some simplified setups, a line of recent works have tried to study this convergence problem from the mathematical theory perspective. For instance, \cite{kawaguchi2021theory} studies this problem for implicit models with linear activation function. By applying an extra softmax layer on the shared weight matrix, the well-posedness challenge is resolved. As a result, they are able to establish the global linear convergence for gradient flow (GF). Unfortunately, their results cannot be extended to nonlinear activation, especially for the nonsmooth ReLU activation, which are critical to the learnability of deep neural networks. Recently, \cite{gao2021global} facility the training process by using a skip connection to the output and introduces a scaling factor to scale the shared weight matrix of the network, which can show there exists a small enough scaling constant which keeps the forward propagation well-posed throughout training. To show the convergence of ReLU-activated implicit neural networks, they take a neural tangent kernel (NTK) method \cite{jacot2018neural}, where the dynamic of the network prediction is governed by a Gram matrix that remains positive definite during training. As a result, they successfully establish the global linear convergence for both GF and GD as long as the width $m$ of the network is quadratic in the sample size $N$, \ie, $m=\Omega(N^2)$. However, their results can only be applied to restricted range of implicit networks due to the special choice of the output layer.

\textbf{Main contribution.} In this paper, we propose to establish the global convergence
results for implicit neural networks with the \textit{nonlinear}
ReLU activation function and \textit{regular} output layer. Specifically, we provide sufficient conditions for the initialization under which GF and GD are guaranteed to converge to a global minimum at a linear rate. Then, we show that all these initial assumptions can be satisfied by a subset of initialization where the network has linear width in the sample size, \ie, $m=\Omega(N)$. For popular random initialization, we show that these initial conditions are satisfied with a high probability (w.h.p.) if the width of a network is quadratic of the sample size, \ie, $m=\Omega(N^2)$. Although these results with similar order of overparameterization have been obtained for finite-depth feed-forward neural networks \cite{huang2020dynamics,nguyen2020global,nguyen2021proof}, it is worth noting that this is the first time, to our best knowledge, that such results are provided for ReLU-activated implicit networks, which could have \textit{infinitely} many layers.

\section{Preliminaries of Implicit Deep Learning}
\textbf{Notation}: For a vector $x$, we use $\norm{x}$ to denote its
Euclidean norm. For a matrix $A$, $\norm{A}$ is
its operator norm, and $\sigma_{\min}(A)$ and $\sigma_{\max}(A)$ denote its smallest and largest singular values, respectively. If $A$ is a square matrix, then
$\lambda_{\min}(A)$ and $\lambda_{\max}(A)$ denote the
smallest and largest eigenvalue of $A$, respectively. We use $\vect{A}$ to denote the vectorization operation applied on the matrix $A$. Given a function $Y=f(X)$, the derivative $\partial f/\partial X$ is defined by $\vect{dY}  = \left(\partial f/\partial X\right)^TdX$, where $X$ and $Y$ can be scalars, vectors, and matrices. We also denote $[N]:=\{1,2,\cdots, N\}$.

Let $X\in \reals^{N\times d}$ and $y\in \reals^N$ be the training data. The implicit neural network we consider in this paper has the transition at the $\ell$-th layer in the following form
\begin{align}
	Z^{\ell}=\sigma(\gamma Z^{\ell-1}A + \Phi), \quad \ell\geq 1,\label{eq: transition}
\end{align}
where $Z^{\ell}\in \reals^{N\times m}$ is the output of the $\ell$-layer with $Z^{0}=0$, the $i$-th row of $\Phi\in \reals^{N\times m}$ is the feature vector $\phi(x)$ under the feature map $\phi:\reals^{d}\rightarrow \reals^m$, $A\in \reals^{m\times m}$ is the weight matrix shared among all implicit layers, $\sigma(u) = \max\{0,u\}$ is the ReLU activation function, and $\gamma\in (0,1)$ is a fixed scalar. As will be shown in Section~\ref{sec: forward}, the choice of $\gamma$ is essential to ensure the existence of the limit $Z=\lim_{\ell\rightarrow \infty} Z^{\ell}$. As $\ell\rightarrow \infty$, an implicit neural network can be considered as a neural network with \textit{infinitely} many layers. Consequently, $Z$ is not only the limit of the sequence $\{Z^{\ell}\}_{\ell=1}^{\infty}$ but also an equilibrium point or fixed point of the following equilibrium equation:
\begin{align}
	Z = \sigma(\gamma ZA + \Phi).\label{eq:equi eq}
\end{align}
In general, the feature map $\phi$ is a nonlinear function, which extracts the hidden features from the low-dimensional input. In this paper, we consider the feature map $\phi$ as a simple one layer neural network that is activated by ReLU activation function, \ie, 
\begin{align}
	\Phi:=\phi(X)=\sigma(XW),\label{eq: def Phi}
\end{align}
where $W\in \reals^{d\times m}$ is also a trainable weight matrix. The training loss is given by
\begin{align}
	L(\theta)=\frac{1}{2}\norm{\hat{y}-y}^2,
\end{align}
where $\hat{y}=Zb$ is the prediction with weight vector $b\in \reals^m$ and $\theta:=\vect{W,A,b}$ is the collection of all trainable parameters.

\section{Well-Posedness of the Forward and Backward Propagation}\label{sec: forward}
In this section, we establish sufficient conditions for the equilibrium equation Eq.\eqref{eq:equi eq} to be well-posed in the sense that the existence of the equilibrium point $Z$ is uniquely determined. Provided the existence of the equilibrium point $Z$, we can derive the gradients of the parameters by using the implicit function theorem instead of back-propagating all intermediate layers.
Previous work \cite{gao2021global} has shown the existence of the unique equilibrium point in the vector-based equilibrium equation Eq. \eqref{eq:equi eq} if the scalar $\gamma > 0$ is chosen small enough. We extend this result to a general matrix-based mapping.
\begin{lemma}\label{lemma: forward}
	Suppose $\norm{A}\leq M$ for some constant $M > 0$ and choose the scalar $\gamma>0$ such that $\gamma_0:=\gamma M < 1$. Then the existence of the fixed point $Z$ is uniquely determined. Moreover, we have $\norm{Z^{\ell}}_F\leq \frac{1}{1-\gamma_0}\norm{\Phi}_F$ for all $\ell$, hence $\norm{Z}_F\leq \frac{1}{1-\gamma_0}\norm{\Phi}_F$.
\end{lemma}

Lemma~\ref{lemma: forward} shows that the transition Eq.\eqref{eq: transition} is a contraction mapping if a small enough scalar $\gamma$ is selected for which $\gamma\norm{A}<1$. The proof is provided in Appendix~\ref{app: forward}. However, the operator norm of $A(k)$ changes throughout the training process. In general, a fixed scalar $\gamma$ cannot guarantee the well-posedness during the entire training. That is the main reason \cite{chen2018neural,bai2019deep,bai2021stabilizing,kawaguchi2021theory} observe the number of forward iterations required to find a fixed point gradually grows with training epochs. Thus, a simple strategy is to find a sequence $\{\gamma_k\}_{k=1}^{\infty}$ of appropriate scalars for each iteration that ensure the equilibrium equation are kept well-posed over all iterations. However, finding an appropriate scalars at each iteration is computationally expensive, since it needs to compute quantities related to the operator norm of the weight matrix $A(k)$ (\eg, \cite{el2019implicit,bai2021stabilizing}). Fortunately, we can show $\norm{A(k)}$ is uniformly upper bounded by some constant. As a result, there indeed is a small constant for which the forward pass is kept well-posed throughout training.

A finite-depth neural network needs to store all intermediate parameters and apply backpropagation to compute the gradients of each weight matrix or vector. Surprisingly, this expensive computation is not necessary for implicit neural networks. Instead, the implicit function theorem provides an efficient way to derive the gradients since the fixed point $Z$ is a root of the function $f$ given by
\begin{align}
	f(Z,A,W):=Z-\sigma(\gamma ZA+\Phi).
\end{align}
We can easily show that the partial derivative $\partial f/\partial Z$ is invertible, provided $\norm{A} < \gamma^{-1}$. As a result, we obtain the partial derivatives $\partial Z/\partial A$ and $\partial Z/\partial W$ by applying the implicit function theorem. Then the gradients are obtained by using the chain rule. Appendix~\ref{app: gradient} includes the detailed derivation.

\begin{lemma}\label{lemma: gradient}
	Suppose $\norm{A}\leq M$ for some constant $M > 0$ and choose the scalar $\gamma>0$ such that $\gamma_0:=\gamma M < 1$. Then
	\begin{align}
		\lambda_{\min}\{I_{Nm}-\gamma D(A^T\otimes I_N)\}> 1-\gamma_0 > 0.
	\end{align}		
	Hence the matrix $Q:=I_{Nm}-\gamma D(A^T\otimes I_N)$ is invertible, and 
	\begin{align}
		\frac{\partial L}{\partial W} 
		=&\left[DE \left(I_m\otimes X\right) \right]^T
		Q^{-T}\left(b^T\otimes I_N\right)^T(\hat{y}-y),\\
		\frac{\partial L}{\partial A} 
		=& \gamma \left[D (I_m \otimes Z) \right]^T
		Q^{-T}\left(b^T\otimes I_N\right)^T(\hat{y}-y),\\
		\frac{\partial L}{\partial b} 
		=& Z^T(\hat{y}-y),
	\end{align}
	where $D:=\diag[\vect{\sigma^{\prime}(\gamma  ZA+\Phi)}]$, and $E:=\diag\left[\vect{\sigma^{\prime}(XW)}\right]$.
%	\begin{align}
%		D:=&\diag[\vect{\sigma^{\prime}(\gamma  ZA+\Phi)}],\\
%		E:=&\diag\left[\vect{\sigma^{\prime}(XW)}\right].
%%		Q:=&I_{Nm}-\gamma D(A^T\otimes I_N).
%	\end{align}
	
\end{lemma}

\section{Main results}
In this section, we first study the dynamics of the prediction $\hat{y}(t)$ that is induced by the gradient flow. It can be shown that a time-variant Gram matrix controls the dynamics of the prediction. At the same time, the spectral property of the Gram matrix is consistent throughout the training as long as the network is over-parameterized. Based on the findings in gradient flow analysis, we show that gradient descent with a fixed step size converges to a global minimum of the implicit network at a linear rate. 

\subsection{Continuous time analysis: convergence of gradient flow}
The gradient flow is given by
% \begin{align*}
	$\frac{d\theta}{dt}=-\frac{\partial L(t)}{\partial \theta}$,
% \end{align*}
where $L(t):=L(\theta(t))$ is the corresponding loss function for $\theta(t)$ at time $t$.
By using the chain rule, we derive the dynamics of the prediction $\hat{y}(t)$ in the following lemma, and the derivation is deferred in Appendix~\ref{app: dynamics}.
\begin{lemma}\label{lemma: dynamics prediction}
	Assume $\norm{A(t)}\leq M$ for all $t\geq 0$ and choose the scalar $\gamma>0$ small enough such that $\gamma_0:=\gamma M < 1$. Then the dynamics of the prediction $\hat{y}(t)$ is given by
	\begin{align}
		\frac{d\hat{y}}{dt}=-H(t)(\hat{y}(t)-y),
	\end{align}
	where
	\begin{align}
		H(t):=&Z(t)Z(t)^T+M(t)M(t)^T+\Pi(t)\Pi(t)^T,\\
		M(t):=&\gamma [b(t)^T\otimes I_N]Q(t)^{-1}D(t)[I_m \otimes Z(t)],\\
		\Pi(t):=&\left[b(t)^T\otimes I_N\right]Q(t)^{-1}D(t)E(t)\left[I_m\otimes X\right].
	\end{align}
\end{lemma}
Clearly, the matrix $H(t)$ is positive semidefinite. If there exists a strictly positive scalar $\lambda_0>0$ for which $\lambda_{\min}(H(t))\geq \lambda_0$ for all $t\geq 0$, then $L(t)$ consistently decreases to zero at a linear rate, \ie, $L(t)\leq \exp\left\{-\lambda_0t\right\}L(0)$. Thus, the problem is reduced to show that the smallest singular value of at least one of the matrices $Z(t)$, $M(t)$, and $\Pi(t)$ is lower bounded throughout the training. By using simple matrix analysis results, however, we obtain the following inequalities:
\begin{align*}
	\sigma_{\min}\left[M(t)\right]\geq& \frac{\gamma}{1-\gamma_0} \norm{b(t)}\min_{i}\{D_{ii}\}\sigma_{\min}\left[Z(t)\right],\\
	\sigma_{\min}\left[\Pi(t)\right]\geq &\frac{1}{1-\gamma_0}\norm{b(t)}\min_{i}\{D_{ii}\}\min_{i}\{E_{ii}\}\sigma_{\min}\left(X\right).
\end{align*}

Thus, to lower bound the singular values of matrices $M(t)$ and $\Pi(t)$, one must make extra assumptions on the data sample $X$ and activation $\sigma$. An example method is introduced in \cite{nguyen2020global}, where the neural network has to follow a pyramidal structure and the activation function has to be sufficiently smooth. This method does not hold for ReLU activation due to the non-smoothness. In contrast, we adopt the method proposed by \cite{nguyen2021proof} for finite-depth ReLU neural network, where their analysis focuses on the evolution of the last layer of the network, that is, $Z(t)$ in the implicit neural network. With appropriate assumptions on the initial conditions, we can establish the global convergence result for the gradient flow in Theorem~\ref{thm: gradient flow}, and the entire proof is provided in Appendix~\ref{app: gradient flow}.
\begin{theorem}\label{thm: gradient flow}
	Let $C_1, C_2, C_3>0$ be given positive numbers. Denote $\alpha_0:=\sigma_{\min}(Z(0))$, $\lambda_1:=\norm{W(0)}  +C_1$, $\lambda_2:=\norm{A(0)}+C_2$, and $\lambda_3:=\norm{b(0)} + C_3$. Choose $\gamma > 0$ small enough for which $\gamma_0:=\gamma \lambda_2 < 1$.
	Assume the following conditions are satisfied at initialization
	\begin{align}
		\alpha_0^2\geq& \frac{4}{1-\gamma_0}
		\lambda_0\norm{X}_F\norm{\hat{y}(0)-y},\label{eq:GF1}\\
		\alpha_0^3\geq& \left[1+\frac{\gamma_0^2}{(1-\gamma_0)^2}\frac{\lambda_1^2}{\lambda_2^2}\right]\frac{8\lambda_3}{(1-\gamma_0)^2}
		\norm{X}_F^2 \norm{\hat{y}(0)-y},\label{eq:GF2}
	\end{align} 
	where $\lambda_0:=\max\left\{\frac{\lambda_1}{C_3}, \frac{\gamma_0\lambda_1\lambda_3}{(1-\gamma_0)C_2\lambda_2}, \frac{\lambda_3}{C_1}\right\}$.
%	\begin{align}
%		\lambda_0:=\max\left\{\frac{\lambda_1}{C_3}, \frac{\gamma_0\lambda_1\lambda_3}{(1-\gamma_0)C_2\lambda_2}, \frac{\lambda_3}{C_1}\right\}.\label{eq:lambda}
%	\end{align}
	Then for all $t\geq 0$ the followings hold:
	
	\begin{enumerate}[label=(\roman{*})]
		\item $\norm{W(t)}\leq \lambda_1$, $\norm{A(t)}\leq \lambda_2$, $\norm{b(t)}\leq \lambda_3$,
		%		\item $\norm{A(t)}\leq \lambda_2$, 
		%		\item $\norm{b(t)}\leq \lambda_3$,
		\item $\sigma_{\min}(Z(t))\geq \alpha_0/2$,
		\item $L(t)\leq \exp\{-\alpha_0^2t/2\}L(0)$, where $L(t):=L(\theta(t))$.
	\end{enumerate}
\end{theorem}
Theorem~\ref{thm: gradient flow} shows that the operator norm $A(t)$ is upper bounded by the constant $\lambda_2$ throughout training, \ie, $\norm{A(t)}\leq \lambda_2$. By choosing $\gamma>0$ small enough, the forward propagation is kept well-posed throughout training. Moreover, Theorem~\ref{thm: gradient flow} also indicates that the smallest singular value of $Z(t)$ is lower bounded by the constant $\alpha_0/2$ during training. As a result, the training loss consistently decreases to zero at a linear rate. It is worth noting that the network only needs to have linear width $m$ in the sample size $N$ as long as the initial conditions \eqref{eq:GF1}-\eqref{eq:GF2} are satisfied. Section~\ref{sec:initial} will provide concrete examples to show that these initial conditions are indeed satisfied by using only linear widths.

\subsection{Discrete time analysis: convergence of gradient descent}
By applying the Euler method to the gradient flow with stepsize  $\eta>0$, we obtain the gradient descent as follows
% \begin{align*}
	$\theta(k+1)=\theta(k)-\eta \frac{\partial L(k)}{\partial \theta}$.
% \end{align*}
Unlike the continuous analysis in the gradient flow, we don't have the explicit formula of the dynamics of the prediction $\hat{y}(k)$ in the discrete time analysis. Instead, we need to first show the difference of equilibrium points $Z(k)$ in two consecutive iterations. The following result serves this purpose by providing a bound between two equilibrium points based on their parameters. We defer the proof in Appendix~\ref{app: Z_a-Z_b}.

\begin{lemma}\label{lemma: Z_a-Z_b}
	Given matrices $(W_a, A_a)$ and $(W_b, A_b)$, let $\lambda_1:=\max\{\norm{W_a}, \norm{W_b}\}$, and $\lambda_2:= \max\{\norm{A_a}, \norm{A_b}\}$. Choose $\gamma > 0$ small enough for which $\gamma_0:=\gamma \lambda_2 < 1$. Then the corresponding equilibrium points exist and are denoted by $Z_a$ and $Z_b$, respectively. Moreover, we have
	\begin{align}
		\norm{Z_a-Z_b}\leq \frac{\norm{X}_F}{1-\gamma_0}\left[\frac{\gamma_0}{1-\gamma_0}\frac{\lambda_1}{\lambda_2}\norm{A_a-A_b} + \norm{W_a-W_b}\right].
	\end{align}
%	where
%	\begin{align}
%		Q:=\left[\frac{\gamma_0}{1-\gamma_0}\frac{\lambda_1}{\lambda_2}\norm{A_a-A_b} + \norm{W_a-W_b}\right]\norm{X}_F.
%	\end{align}
\end{lemma}
Lemma~\ref{lemma: Z_a-Z_b} allows us to derive the relationship of the predictions $\hat{y}(k)$ between two consecutive iterations. The loss can be shown to consistently decrease as long as the step size selected is small enough. To satisfy the conditions of the step size, more initial conditions are needed to obtain the convergence result for gradient descent. The convergence result of the gradient descent is provided in the following theorem, and the proof is included in Appendix~\ref{app: gradient descent}.
\begin{theorem}\label{thm: gradient descent}
	Let $C_1, C_2, C_3>0$ be given positive numbers. Denote $\alpha_0:=\sigma_{\min}(Z(0))$, $\lambda_1:=\norm{W(0)}  +C_1$, $\lambda_2:=\norm{A(0)}+C_2$, and $\lambda_3:=\norm{b(0)} + C_3$. Choose $\gamma > 0$ small enough for which $\gamma_0:=\gamma \lambda_2 < 1$.
	Assume the following conditions are satisfied at initialization
	\begin{align}
		\alpha_0^2\geq& \frac{8}{1-\gamma_0}\lambda_0\norm{X}_F\norm{\hat{y}(0)-y}\label{eq:GD1}\\
		\alpha_0^3\geq& \left[1 + \frac{\gamma_0^2}{(1-\gamma_0)^2}\frac{\lambda_1^2}{\lambda_2^2}\right]
		\frac{16\lambda_3}{(1-\gamma_0)^2}
		\norm{X}_F^2\norm{\hat{y}(0)-y}\label{eq:GD2}\\
		\alpha_0^2\geq& 
		\left[1+\frac{\gamma_0^2}{(1-\gamma_0)^2}\frac{\lambda_1^2}{\lambda_2^2}\right]\frac{16\lambda_3^2}{(1-\gamma_0)^2}\norm{X}_F^2
		\label{eq:GD3}
	\end{align} 
	where $\lambda_0$ is defined in Theorem~\ref{thm: gradient flow}.
%	Eq.\eqref{eq:lambda}.
	Choose step size $\eta>0$ such that 
	\begin{align}
		\eta< \min\left\{ \frac{4}{\alpha_0^2},
		2(1-\gamma_0)^2 \overline{\lambda} \lambda_1^{-4}\lambda_3^{-2}\norm{X}_F^{-2}
		\right\},
		\label{eq:step size}
	\end{align} 
	where $\overline{\lambda}:=\left[1+\frac{\gamma_0^2}{(1-\gamma_0)^2}\frac{\lambda_1^2}{\lambda_2^2}\right]\left(\left[1+\frac{\gamma_0^2}{(1-\gamma_0)^2}\frac{\lambda_1^2}{\lambda_2^2}\right]\lambda_1^{-2}+\lambda_3^{-2}\right)^{-2}$.
%	\begin{align}
%		\overline{\lambda}:=\frac{\left[1+\frac{\gamma_0^2}{(1-\gamma_0)^2}\frac{\lambda_1^2}{\lambda_2^2}\right]}{\left(\left[1+\frac{\gamma_0^2}{(1-\gamma_0)^2}\frac{\lambda_1^2}{\lambda_2^2}\right]\lambda_1^{-2}+\lambda_3^{-2}\right)^2}.
%	\end{align}
	Then the followings holds $\forall k\geq 0$
	\begin{enumerate}[label=(\roman{*})]
		\item $\norm{W(k)}\leq \lambda_1$, $\norm{A(k)}\leq \lambda_2$, $\norm{b(k)}\leq \lambda_3$,
		\item $\sigma_{\min}(Z(k))\geq \alpha_0/2$,
		\item $L(k)\leq (1-\eta\alpha_0^2/4)^kL(0)$, where $L(k):=L(\theta(k))$.
	\end{enumerate}
\end{theorem}
With some extra conditions on initialization and stepsize, Theorem~\ref{thm: gradient descent} obtains similar results for gradient descent as in Theorem~\ref{thm: gradient flow} for gradient flow. In particular, the well-posedness is guaranteed throughout training using a singular constant $\gamma$. The smallest singular value $\sigma_{\min}(Z(t))$ remains greater than the constant $\alpha_0/2$ so that the loss $L(k)$ continuously decreases to zero at a linear rate. Similarly, these results only need the implicit network with linear width in the sample size $N$. However, these results are not directly comparable because it remains unclear how likely the initial conditions \eqref{eq:GD1}-\eqref{eq:GD3} are satisfied. Thus, we reserve Section~\ref{sec:initial} to further analyze on the satisfiability of the initial assumptions. Specifically, we address this issue by providing concrete examples by which the initial assumptions are indeed satisfied by only using linear widths.

\section{Satisfiability of the initial assumptions}\label{sec:initial}
This section provides concrete examples for which the initial conditions \eqref{eq:GD1}-\eqref{eq:GD3} of Theorem~\ref{thm: gradient descent} are satisfied.
Since the equilibrium point $Z(k)$ is provided implicitly, the singular values of $Z(k)$ are difficult to determine, even at initialization. What exacerbates the problem is the nonlinearity of the ReLU activation $\sigma$. Fortunately, we can find a subset of initialization that can easily induce the singular values of $Z$ by using homogeneity and nonnegativity of $\sigma$. The following result provides an explicit expression for the equilibrium point $Z$ in terms of the feature matrix $\Phi$ as long as $A$ is appropriately initialized. The proof is given in Appendix~\ref{app: initial}.

\begin{lemma}\label{lemma: initial}
	Assume $A_{ij}\geq 0$ for all $i,j\in [m]$ and $\gamma>0$ is chosen for which $\norm{A} < \gamma^{-1}$. Then $Z=\Phi \left(I_m-\gamma A\right)^{-1}$.
\end{lemma}

Suppose that we are given $A(0)\in \reals^{m\times m}$ with $A_{ij}\geq 0$ for all $ i,j\in [m]$ and $\norm{A(0)}\leq \lambda_2$ for some positive constant $\lambda_2$. By choosing $\gamma>0$ small enough for which $\gamma_0=\gamma\lambda_2 < 1$, Lemma~\ref{lemma: initial} implies that $\alpha_0:=\sigma_{\min}[Z(0)]\geq \sigma_{\min}[\Phi(0)]/(1+\gamma_0)$. Therefore, the satisfiability of the assumption made in Theorem~\ref{thm: gradient descent} is reduced to find an appropriate $\Phi(0)$ or $W(0)$ for which $\sigma_{\min}[\Phi(0)]/(1+\gamma_0)$ satisfies the initial conditions \eqref{eq:GD1}-\eqref{eq:GD3}. Some mostly recent literature  \cite{zhang2021understanding,du2019gradient,nguyen2020global,nguyen2021tight,nguyen2021proof} have been provided rigorously theoretical analyses and concrete examples to demonstrate that the initial conditions are easily satisfied with different types of initialization. The rest of this section provides concrete examples to show a subset of initialization that satisfies these assumptions.

\subsection{Linear width $m=\Omega(N)$ suffices the initial conditions}
Specifically, we apply Theorem~\ref{thm: gradient descent} for $C_1=C_2=C_3=1$. Let $\theta(0):=\vect{W(0), A(0), b(0)}$, where we choose $W(0)$ such that $\sigma_{\min}[\Phi(0)] > 0$, $A(0)_{ij}\geq0$ with $\norm{A(0)}=\norm{W(0)}$, and $b(0)=0$. For $\sigma_{\min}[\Phi(0)]>0$, a concrete example is that $W(0)$ is chosen according to LeCun's initialization (\eg, see \cite[Section~3.1]{nguyen2021proof},\cite[Section~3.1]{nguyen2020global}). It is easy to come up with examples for which $A(0)_{ij}\geq 0$ and $\norm{A(0)}=\bigo{\norm{W(0)}}$. For example, $A(0):=\norm{W(0)}I_m$. Thus, for simplicity, we assume $\norm{A(0)}=\norm{W(0)}$, then $\lambda_2 = \lambda_1$. Since $b(0)=0$, we have $\lambda_3=1$, and $\norm{\hat{y}(0)-y}=\norm{y}$. Next, we can choose $\gamma>0$ small enough for which $\gamma_0=1/2$. It follows from Lemma~\ref{lemma: initial} that the initial conditions \eqref{eq:GD1}-\eqref{eq:GD3} become
\begin{align}
	\left(\frac{2}{3}\sigma_{\min}[\Phi(0)]\right)^2\geq&16\lambda_1\norm{X}_F\norm{y}\label{eq:condition1}\\
	\left(\frac{2}{3}\sigma_{\min}[\Phi(0)]\right)^3\geq&128\norm{X}_F^2\norm{y}\label{eq:condition2}\\
	\left(\frac{2}{3}\sigma_{\min}[\Phi(0)]\right)^2\geq&128\norm{X}_F^2.\label{eq:condition3}
\end{align} 
Let $\tilde{\theta}(0):=\beta \theta(0)$ for some $\beta > 0$ and denote $\alpha_1:=\sigma_{\min}[\Phi(0)]$. 
Condition \eqref{eq:condition1} at $\tilde{\theta}(0)$ becomes
\begin{align}
	\left(\frac{2}{3}\beta\alpha_1\right)^2\geq&16(\beta\norm{W(0)}+1)\norm{X}_F\norm{y}.\label{eq:linear width 1}
\end{align}
The left-hand side (LHS) of the above inequality is a polynomial of degree $2$ in $\beta$, whereas the RHS is linear in $\beta$. Thus, as $\norm{W(0)}$ is fixed in the inequality above, the inequality \eqref{eq:linear width 1} is satisfied as long as $\beta$ is chosen large enough. Similarly, the condition \eqref{eq:condition2} and \eqref{eq:condition3} are satisfied for $\tilde{\theta}(0)$ if 
\begin{align}
	\left(\frac{2}{3}\beta\alpha_1\right)^3\geq&128\norm{X}_F^2\norm{y},\label{eq:linear width 2}\\
\left(\frac{2}{3}\beta\alpha_1\right)^2\geq&128\norm{X}_F^2.\label{eq:linear width 3}
\end{align}
The RHS of the above inequalities are all constants, whereas the LHS are polynomials of degree $3$ and $2$ in $\beta$, respectively. Thus, the conditions \eqref{eq:linear width 2} and \eqref{eq:linear width 3} are also satisfied for large enough $\beta$. As a result, all the initial conditions of Theorem~\ref{thm: gradient descent} are satisfied at $\tilde{\theta}(0)$ for large $\beta$. 

\subsection{Random initialization: width $m=\Omega(N^2)$ suffices the initial conditions}
We show that conditions \eqref{eq:GD1}-\eqref{eq:GD3} are satisfied under similar random initialization as LeCun's Initialization \cite{lecun2012efficient}, Xavier Initialization \cite{glorot2010understanding}, and He Initialization \cite{he2015delving}, provided a stronger condition $m=\Omega(N^2)$.
For simplicity, we assume $\norm{x_i}=1$ and $\abs{y_i}=\bigo{1}$ for all $i\in [N]$. Then $\norm{X}_F=\sqrt{N}$ and $\norm{y}=\bigo{\sqrt{N}}$. Consider initialization
\begin{align}
	b(0)_{i}\overset{i.i.d.}{\sim} \mathcal{N}(0,1/m),
	\quad
	A(0)_{ij}\overset{i.i.d.}{\sim} \abs{\mathcal{N}}(0,1),
	\quad
	W(0)_{ij}\overset{i.i.d.}{\sim} \mathcal{N}(0,1),
\end{align}
where $\abs{\mathcal{N}}$ stands for \textit{half-normal distribution}. It follows from Theorem~4.4.5 of \cite{vershynin2018high} that (with high probability) $\norm{b(0)}= \bigo{1}$, $\norm{W(0)}= \bigo{\sqrt{m}}$, and $\norm{A(0)}= \bigo{\sqrt{m}}$. For $C_1=C_2=C_3=1$, we have $\lambda_1= \bigo{\sqrt{m}}$, $\lambda_2= \bigo{\sqrt{m}}$, $\lambda_3= \bigo{1}$. 
We can choose $\gamma>0$ small enough to ensure $\gamma_0 = 1/2$. By using standard concentration argument and Lemma~\ref{lemma: initial}, we obtain
\begin{align*}
	\norm{\hat{y}(0)} =\norm{Z(0)b(0)}=\bigo{\Norm{\Phi(0)\left[I_m-\gamma A(0)\right]^{-1}}}=\bigo{\norm{X}_F}.
\end{align*}
Since $\norm{X}_F=\sqrt{N}$ and $\norm{y}= \bigo{\sqrt{N}}$, the initial conditions \eqref{eq:GD1}-\eqref{eq:GD3} are reduced to
\begin{align}
	\alpha_1^2=\Omega\left(N\sqrt{m}\right).
\end{align} 

Using Matrix-Chernoff inequality, one can easily show $\alpha_1^2\geq m\lambda_*/4$ (see Lemma~5.2 of \cite{nguyen2021tight}) with a probability of at least $1-\delta$,
% \begin{align}
% 	\alpha_1^2\geq m\lambda_*/4,
% \end{align}
if $m= \tilde{\Omega}(N/\lambda_*)$ holds, where
$\lambda_*=\lambda_{\min}\left(G_*\right)$ with $G_*:=\E_{w\sim \mathcal{N}(0,I_d)}\left[\sigma(Xw)\sigma(Xw)^T\right]$
and $\tilde{\Omega}$ omits logarithmic factors depending on
$\delta$. It can be shown $\lambda_* > 0$ under some mild data
assumption. For example, $\lambda_*>0$ if no two data are
parallel to each other by Lemma~3.2 of \cite{gao2021global}. Additionally, we have
\begin{align*}
	\lambda_*\leq \frac{\trace(G_*)}{N}=\frac{\mathbb{E}\norm{\sigma(Xw)}^2}{N}
	\leq \frac{\mathbb{E}\norm{Xw}^2}{N}
	=\frac{\norm{X}_F^2}{N}=1.
\end{align*}

Therefore, all the initial conditions
are satisfied for $m=\Omega(N^2\lambda_*^{-2})$. Then, if we additionally assume that the data points follow some sub-Gaussian distribution, \cite[Theorem~3.3]{nguyen2020global} implies that $\lambda_*=\Theta(1)$. Thus, all initial conditions are satisfied for $m=\Omega(N^2)$.

\section{Related works}
The convergence problem of gradient flow and gradient descent for standard deep networks has been studied recently \cite{du2019gradient,allen2019convergence,zou2020gradient,nguyen2020global,arora2019exact,nguyen2021proof}. To show the convergence, the most common strategy is NTK method \cite{jacot2018neural}, where it can be shown that dynamic of the prediction is governed by a gram matrix whose smallest eigenvalues is kept strictly positive during training. The previous work of  \cite{du2019gradient,allen2019convergence,zou2020gradient,arora2019exact} requires all hidden layers have large widths since their analyses rely on studying various quantities related to the changes in the activation patterns during training. In \cite{nguyen2020global}, the authors claim one wide hidden layer is enough if the network follows a pyramidal topology and the derivative of activation is lower bounded. Moreover, they prove that linear width is enough for overparameterization, while previous works need a width of at least $\Omega(N^8)$. The most recent study \cite{nguyen2021proof} further confirms linear width suffices global convergence for ReLU activated networks by focusing the analysis on the last hidden layer so that the pyramidal topology is not necessary. However, none of the previous works can be applied to implicit networks directly since implicit networks have infinitely many layers, and equilibrium equations may not be well-posed during training.

The well-posedness of the forward propagation is the main challenge for any infinite-depth or implicit-depth networks. A line of recent works \cite{winston2020monotone,bai2021stabilizing,chen2018neural,bai2019deep} have shown the instability of forward propagation in implicit networks, and the probability of divergent forward pass gradually raises with training epochs. Several strategies are suggested to deal with the well-posedness problem, \eg, adding constraints \cite{el2019implicit} or regularization\cite{bai2021stabilizing}, reformulates the forward propagation \cite{winston2020monotone}. Unfortunately, none of them can theoretically guarantee convergence. By using an extra softmax layer on the weight matrix, \cite{kawaguchi2021theory} proves the convergence of the gradient flow, but the result only holds for linear activation while nonlinearity is critical for learnability. To deal with well-posedness, our strategy is similar to \cite{gao2021global} by introducing a scalar in front of the weight matrix. The well-posedness problem is resolved since we can find a sufficiently small constant scalar. The novelty of this paper is that implicit network considered in this paper has regular output while \cite{gao2021global} uses a skip connection to facilitate training process. Moreover, this is the first work, to our best knowledge, that establishes the global convergence for an infinite-depth (weight tying) network with linear width.

\section{Experimental Results}
\begin{figure*}
	\centering
	\includegraphics[width=.31\textwidth]{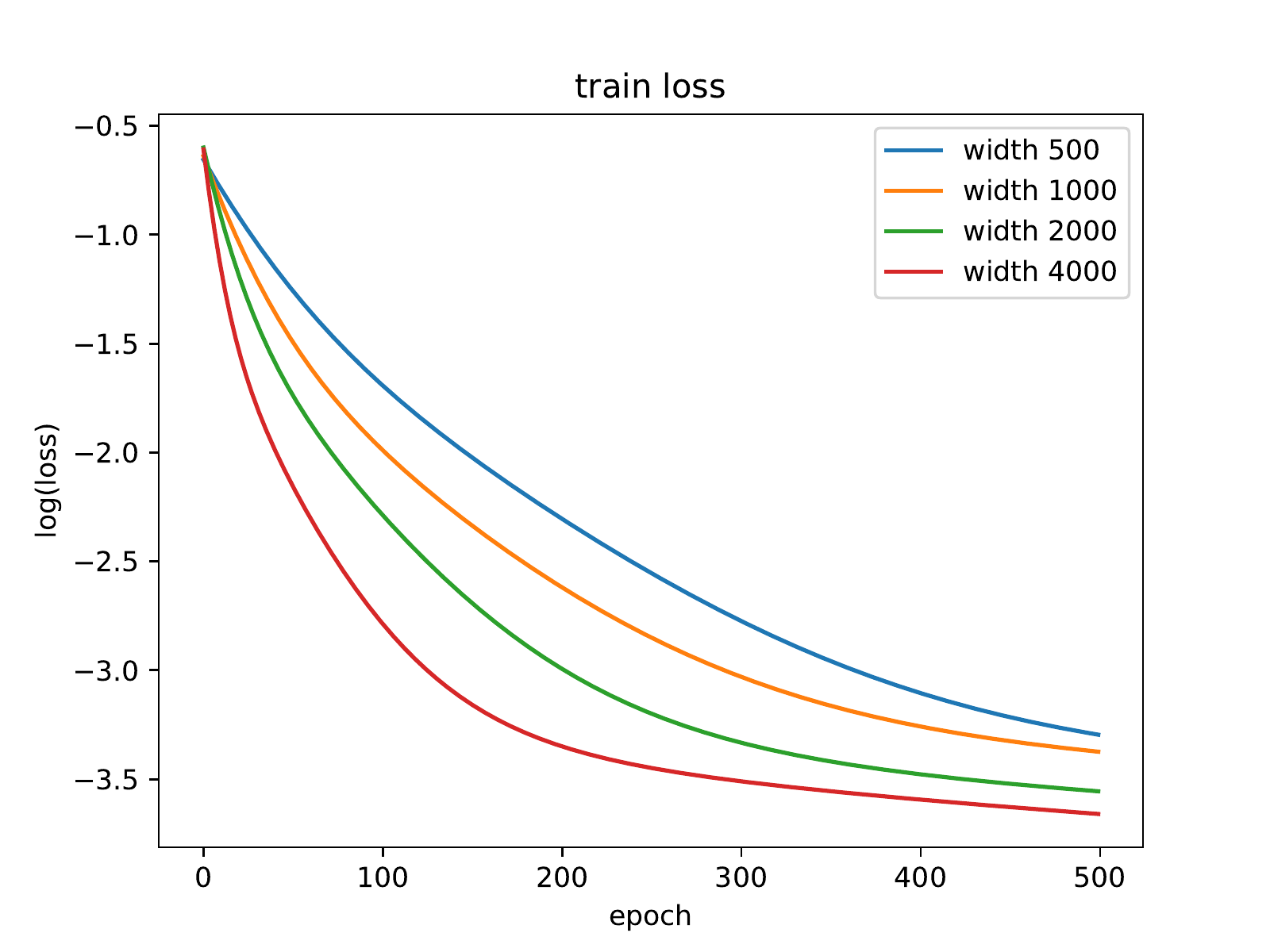}
	\includegraphics[width=.31\textwidth]{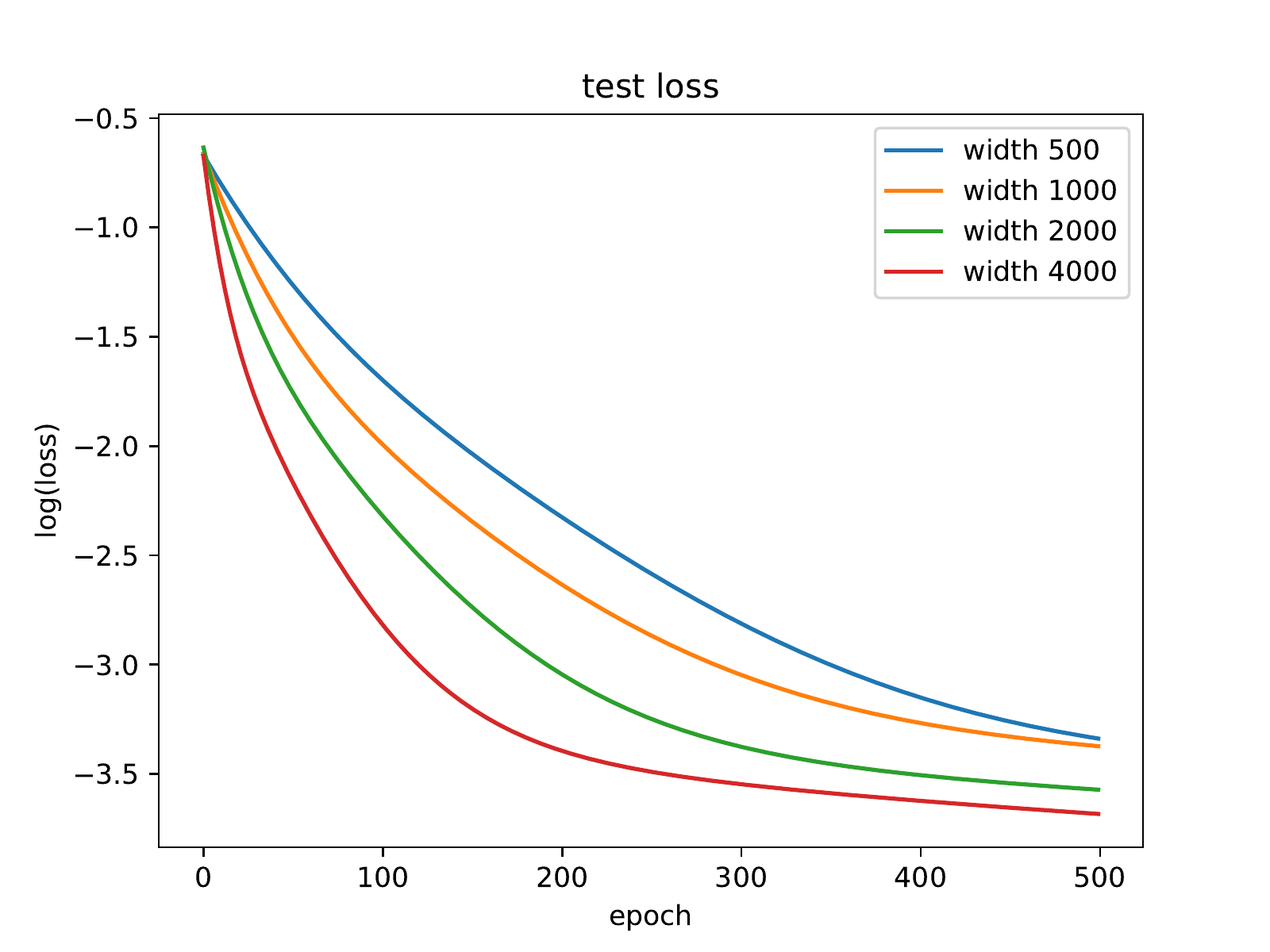}
	\includegraphics[width=.31\textwidth]{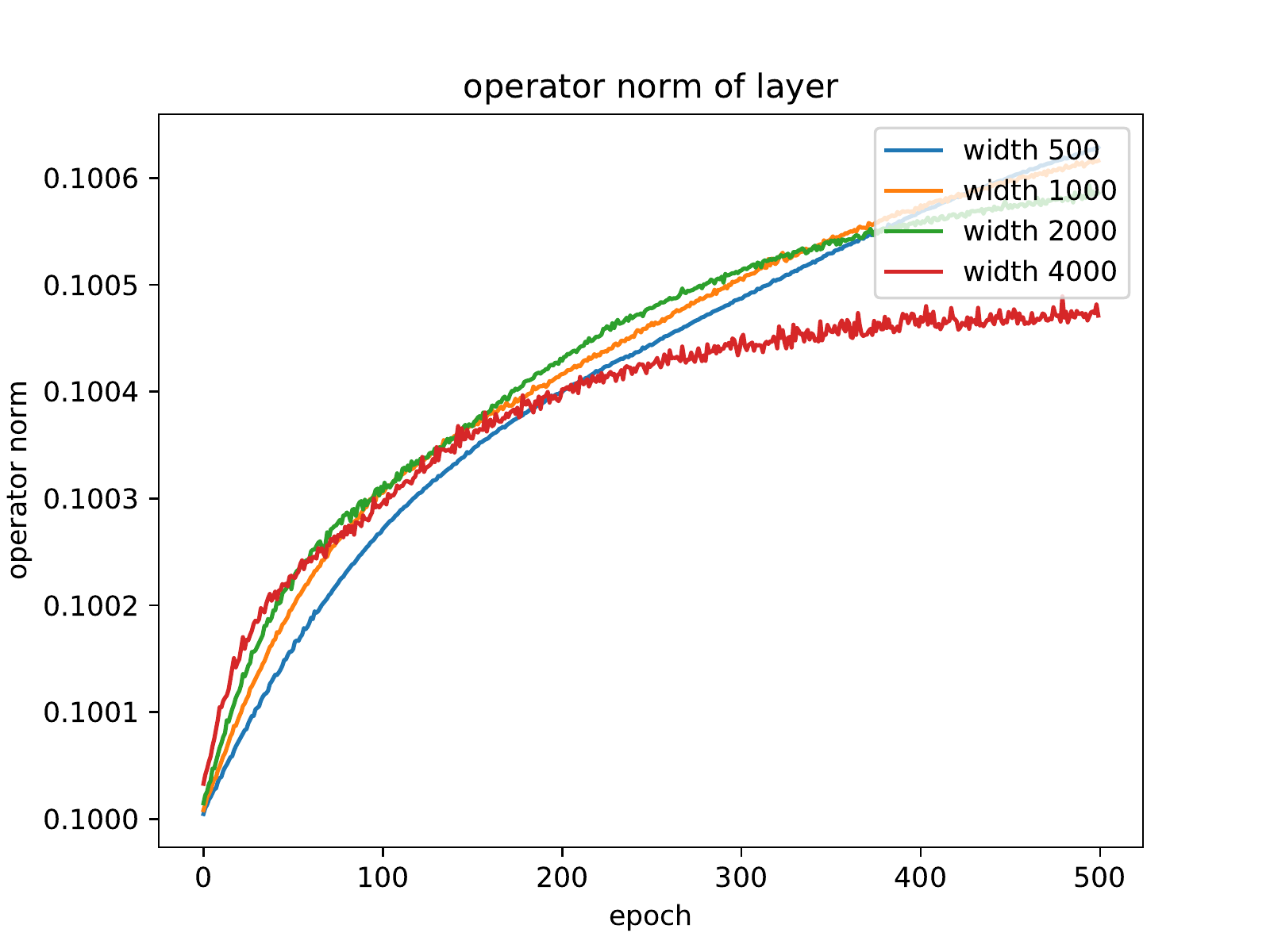}
	\caption{We evaluate the impact of the \textbf{width $m$} on the
		training loss, test loss, and operator norm of the scaled
		matrix $\gamma A(k)$ on the modified dataset of MNIST.}
	\label{fig:fig1}
	% \vspace{-.2in}
\end{figure*}

We evaluate our results use real-world datasets like MNST, FashionMNST, CIFAR10, and SVHN.
\textbf{Experimental Setups.}
% \subsection{Experimental Setups}
For each dataset, we use classes $0$ and $1$, and
$500$ samples are randomly drawn from each class to generate the
training dataset of $N=1000$. All data samples are
converted to gray scale and resized to $28\times 28$. We also normalize each data to have unit norm. We run
$500$ epochs of gradient descent with a fixed step-size. 
It follows from the analysis of Section~\ref{sec:initial} that we initialize $W(0)_{ij}\overset{\text{i.i.d.}}{\sim}\mathcal{N}(0,1/m)$. For simplicity, we set $A(0)=I_m$ and $b(0)=0$, so that $\lambda_1=\lambda_2=\lambda_3=\bigo{1}$. Let $\theta(0):=\vect{W(0), A(0), b(0)}$. Then we multiply $\theta(0)$ by a large $\beta>0$ if $\theta(0)$ does not satisfies the initial conditions \eqref{eq:GD1}-\eqref{eq:GD3}. Moreover, it follows from Theorem~\ref{thm: gradient descent} that we can set $\eta=N^{-1}$, \ie, $\eta=10^{-3}$, to ensure consistent descent in loss function.

\subsection{Over-Parameterization Study}

In this work, some of our finds are based on the setting of over-parameterization. Thus, we first study how overparameterization affects the convergence rates of the gradient descent method. In addition, we also study the impact of overparameterized implicit networks on unseen test data. Similarly, the test data is constructed by randomly selecting $500$ unseen samples from each class. Third, we study how overparameterization affects the changes of $\norm{A(k)}$ from its initialization, since consistent $\norm{A(k)}$ is critical for the well-posedness of the forward propagation. Here, ``operator norm'' in the plots denotes $\gamma\norm{A(k)}$ with $\gamma=0.1$.

Due to the space limit, Figure~\ref{fig:fig1} only includes the result of the dataset MNIST, and the rest results for other real datasets are included in Appendix~\ref{app: exp}. The first sub-figure in Figure~\ref{fig:fig1} shows that as $m$ becomes larger, better convergence rates can be observed. Accordingly, the second sub-figure shows that the neural networks can achieve lower test loss as $m$ becomes larger. The third sub-figure shows that the operator norms are changed throughout the training process, but overall the operator norms are approximately equal to their initialization, \ie, $\gamma$, since we set $A(0)=I_m$.

\subsection{Hyper-parameter Study of $\gamma$}
\begin{figure*}
	\centering
	\includegraphics[width=.31\textwidth]{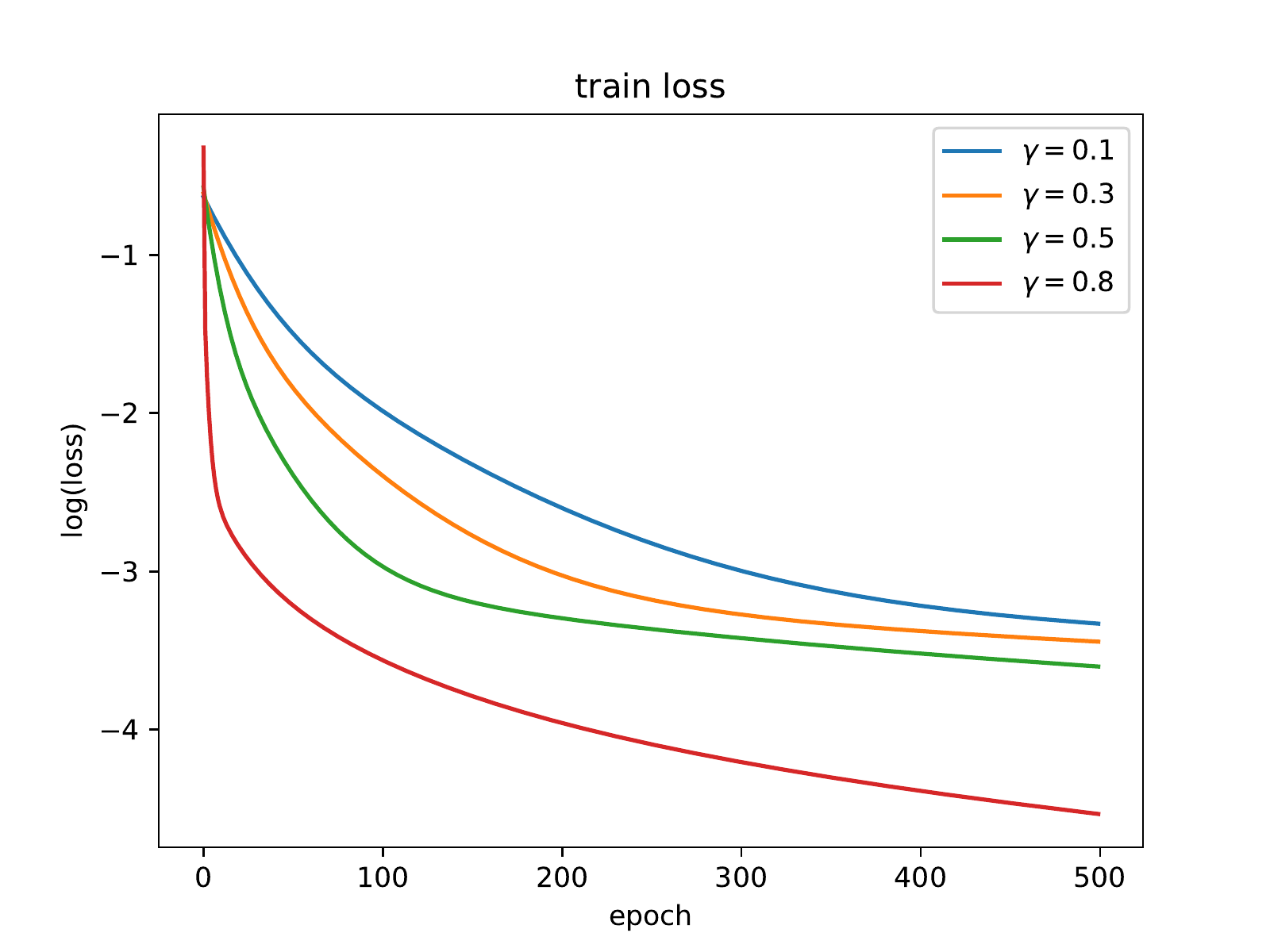}
	\includegraphics[width=.31\textwidth]{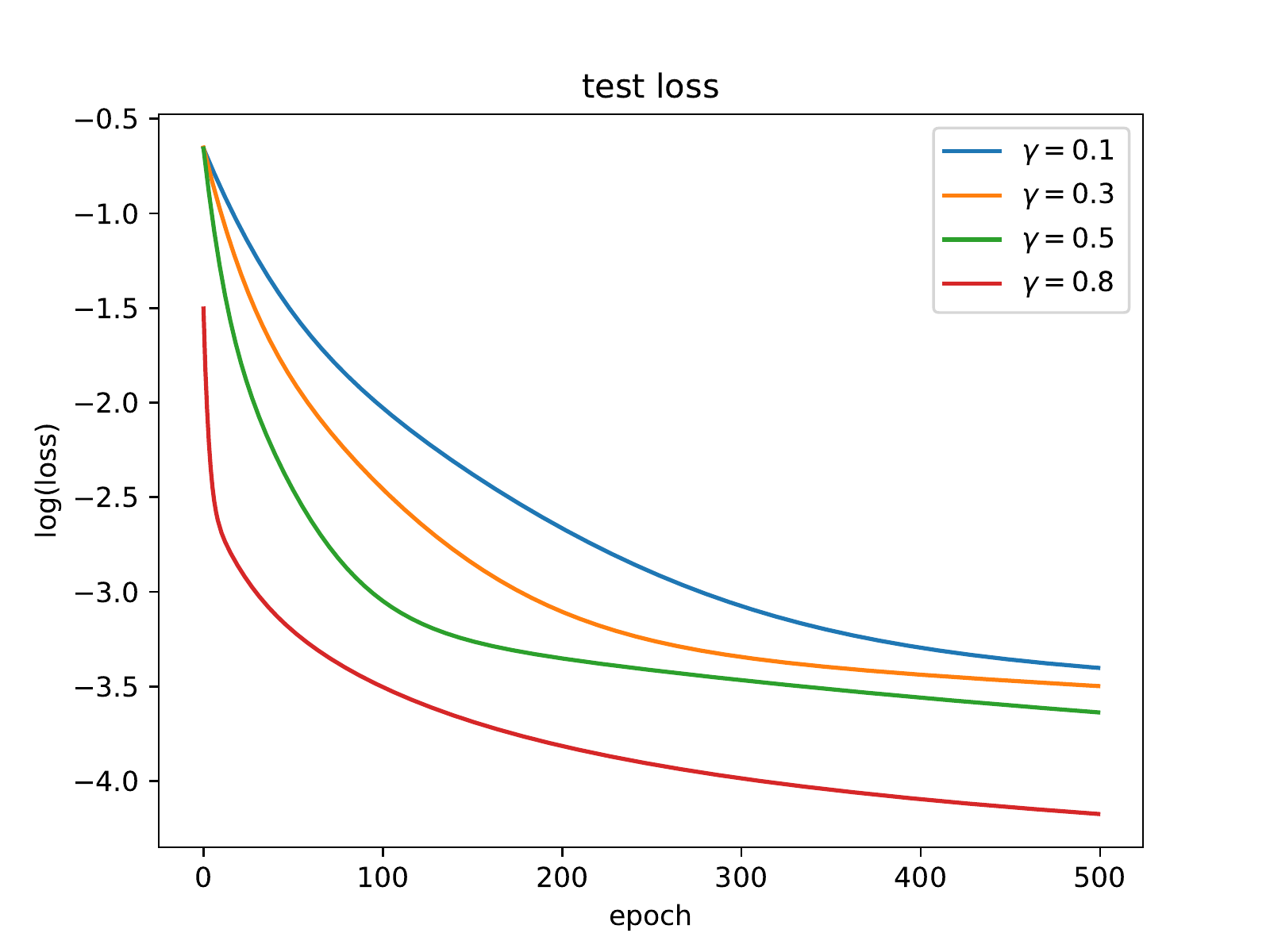}
	\includegraphics[width=.31\textwidth]{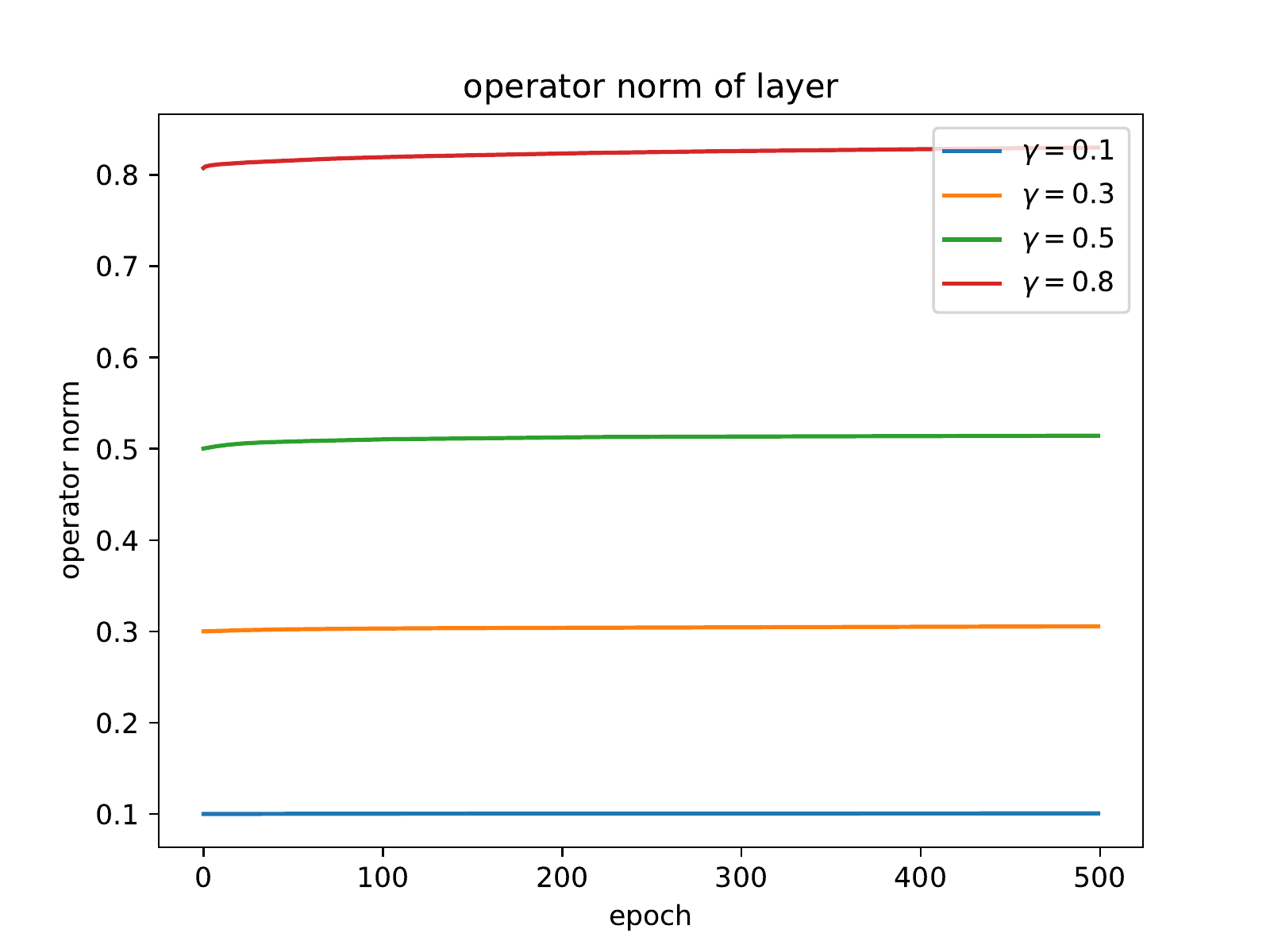}
	\caption{We evaluate the impact of \textbf{choice $\gamma$} on the
		training loss, test loss, and operator norm of the scaled
		matrix $\gamma A(k)$ on the modified dataset of MNIST.}
	\label{fig:fig2}
\end{figure*}

In this work, we introduce the scalar $\gamma$ to solve the
well-posedness of implicit neural networks. Since $\gamma$ is
critical for our theoretical finds, it is worth to conduct a
series of experiments to study the impact
of this new hyper-parameter. Recall that Theorem~\ref{thm:
	gradient descent} shows $m=\Omega(N)$ is enough to ensure the
convergence of gradient descent. With $N=1000$, we set $m=1000$
and keep the rest setups the same as before. We choose the values
of $\gamma$ from $\{0.1, 0.3,0.5,0.8\}$ such that it covers a
reasonable range of values for $\gamma$. Here, we study how the
different choices of hyper-parameter $\gamma$ impact the training
and test performances. We summarize the
results on MNIST in Figure~\ref{fig:fig2}. The rest results are
included in Appendix~\ref{app: exp} to save the space. From
Figure~\ref{fig:fig2}, we can see the lower training loss is
obtained by using relatively larger $\gamma$. Accordingly, the
achieved test loss is also lower when $\gamma$ becomes larger.
These observations show that $\gamma$ has impact on training and
testing performances.

\begin{wraptable}{r}{0.4\textwidth}\vspace{-12pt}
	% \begin{table}
		\centering
		\begin{tabular}{ |c| c c c c|}
			\hline
			$\gamma$ & 0.1 & 0.3 & 0.5 & 0.8 \\ \hline
			iteration $\#$ & 6  & 9 & 15 & 47.5\\ 
			\hline
		\end{tabular}
		\caption{Impact of the \textbf{choice} $\gamma$ on forward iteration to converge on the modified dataset of MNIST.}
		\label{tab:tab1}
		\vspace{-8pt}
	% \end{table}
\end{wraptable}
On the other hand, we also test how the choice of $\gamma$
impacts the number of iteration forward propagation needed to
find the equilibrium point. We stop the forward propagation if
either $\norm{Z^{\ell+1}-Z^{\ell}}\leq 10^{-2}$ or reaches the
max iteration of $100$. Table~\ref{tab:tab1} contains the
averaged number of forward iteration for different $\gamma$
values to find the equilibrium point $Z$ on MNIST. Although Figure~\ref{fig:fig2} shows the lower training
and test losses can be obtained by using relatively larger
$\gamma$, Table~\ref{tab:tab1} indicates larger $\gamma$ results
in more iterations needed in the forward propagation to find the
equilibrium point. Moreover, as the operator norm of $\gamma
A(k)$ becomes larger as growth of epoch, forward propagation
takes even more iterations to converge. This scenario is also
observed in some previous works
\cite{chen2018neural,bai2019deep,bai2021stabilizing}. Thus, there
is probably a trade-off between the convergence speed and
well-posedness of implicit neural networks.

%\vspace{-.1in}
\subsection{Hyper-Parameter Study of Learning Rate}
\begin{figure*}
	\centering
	\includegraphics[width=.31\textwidth]{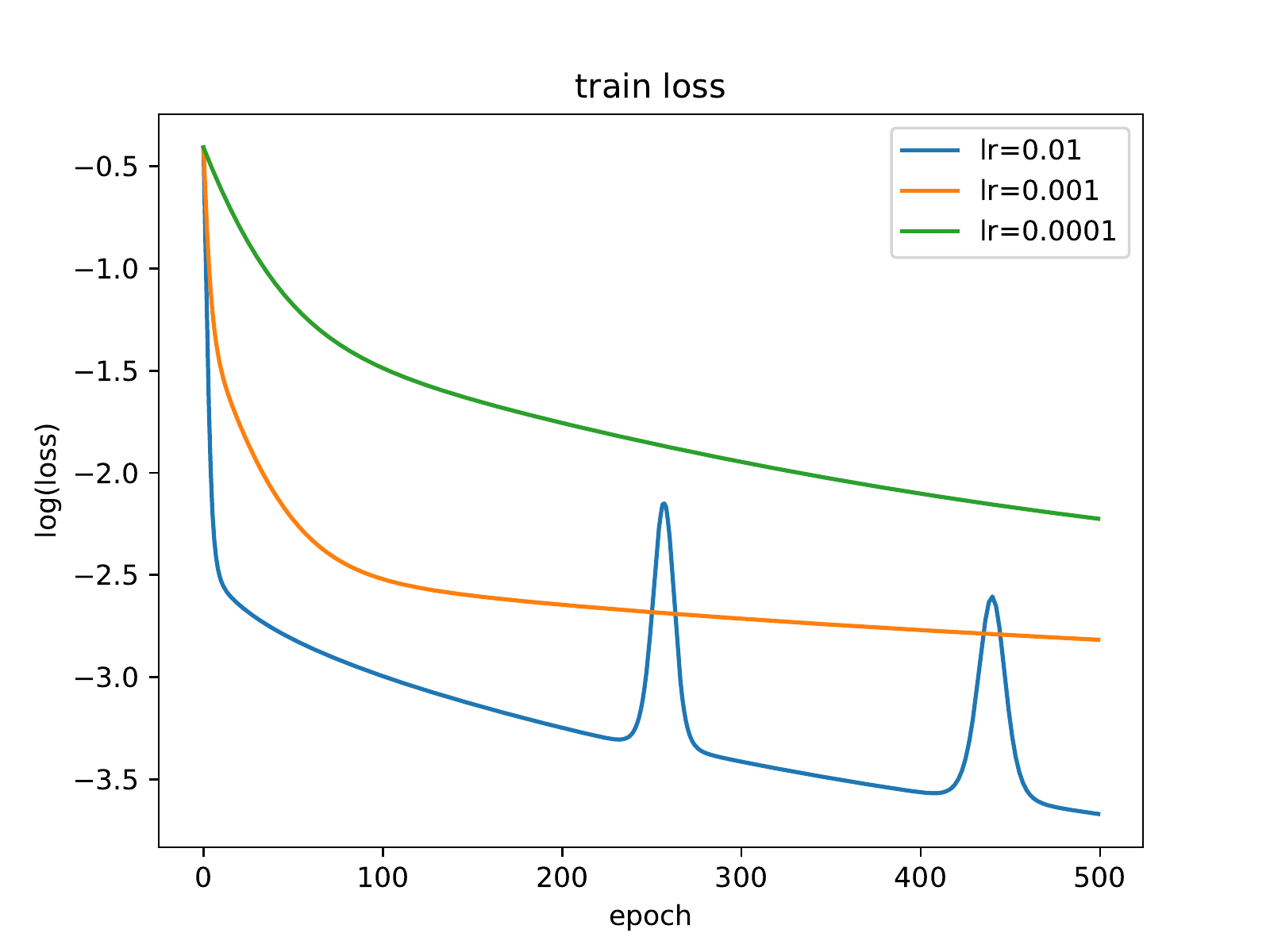}
	\includegraphics[width=.31\textwidth]{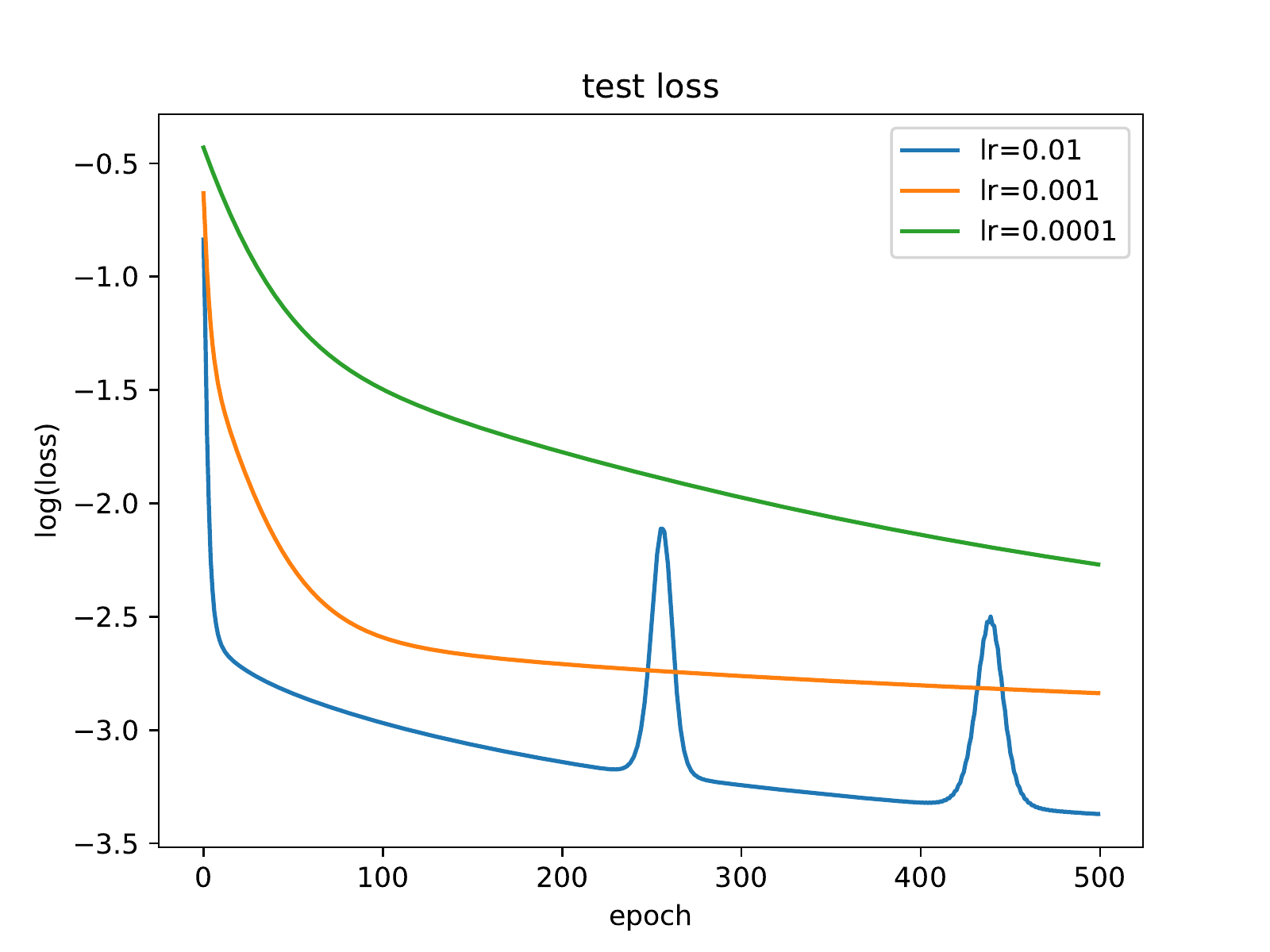}
	\includegraphics[width=.31\textwidth]{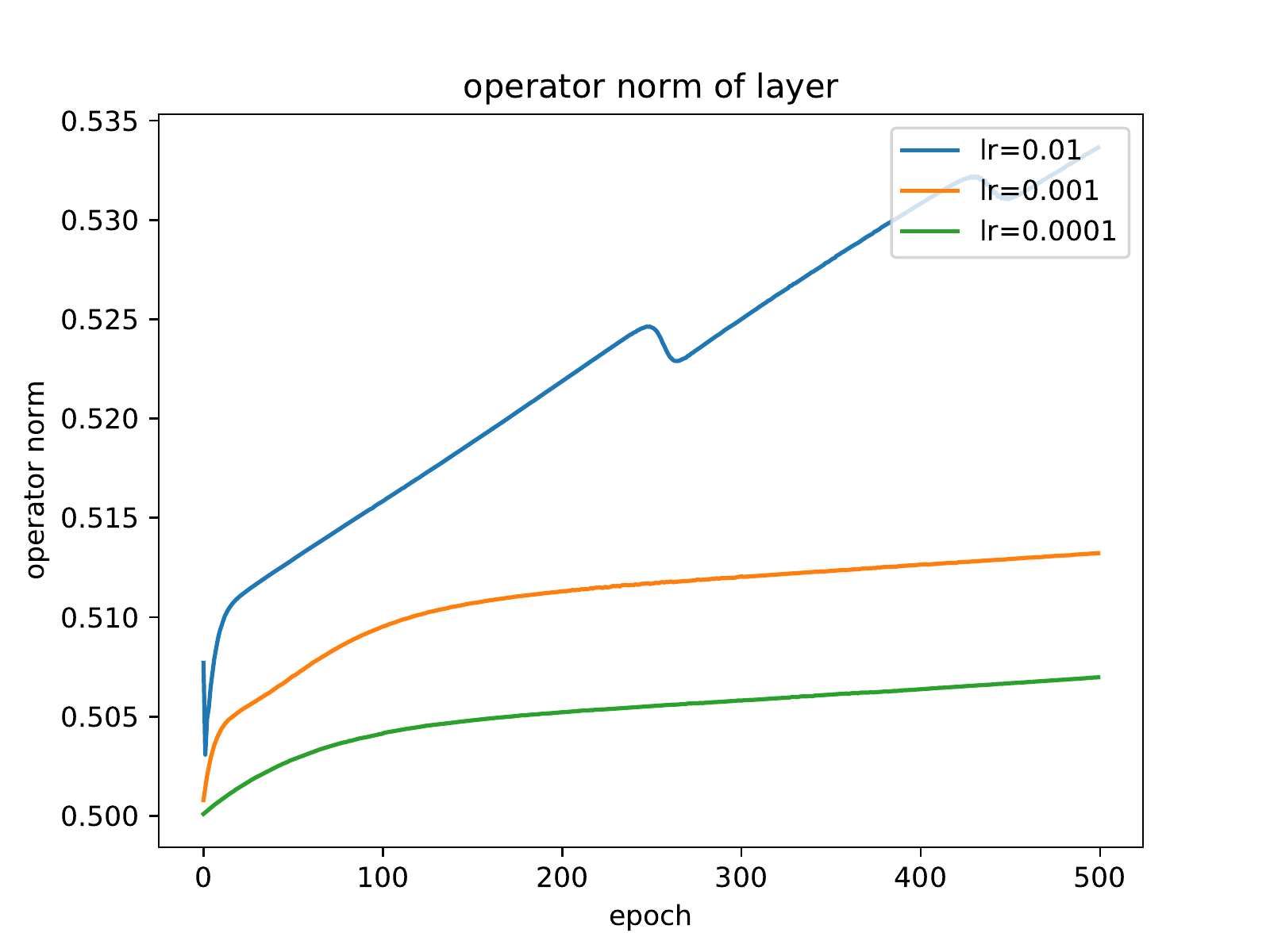}
	\caption{We evaluate the impact of the \textbf{step size $\eta$} on the
		training loss, test loss, and operator norm of the scaled
		matrix $\gamma A(k)$ on the modified dataset of FashionMNIST.}
	\label{fig:fig3}
	% \vspace{-.2in}
\end{figure*}

In general,  the step size or learning rate $\eta$ as a hyper-parameter is significant on both training and test performance. However, the choice of $\eta$ mostly likely is based on the practical experiments. Based on some theoretical analyses, $\eta= \bigo{L^{-1}}$ is one of the widely obtained result, where $L$ is the Lipschitz constant by assuming the gradient of neural network is Lipschitz continuous. Unfortunately, the value of $L$ is generally unknown or relatively large \cite{nguyen2020global}. In Section~\ref{sec:initial}, we show $\eta=\bigo{N^{-1}}$ is enough to guarantee the convergence of the gradient descent, which is much larger than many previous results $\eta=\bigo{N^{-2}}$ \cite{du2018gradient,du2019gradient,gao2021global}, even the implicit neural network could have infinitely many layers. Thus, we also construct a series of numerical experiments to justify the choice of step size $\eta$. With almost the same setup as before, we set the values for step size $\eta$ from $\{10^{-2}, 10^{-3}, 10^{-4}\}$. The corresponding train losses and test losses are illustrated in Figure~\ref{fig:fig3}. The first sub-figure in Figure~\ref{fig:fig3} shows that the objective value does not consistently decrease if we choose the step size $\eta$ larger than the suggested value in Theorem~\ref{thm: gradient descent}. Accordingly, the second sub-figure in Figure~\ref{fig:fig3} shows instability of the test loss for large step size. In addition, the operator norm of $\gamma A(k)$ raises fast to a relatively large value, which probably causes divergence of the forward propagation. Therefore, choosing step size $\eta$ larger than $N^{-1}$ is probably causing the divergence of the forward propagation and the entire training process. This experimentally indicates the result of the step size $\eta$ in Theorem~\ref{thm: gradient descent} is tight.

%\vspace{-.1in}
\section{Conclusion}
This paper studies the convergence problem of first-order methods such as gradient descent for ReLU implicit networks with infinitely many layers. Specifically, we provide sufficient conditions under which both gradient flow and gradient descent converge to a global minimum at a linear rate. Moreover, we show that these sufficient conditions can be indeed satisfied by some initialization as long as the width $m$ is linear in the sample size $N$, \ie, $m=\Omega(N)$, even when the implicit networks have infinitely many layers and are activated by ReLU. Moreover, we also show that popular random initializations satisfy sufficient conditions under a stronger condition where the width is quadratic of the sample size, \ie, $m=\Omega(N^2)$.

\clearpage
\bibliography{example_paper}
\bibliographystyle{plain}
%%%%%%%%%%%%%%%%%%%%%%%%%%%%%%%%%%%%%%%%%%%%%%%%%%%%%%%%%%%%

\clearpage
\appendix

\section{Appendix}
\subsection{Proof of Lemma~\ref{lemma: forward}}\label{app: forward}
For each $\ell$, we have
\begin{align*}
	\Norm{Z^{\ell+1}-Z^{\ell}}_F
	=&\Norm{\sigma(\gamma Z^{\ell}A + \Phi) - \sigma(\gamma Z^{\ell-1}A + \Phi)}_F\\
	\leq & \gamma \norm{Z^{\ell}A - Z^{\ell-1}A}_F\quad\text{by the Lipschitz continuity of $\sigma$}\\
	\leq & \gamma\norm{A} \norm{Z^{\ell} - Z^{\ell-1}}_F\\
	<& \gamma_0 \norm{Z^{\ell} - Z^{\ell-1}}_F.
\end{align*}
Repeating the above argument $\ell$ times yields
\begin{align*}
	\Norm{Z^{\ell+1}-Z^{\ell}}_F\leq \gamma_0^{\ell}\Norm{Z^{1}-Z^0}_F
	=\gamma_0^{\ell}\Norm{Z^{1}}_F
	=\gamma_0^{\ell}\Norm{\Phi}_F,
\end{align*}
where the second last equality is due to $Z^{0}=0$. Therefore, for all $p, q$ with $p\leq q$, we have
\begin{align*}
	\norm{Z^{p}-Z^{q}}_F
	\leq& \norm{Z^{p}-Z^{p+1}}_F + \cdots + \norm{Z^{q-1}-Z^{q}}_F\\
	\leq & \gamma_0^{p}\norm{\Phi}_F + \cdots + \gamma_0^{q-1}\Norm{\Phi}_F\\
	\leq & \gamma_0^{p}\norm{\Phi}_F\left(1+\gamma_0+\gamma_0^2+\cdots\right)\\
	=&\frac{\gamma_0^{p}}{1-\gamma_0}\norm{\Phi}.
\end{align*} 
Let $p,q\rightarrow \infty$, then $\norm{Z^{p}-Z^{q}}_F\rightarrow 0$. Thus, $\{Z^{\ell}\}_{\ell=0}^{\infty}$ is a Cauchy sequence in $\reals^{N\times m}$. By the completeness of $\reals^{N\times m}$, $\{Z^{\ell}\}_{\ell=0}^{\infty}$ converges to the unique limit $Z$, which is the equilibrium point. Now, let $p=0$ and $q=\ell$, then we have
\begin{align*}
	\norm{Z^{\ell}}_F\leq \frac{1}{1-\gamma_0}\Norm{\Phi}_F.
\end{align*}
Moreover, let $\ell\rightarrow \infty$, we obtain $\norm{Z}_F\leq \frac{1}{1-\gamma_0}\Norm{\Phi}_F$.

\subsection{Proof of Lemma~\ref{lemma: gradient}}\label{app: gradient}
Since the equilibrium point $Z$ is the root of the function $f$ defined by
\begin{align*}
	f(Z,A,W):=Z-\sigma(\gamma ZA+\Phi),
\end{align*}
then the differential of $f$ is given by
\begin{align*}
	df =& dZ -d \sigma(\gamma  ZA+ \Phi)\\
	=&dZ - \sigma^{\prime}(U)\odot d(\gamma  ZA+\Phi)\\
	=&dZ 
	- \sigma^{\prime}(U)\odot \gamma  (dZ)A
	-\sigma^{\prime}(U) \odot \gamma  ZdA
	-\sigma^{\prime}(U)\odot d\Phi,
\end{align*}
where $U:=\gamma  ZA+\Phi$. 

Taking vectorization on both sides yields
\begin{align*}
	\vect{df}=&\vect{dZ 
		- \sigma^{\prime}(U)\odot \gamma  (dZ)A
		-\sigma^{\prime}(U) \odot \gamma  ZdA
		-\sigma^{\prime}(U)\odot d\Phi}\\
	=&\vect{dZ}
	-\gamma D \vect{dZ A}
	-\gamma  D\vect{ZdA}
	-D \vect{d\Phi}\\
	=&\vect{dZ}
	-\gamma D \left(A^T\otimes I_N\right)\vect{dZ}
	-\gamma  D \left(I_m\otimes Z\right)\vect{dA}
	-D \vect{d\Phi},
\end{align*}
where $D:=\diag[\vect{\sigma^{\prime}(U)}]$. Thus, we obtain
\begin{align*}
	\frac{\partial f}{\partial Z}
	=&\left[I_{Nm}-\gamma D(A^T\otimes I_N)\right]^T\\
	\frac{\partial f}{\partial A}
	=&-\gamma \left[D (I_m \otimes Z) \right]^T\\
	\frac{\partial f}{\partial \Phi}
	=&- D^T.
\end{align*}

By reverse triangle inequality, we have
\begin{align}
	\Norm{I_{Nm}-\gamma D(A^T\otimes I_N)}
	\geq 1-\gamma \norm{D(A^T\otimes I_N)}
	\geq 1-\gamma\norm{A} > 1-\gamma_0 > 0,
\end{align}
where we use $\norm{A}\leq M$ and $\gamma_0 = \gamma M < 1$. Hence, the matrix $Q:=\left[I_{Nm}-\gamma D(A^T\otimes I_N)\right]$ is invertible.

Since the fixed point $Z$ is the root of $f$ and the matrix $Q$ is invertible, by using the \textit{implicit function theorem}, we have
\begin{align*}
	\frac{\partial Z}{\partial A}\frac{\partial f}{\partial Z}
	+\frac{\partial f}{\partial A} = 0,
\end{align*}
which implies
\begin{align*}
	\frac{\partial Z}{\partial A}
	=-\left(\frac{\partial f}{\partial A}\right)
	\left( \frac{\partial f}{\partial Z}\right)^{-1}
	=\gamma \left[D (I_m \otimes Z) \right]^T Q^{-T}.
\end{align*}

Similarly, we have
\begin{align*}
	\frac{\partial Z}{\partial \Phi} = -\left(\frac{\partial f}{\partial \Phi}\right)\left(\frac{\partial f}{\partial Z}\right)^{-1}
	=D^{T}Q^{-T}.
\end{align*}

Moreover, the definition of $\Phi$ implies that
\begin{align*}
	d\Phi = d\sigma(XW)
	= \sigma^{\prime}(XW) \odot X dW.
\end{align*}
Taking vectorization on both sides yields
\begin{align*}
	\vect{d\Phi}
	=E\vect{XdW}
	=E \left(I_m\otimes X\right) \vect{dW}.
\end{align*}
where $E:=\diag\left[\vect{\sigma^{\prime}(XW)}\right]$. Therefore, we have
\begin{align*}
	\frac{\partial \Phi}{\partial W}
	=\left[E \left(I_m\otimes X\right) \right]^T.
\end{align*}

By using the chain rule, we have
\begin{align*}
	\frac{\partial Z}{\partial W}
	=\left(\frac{\partial \Phi}{\partial W}\right)\left(\frac{\partial Z}{\partial \Phi}\right)
	=\left[DE \left(I_m\otimes X\right) \right]^TQ^{-T}.
\end{align*}

Next, we have $\hat{y}=Zb$, so that
\begin{align*}
	d\hat{y} = dZb = (dZ)b + Zdb.
\end{align*}
Taking vectorization on both sides yields
\begin{align*}
	\vect{d\hat{y}} =\left(b^T\otimes I_N\right)\vect{dZ} + Z \vect{db},
\end{align*}
so that we have
\begin{align}
	\frac{\partial \hat{y}}{\partial Z} =& \left(b^T\otimes I_N\right)^T\\
	\frac{\partial \hat{y}}{\partial b}=&Z^T.
\end{align}
By using the chain rule, we obtain
\begin{align}
	\frac{\partial \hat{y}}{\partial A}
	=&\frac{\partial Z}{\partial A}\frac{\partial \hat{y}}{\partial Z}
	=\gamma \left[D (I_m \otimes Z) \right]^T
	Q^{-T}\left(b^T\otimes I_N\right)^T\\
	\frac{\partial \hat{y}}{\partial W}
	=&\frac{\partial Z}{\partial W}\frac{\partial \hat{y}}{\partial Z}
	=\left[DE \left(I_m\otimes X\right) \right]^T
	Q^{-T}\left(b^T\otimes I_N\right)^T.
\end{align}
Note that
\begin{align*}
	dL = (\hat{y}-y)^T d\hat{y}
\end{align*}
so that
\begin{align*}
	\frac{\partial L}{\partial \hat{y}} = \hat{y}-y.
\end{align*}
Therefore, the partial derivative of $L$ with respect to $W$, $A$, and $b$  are given by
\begin{align*}
	\frac{\partial L}{\partial W} 
	=& \frac{\partial \hat{y}}{\partial W}\frac{\partial L}{\partial \hat{y}} 
	=\left[DE \left(I_m\otimes X\right) \right]^T
	Q^{-T}\left(b^T\otimes I_N\right)^T(\hat{y}-y)\\
	\frac{\partial L}{\partial A} 
	=& \frac{\partial \hat{y}}{\partial A}\frac{\partial L}{\partial \hat{y}} 
	=\gamma \left[D (I_m \otimes Z) \right]^T
	Q^{-T}\left(b^T\otimes I_N\right)^T(\hat{y}-y)\\
	\frac{\partial L}{\partial b} 
	=& \frac{\partial \hat{y}}{\partial b}\frac{\partial L}{\partial \hat{y}} 
	=Z^T(\hat{y}-y).
\end{align*}

\subsection{Proof of Lemma~\ref{lemma: dynamics prediction}}\label{app: dynamics}
The gradient flows are given by
\begin{align*}
	\frac{d \vect{W}}{dt} = -\frac{\partial L}{\partial W},
	\quad
	\frac{d \vect{A}}{dt} = -\frac{\partial L}{\partial A},
	\quad
	\frac{d b}{dt} = -\frac{\partial L}{\partial b}.
\end{align*}
By using the chain rule, we obtain the dynamics system of the equilibrium point as follows
\begin{align*}
	\frac{d \vect{Z}}{dt}
	=&\left(\frac{\partial Z}{\partial A}\right)^T\left(\frac{d\vect{A}}{dt}\right)  +\left(\frac{\partial Z}{\partial W}\right)^{T}\left(\frac{d\vect{W}}{dt}\right)\\
	=&-\left(\frac{\partial Z}{\partial A}\right)^T\left(\frac{\partial L}{\partial A}\right) -\left(\frac{\partial Z}{\partial W}\right)^{T}\left(\frac{\partial L}{\partial W}\right)\\
	=&-\left[\left(\frac{\partial Z}{\partial A}\right)^T\left(\frac{\partial Z}{\partial A}\right)
	+\left(\frac{\partial Z}{\partial W}\right)^{T}\left(\frac{\partial Z}{\partial W}\right)
	\right]\left(\frac{\partial \hat{y}}{\partial Z}\right)\left(\frac{\partial L}{\partial \hat{y}}\right).
\end{align*}
Accordingly, the dynamics of the prediction $\hat{y}$ is given by
\begin{align*}
	\frac{d\hat{y}}{dt}
	=&\left(\frac{\partial \hat{y}}{\partial b}\right)^T\left(\frac{d b}{dt}\right)
	+\left(\frac{\partial \hat{y}}{\partial Z}\right)^T\left(\frac{d \vect{Z}}{d t}\right)\\
	=&-\left\{\left(\frac{\partial \hat{y}}{\partial b}\right)^{T}\left(\frac{\partial \hat{y}}{\partial b}\right)
	+\left(\frac{\partial \hat{y}}{\partial A}\right)^T\left(\frac{\partial \hat{y}}{\partial A}\right)
	+\left(\frac{\partial \hat{y}}{\partial W}\right)^{T}
	\left(\frac{\partial \hat{y}}{\partial W}\right)\right\}\left(\frac{\partial L}{\partial \hat{y}}\right)\\
	\triangleq&-H(t)(\hat{y}(t)-y),
\end{align*}
where
\begin{align*}
	&\left(\frac{\partial \hat{y}}{\partial b}\right)^{T}\left(\frac{\partial \hat{y}}{\partial b}\right)
	=Z(t)Z(t)^T\\
	&\left(\frac{\partial \hat{y}}{\partial A}\right)^T\left(\frac{\partial \hat{y}}{\partial A}\right)
	=\gamma^2\left\{ \left[D (I_m \otimes Z) \right]^T
	Q^{-T}\left(b^T\otimes I_N\right)^T\right\}^T
	\left\{ \left[D (I_m \otimes Z) \right]^T
	Q^{-T}\left(b^T\otimes I_N\right)^T\right\}\\
	&\left(\frac{\partial \hat{y}}{\partial W}\right)^{T}\left(\frac{\partial \hat{y}}{\partial W}\right)
	=\left\{\left[DE \left(I_m\otimes X\right) \right]^T
	Q^{-T}\left(b^T\otimes I_N\right)^T\right\}^T
	\left\{\left[DE \left(I_m\otimes X\right) \right]^T
	Q^{-T}\left(b^T\otimes I_N\right)^T\right\}.
\end{align*}

\subsection{Proof of Theorem~\ref{thm: gradient flow}}\label{app: gradient flow}
\begin{proof}
	We make the inductive hypothesis as follows for all $0\leq s\leq t$
	\begin{enumerate}[label=(\roman{*})]
		\item $\norm{W(s)}\leq \lambda_1$, 
		$\norm{A(s)}\leq \lambda_2$, 
		$\norm{b(s)}\leq \lambda_3$,
		\item $\sigma_{\min}(Z(s))\geq \alpha_0/2$,
		\item $\norm{\hat{y}(s)-y}^2\leq \exp\{-(\alpha_0^2/2)s\}\norm{\hat{y}(0)-y}^2$.
	\end{enumerate}
	For any $0\leq s\leq t$, we have 
	\begin{align}
		\norm{\Phi(s)}_F 
		= \norm{\sigma(XW(s))}_F
		\leq \norm{XW(s)}_F
		\leq \norm{X}_F\norm{W(s)}
		\leq \lambda_1\norm{X}_F,\label{eq: Phi}
	\end{align}
	where the last inequality follows from the inductive hypothesis (i). Then Lemma~\ref{lemma: forward} implies that
	\begin{align}
		\Norm{Z(s)}_F\leq \frac{1}{1-\gamma_0}\norm{\Phi(s)}_F\leq \frac{\lambda_1}{1-\gamma_0}\Norm{X}_F.
		\label{eq: Z(s)}
	\end{align}
	
	Then we can bound the partial derivative of $L$ with respect to $b$ as follows
	\begin{align}
		\norm{(\partial L/\partial b)(s)}
		=&\norm{Z(s)^T(\hat{y}(s)-y)}\nonumber\\
		\leq& \norm{Z(s)}\norm{\hat{y}(s)-y}\nonumber\\
		\leq&  \frac{\lambda_1}{1-\gamma_0} \norm{X}_F\norm{\hat{y}(s)-y},\quad\text{by Eq.\eqref{eq: Z(s)}}\\
		\leq& \frac{\lambda_1}{1-\gamma_0}\norm{X}_F\exp\left\{-(\alpha_0^2/4)s\right\}\norm{\hat{y}(0)-y},
	\end{align}
	where the last inequality is due to the inductive hypothesis (iii). Then we can bound the difference between $b(t)$ and $b(0)$ as follows
	\begin{align*}
		\norm{b(t)-b(0)}\leq \int_{0}^{t}\norm{(\partial L/\partial b)(s)} ds
		\leq\frac{4\lambda_1}{\alpha_0^2(1-\gamma_0)}\norm{X}_F \norm{\hat{y}(0)-y}\leq C_3,
	\end{align*}
	where the last inequality follows from the initial assumptions \eqref{eq:GF1}-\eqref{eq:GF2}. By Weyl’s inequality, we obtain
	\begin{align*}
		\norm{b(t)}
		\leq \norm{b(t)-b(0)} + \norm{b(0)}\leq  \lambda_3.
	\end{align*}
	
	With the similar argument, we can bound the partial derivative of $L$ with respect to $W$
	\begin{align*}
		\Norm{(\partial L/\partial W)(s)}
		=&\Norm{\left[D(s)E(s) \left(I_m\otimes X\right) \right]^TQ(s)^{-T}\left(b(s)^T\otimes I_N\right)^T(\hat{y}(s)-y)}\\
		\leq &\norm{D(s)}\norm{E(s)}\norm{X}\norm{Q(s)^{-1}}\norm{b(s)}\norm{\hat{y}(s)-y}\\
		\leq &\norm{X}_F\cdot \frac{1}{1-\gamma_0}\cdot \lambda_3\cdot \norm{\hat{y}(s)-y}\\
		\leq &\frac{\lambda_3}{1-\gamma_0}\norm{X}_F \exp\left\{-(\alpha_0^2/4)s\right\}\norm{\hat{y}(0)-y},
	\end{align*}
	and the difference between $W(t)$ and $W(0)$
	\begin{align*}
		\norm{W(t)-W(0)}
		\leq \norm{W(t)-W(0)}_F
		\leq \int_0^t\Norm{(\partial L/\partial W)(s)} ds
		\leq\frac{4\lambda_3}{\alpha_0^2(1-\gamma_0)}\norm{X}_F \norm{\hat{y}(0)-y}\leq C_1,
	\end{align*}
	which further implies
	\begin{align*}
		\norm{W(t)}\leq \norm{W(t)-W(0)} + \norm{W(0)}\leq  \lambda_1.
	\end{align*}
	
	Similarly, we can bound the partial derivative of $L$ with respect to $A$ as follows
	\begin{align*}
		\Norm{(\partial L/\partial A)(s)}
		=&\Norm{\gamma \left[D(s) (I_m \otimes Z(s)) \right]^T
			Q(s)^{-T}\left(b(s)^T\otimes I_N\right)^T(\hat{y}(s)-y)}\\
		\leq &\gamma \norm{D(s)}\norm{Z(s)}\norm{Q(s)^{-1}}\norm{b(s)}\norm{\hat{y}(s)-y}\\
		\leq &\gamma \cdot \frac{\lambda_1}{1-\gamma_0}\norm{X}_F\cdot\frac{1}{1-\gamma_0}\cdot\lambda_3\cdot\norm{\hat{y}(s)-y},\quad\text{By $\eqref{eq: Z(s)}$ and inductive hypothesis (iii)}\\
		= &\frac{\gamma_0 }{(1-\gamma_0)^2}\cdot \frac{\lambda_1 \lambda_3}{\lambda_2}\cdot \norm{X}_F\cdot \norm{\hat{y}(s)-y},\quad \text{By }\gamma_0=\gamma\lambda_2\\
		\leq &\frac{\gamma_0}{(1-\gamma_0)^2}\frac{\lambda_1\lambda_3}{\lambda_2}\norm{X}_F\exp\{-(\alpha_0^2/4)s\}\norm{\hat{y}(0)-y},
	\end{align*}
	and the difference between $A(t)$ and $A(0)$ as follows
	\begin{align*}
		\norm{A(t)-A(0)}
		\leq& \norm{A(t)-A(0)}_F\\
		\leq& \int_0^t\Norm{(\partial L/\partial A)(s)} ds\\
		\leq &\frac{4\gamma_0}{\alpha_0^2(1-\gamma_0)^2}\frac{\lambda_1\lambda_3}{\lambda_2}\norm{X}_F \norm{\hat{y}(0)-y}\\
		\leq& C_2.
	\end{align*}
	Therefore, we obtain
	\begin{align*}
		\norm{A(t)}\leq \norm{A(t)-A(0)} + \norm{A(0)} \leq  \lambda_2.
	\end{align*}
	
	It follows from the inductive hypothesis and analysis above that $\norm{A(s)}\leq \lambda_2$ for all $0\leq s\leq t$. Since $\gamma_0=\lambda_2\gamma<1$, it follows from Lemma~\ref{lemma: forward} that the unique equilibrium point $Z(s)$ always exists for all $0\leq s\leq t$. Then we can derive the dynamics of the equilibrium point as follows
	\begin{align*}
		\frac{d\vect{Z}}{dt}
		=&\left(\frac{\partial Z}{\partial A}\right)^T\left(\frac{d\vect{A}}{dt}\right)
		+\left(\frac{\partial Z}{\partial W}\right)^T\left(\frac{d\vect{W}}{dt}\right)\\
		=&\left(\frac{\partial Z}{\partial A}\right)^T\left(-\frac{\partial L}{\partial A}\right)
		+\left(\frac{\partial Z}{\partial W}\right)^T\left(-\frac{\partial L}{\partial W}\right)\\
		=&-\left[\left(\frac{\partial Z}{\partial A}\right)^T\left(\frac{\partial Z}{\partial A}\right)+\left(\frac{\partial Z}{\partial W}\right)^T\left(\frac{\partial Z}{\partial W}\right)\right]\left(\frac{\partial \hat{y}}{\partial Z}\right)\left(\frac{\partial L}{\partial \hat{y}}\right),
	\end{align*}
	Note that for each $0\leq s\leq t$, we have
	\begin{align*}
		\norm{(\partial Z/\partial A)(s)}
		=&\Norm{\gamma \left[D(s) (I_m \otimes Z(s)) \right]^T Q(s)^{-T}}\\
		\leq & \gamma\cdot \frac{\lambda_1}{(1-\gamma_0)}\cdot \norm{X}_F\cdot \frac{1}{(1-\gamma_0)}\\
		= &  \frac{\gamma_0}{(1-\gamma_0)^2} \frac{\lambda_1}{\lambda_2}\norm{X}_F,
	\end{align*}
	and
	\begin{align*}
		\norm{(\partial Z/\partial W)(s)}
		=&\Norm{\left[D(s)E(s) \left(I_m\otimes X\right) \right]^TQ(s)^{-T}}\\
		\leq & \norm{X}_F\norm{Q(s)^{-1}}\\
		\leq & \frac{1}{(1-\gamma_0)}\norm{X}_F
	\end{align*}
	and
	\begin{align*}
		\Norm{\left(\frac{\partial \hat{y}}{\partial Z}\right)\left(\frac{\partial L}{\partial \hat{y}}\right)}
		\leq\Norm{b(s)}\norm{\hat{y}(s)-y}
		\leq \lambda_3 \exp\left\{-(\alpha_0^2/4)s\right\}\norm{\hat{y}(0)-y}.
	\end{align*}
	Therefore, we obtain
	\begin{align*}
		\Norm{\frac{d\vect{Z}}{ds}}
		\leq &\Big(\norm{(\partial Z/\partial A)(s)}^2 + \norm{(\partial Z/\partial W)(s)}^2\Big)
		\Norm{\left(\frac{\partial \hat{y}}{\partial Z}\right)\left(\frac{\partial L}{\partial \hat{y}}\right)}\\
		\leq &\left[\frac{\gamma_0^2}{(1-\gamma_0)^2}\frac{\lambda_1^2}{\lambda_2^2}+1\right]
		\frac{\lambda_3}{(1-\gamma_0)^2}\norm{X}_F^2 \exp\left\{-(\alpha_0^2/4)s\right\}\norm{\hat{y}(0)-y},
	\end{align*}
	and
	\begin{align*}
		\Norm{Z(t)-Z(0)}
		\leq \Norm{Z(t)-Z(0)}_F
		\leq &\int_0^t \Norm{\frac{d\vect{Z}}{ds}} ds\\
		\leq & \left[\frac{\gamma_0^2}{(1-\gamma_0)^2}\frac{\lambda_1^2}{\lambda_2^2}+1\right]\frac{4\lambda_3}{\alpha_0^2(1-\gamma_0)^2}
		\norm{X}^2 \norm{\hat{y}(0)-y}\\
		\leq &\alpha_0/2,
	\end{align*}
	where the last inequality follows from the initial assumption.
	
	Since $\sigma_{\min}{Z(0)}=\alpha_0$, Weyl’s inequality implies that
	\begin{align*}
		\sigma_{\min}\left\{Z(t)\right\}\geq \sigma_{\min}\left\{Z(0)\right\}-\norm{Z(0)-Z(t)}\geq \alpha_0/2.
	\end{align*}
	As a result, we have 
	\begin{align*}
		\lambda_{\min}\left\{H(t)\right\}\geq \lambda_{\min}\left\{Z(t)Z(t)^T\right\}
		=\sigma_{\min}\{Z(t)\}^2\geq \alpha_0^2/4,
	\end{align*}
	and so that 
	\begin{align*}
		\frac{d}{dt}L(\theta(t))=&
		\frac{d}{dt}\left(\frac{1}{2}\norm{\hat{y}(t)-y}^2\right)\\
		=&-(\hat{y}(t)-y)^T\frac{d\hat{y}(t)}{dt}\\
		=&-(\hat{y}(t)-y)^TH(t)(\hat{y}(t)-y)\\
		\leq& -(\alpha_0^2/4)\norm{\hat{y}(t)-y}^2\\
		=&-(\alpha_0^2/2)L(\theta(t)).
	\end{align*}
	Solving the above ordinary differential equation yields
	\begin{align*}
		L(\theta(t))\leq \exp\left\{-(\alpha_0^2/2)t\right\}L(\theta(0)).
	\end{align*}
	This completes the proof.
\end{proof}

\subsection{Proof of Lemma~\ref{lemma: Z_a-Z_b}}\label{app: Z_a-Z_b}
Since $\norm{A_a} < \gamma^{-1}$ and $\norm{A_b} < \gamma^{-1}$, Lemma~\ref{lemma: forward} implies the corresponding equilibrium points $Z_a$ and $Z_b$ are uniquely determined. Note that $Z_i=\lim_{\ell\rightarrow \infty} Z_i^{\ell}$ for $i\in \{a,b\}$. Then for any $\ell\geq 1$, we have
\begin{align*}
	\norm{Z_a^{\ell+1} - Z_b^{\ell+1}}_F
	=&\norm{\sigma(\gamma Z_a^{\ell}A_a + \Phi_a) - \sigma(\gamma Z_b^{\ell}A_b + \Phi_b)}_F\\
	\leq&\norm{\gamma Z_a^{\ell}A_a + \Phi_a-\gamma Z_b^{\ell}A_b - \Phi_b}_F\\
	\leq &\gamma\norm{Z_a^{\ell}A_a -Z_a^{\ell}A_b}_F + \gamma\norm{Z_a^{\ell}A_b-Z_b^{\ell}A_b}_F + \norm{\Phi_a-\Phi_b}_F\\
	\leq &\gamma \norm{Z_a^{\ell}}\norm{A_a-A_b}_F + \gamma\norm{Z_a^{\ell}-Z_b^{\ell}}_F\norm{A_b} + \norm{\Phi_a - \Phi_b}_F.
\end{align*} 
In the rest of the proof, we will bound each term above. Lemma~\ref{lemma: forward} implies that
\begin{align*}
	\gamma \norm{Z_a^{\ell}}\norm{A_a-A_b}_F
	\leq \gamma\cdot \frac{1}{1-\gamma_0}\norm{\Phi_a}_F\cdot \norm{A_a-A_b}
	\leq \frac{\gamma_0}{1-\gamma_0}\frac{\lambda_1}{\lambda_2}\norm{X}_F\norm{A_a-A_b}.
\end{align*}
Since $\norm{A_b}\leq \lambda_2$, the second term has the following inequality
\begin{align*}
	\gamma\norm{Z_a^{\ell}-Z_b^{\ell}}\norm{A_b}_F\leq \gamma_0\norm{Z_a^{\ell} - Z_b^{\ell}}.
\end{align*}
By using the Lipschitz continuity of $\sigma(\cdot)$, we have
\begin{align*}
	\norm{\Phi_a - \Phi_b}_F=\norm{\sigma(XW_a)-\sigma(XW_b)}_F\leq \norm{X}_F\norm{W_a-W_b}.
\end{align*}
Combining all results together yields
\begin{align*}
	\norm{Z_a^{\ell+1} - Z_b^{\ell+1}}_F\leq \gamma_0\norm{Z_a^{\ell} - Z_b^{\ell}}_F 
	+ \left( \frac{\gamma_0}{1-\gamma_0}\frac{\lambda_1}{\lambda_2}\norm{X}_F\norm{A_a-A_b} + \norm{X}_F\norm{W_a-W_b}\right)
\end{align*}
Let $C:=\frac{\gamma_0}{1-\gamma_0}\frac{\lambda_1}{\lambda_2}\norm{X}_F\norm{A_a-A_b} + \norm{X}_F\norm{W_a-W_b}$ and apply the same argument $\ell$ times yields
\begin{align*}
	\norm{Z_a^{\ell+1} - Z_b^{\ell+1}}_F\leq \gamma_0^{\ell}\norm{Z_a^{1} - Z_b^{1}}_F
	+\left(1+\gamma_0+\cdots+\gamma_0^{\ell-1}\right)C.
\end{align*}
Let $\ell\rightarrow \infty$. The continuity of the operator norm implies that
\begin{align*}
	\norm{Z_a-Z_b}_F\leq \frac{1}{1-\gamma_0}C.
\end{align*}

\subsection{Proof of Theorem~\ref{thm: gradient descent}}\label{app: gradient descent}
\begin{proof}
	We make the inductive hypothesis as follows for all $0\leq s\leq k$
	\begin{enumerate}[label=(\roman{*})]
		\item $\norm{W(s)}\leq \lambda_1$, 
		$\norm{A(s)}\leq \lambda_2$, 
		$\norm{b(s)}\leq \lambda_3$,
		\item $\sigma_{\min}(Z(s))\geq \alpha_0/2$,
		\item $L(\theta(s))\leq (1-\eta\alpha_0^2/4)^sL(\theta(0))$.
	\end{enumerate}
	For any $0\leq s\leq k$, the inductive hypothesis implies that
	\begin{align}
		\norm{\Phi(s)}=\norm{\sigma(XW(s))}\leq \lambda_1\norm{X}.\label{eq1: Phi}
	\end{align}
	By using Eq. \eqref{eq: Z(s)}, the partial derivative of $L$ with respect to $b$ can be bounded as follows
	\begin{align}
		\norm{(\partial L/\partial b)(s)}
		=&\norm{Z(s)^T(\hat{y}(s)-y)}\nonumber\\
		\leq& \norm{Z(s)}\norm{\hat{y}(s)-y}\nonumber\\
		\leq & \frac{\lambda_1}{1-\gamma_0}\norm{X}_F\norm{\hat{y}(s)-y},\\
		\leq &\frac{\lambda_1}{1-\gamma_0}\norm{X}_F \norm{\hat{y}(0)-y} \beta^s,
	\end{align}
	where $\beta:=\sqrt{1-\eta\alpha_0^2/4}$, and the second inequality is due to Eq.\eqref{eq1: Phi}, and the last inequality is by inductive hypothesis (iii). Then the difference between $b(k+1)$ and $b(0)$ is given by
	\begin{align*}
		\Norm{b(k+1)-b(0)}
		\leq& \eta\sum_{s=0}^{k} \Norm{(\partial L/\partial b)(s)}\\
		\leq& \eta \frac{\lambda_1}{1-\gamma_0}\norm{X}_F \norm{\hat{y}(0)-y}\cdot\sum_{s=0}^{k} \beta^s\\
		\leq&\frac{\eta}{1-\beta} \frac{\lambda_1}{1-\gamma_0}\norm{X}_F \norm{\hat{y}(0)-y}\\
		\leq&\frac{8}{\alpha_0^2} \frac{\lambda_1}{1-\gamma_0}\norm{X}_F \norm{\hat{y}(0)-y}\\
		\leq &C_3,	
	\end{align*}
	where the last inequality is due to the initial assumption. Then we obtain
	\begin{align*}
		\norm{b(k+1)}\leq \norm{b(k+1)-b(0)} + \norm{b(0)}\leq \lambda_3.
	\end{align*}
	With the similar argument, we can bound the partial derivative of $L$ with respect to $W$
	\begin{align}
		\Norm{(\partial L/\partial W)(s)}
		=&\Norm{\left[D(s)E(s) \left(I_m\otimes X\right) \right]^TQ(s)^{-T}\left(b(s)^T\otimes I_N\right)^T(\hat{y}(s)-y)}\nonumber\\
		\leq &\norm{X}_F\cdot \norm{Q(s)^{-1}} \cdot \norm{b(s)}\cdot \norm{\hat{y}(s)-y}\nonumber\\
		\leq &\frac{\lambda_3}{1-\gamma_0}\norm{X}_F\norm{\hat{y}(s)-y}\nonumber\\
		\leq &\frac{\lambda_3}{1-\gamma_0}\norm{X}_F\norm{\hat{y}(0)-y}\cdot \beta^s,\label{eq:dL/dW}
	\end{align}
	and the difference between $W(k+1)$ and $W(0)$ is given by 
	\begin{align*}
		\norm{W(k+1)-W(0)}
		\leq& \eta\sum_{s=0}^k\norm{(\partial L/\partial W)(s)}\\
		\leq & \eta \frac{\lambda_3}{1-\gamma_0}\norm{X}_F\norm{\hat{y}(0)-y}\cdot \sum_{s=0}^k\beta^s\\
		\leq & \frac{\eta}{1-\beta} \frac{\lambda_3}{1-\gamma_0}\norm{X}_F\norm{\hat{y}(0)-y}\\
		\leq &\frac{8}{\alpha_0^2} \frac{\lambda_3}{1-\gamma_0}\norm{X}_F\norm{\hat{y}(0)-y}\\
		\leq & C_1,
	\end{align*}
	where the last inequality is due to the initial assumption \eqref{eq:GD1}. 
	This implies
	\begin{align*}
		\norm{W(k+1)}\leq \norm{W(k)-W(0)} + \norm{W(0)}\leq  \lambda_1.
	\end{align*}
	
	Similarly, we can bound the partial derivative of $L$ with respect to $A$ as follows
	\begin{align}
		\Norm{(\partial L/\partial A)(s)}
		=&\Norm{\gamma \left[D(s) (I_m \otimes Z(s)) \right]^T
			Q(s)^{-T}\left(b(s)^T\otimes I_N\right)^T(\hat{y}(s)-y)}\nonumber\\
		\leq&\gamma\norm{Z(s)}\cdot \norm{Q(s)^{-1}}\cdot\norm{b(s)}\cdot \norm{\hat{y}(s)-y}\nonumber\\
		\leq &\frac{\gamma}{(1-\gamma_0)}\lambda_1 \norm{X}_F\cdot \frac{1}{1-\gamma_0}\cdot\lambda_3\cdot \norm{\hat{y}(s)-y}\nonumber\\
		\leq &\frac{\gamma_0}{(1-\gamma_0)^2}\frac{\lambda_1\lambda_3}{\lambda_2} \norm{X}_F \cdot \norm{\hat{y}(s)-y}\nonumber\\
		\leq &\frac{\gamma_0}{(1-\gamma_0)^2}\frac{\lambda_1\lambda_3}{\lambda_2} \norm{X}_F \cdot \norm{\hat{y}(0)-y}\cdot \beta^s,\label{eq:dL/dA}
	\end{align}
	and the difference between $A(k+1)$ and $A(0)$ is given by 
	\begin{align*}
		\norm{A(k+1)-A(0)}
		\leq &\eta\sum_{s=0}^k \Norm{(\partial L/\partial A)(s)}\\
		\leq &\eta \frac{\gamma_0}{(1-\gamma_0)^2}\frac{\lambda_1\lambda_3}{\lambda_2} \norm{X} \cdot \norm{\hat{y}(0)-y}\sum_{s=0}^k  \beta^s\\
		\leq &\frac{\eta}{1-\beta} \frac{\gamma_0}{(1-\gamma_0)^2}\frac{\lambda_1\lambda_3}{\lambda_2} \norm{X} \cdot \norm{\hat{y}(0)-y}\\
		\leq &\frac{8}{\alpha_0^2} \frac{\gamma_0}{(1-\gamma_0)^2}\frac{\lambda_1\lambda_3}{\lambda_2} \norm{X} \cdot \norm{\hat{y}(0)-y} \\
		\leq & C_2,
	\end{align*}
	where the last inequality is due to the initial assumption~\eqref{eq:GD1}.
	Therefore, we obtain
	\begin{align*}
		\norm{A(k+1)}\leq \norm{A(k+1)-A(0)} + \norm{A(0)} \leq  \lambda_2.
	\end{align*}
	Since $\norm{A(k+1)}\leq \lambda_2$ and $\gamma_0=\gamma\lambda_2 < 1$, Lemma~\ref{lemma: forward} implies the unique equilibrium point $Z(k+1)$ exists.
	
	It follows from Lemma~\ref{lemma: Z_a-Z_b}, Eq.\eqref{eq:dL/dW}, and Eq.\eqref{eq:dL/dA} that for any $s$ and $\ell$, we have
	\begin{align}
		\norm{Z(s+1)-Z(s)}_F
		\leq 
		\left[1 + \frac{\gamma_0^2}{(1-\gamma_0)^2}\frac{\lambda_1^2}{\lambda_2^2}\right]\eta\frac{\lambda_3}{(1-\gamma_0)^2}\norm{X}_F^2\norm{\hat{y}(s)-y}.
	\end{align}
	Therefore, we have
	\begin{align*}
		\norm{Z(k)-Z(0)}_F\leq& \sum_{s=0}^k\norm{Z(s+1)-Z(s)}_F\\
		\leq & \sum_{s=0}^k\left[1 + \frac{\gamma_0^2}{(1-\gamma_0)^2}\frac{\lambda_1^2}{\lambda_2^2}\right]\eta\frac{\lambda_3}{(1-\gamma_0)^2}\norm{X}^2\norm{\hat{y}(0)-y}\cdot \beta^s\\
		\leq &\left[1 + \frac{\gamma_0^2}{(1-\gamma_0)^2}\frac{\lambda_1^2}{\lambda_2^2}\right]\eta\frac{\lambda_3}{(1-\gamma_0)^2}\norm{X}_F^2\norm{\hat{y}(0)-y}\cdot \frac{1}{1-\beta}\\
		\leq &\left[1 + \frac{\gamma_0^2}{(1-\gamma_0)^2}\frac{\lambda_1^2}{\lambda_2^2}\right]\frac{\lambda_3}{(1-\gamma_0)^2}\norm{X}_F^2\norm{\hat{y}(0)-y}\cdot \frac{8}{\alpha_0^2}\\
		\leq &\alpha_0/2,
	\end{align*}
	where the last inequality is due to the initial assumption Eq.\eqref{eq:GD1}-Eq.\eqref{eq:GD3}. As a result, we obtain
	\begin{align}
		\sigma_{\min}\left[Z(k)\right]\geq \sigma_{\min}\left[Z(0)\right] - \norm{Z(k)-Z(0)}\geq \alpha_0/2.
	\end{align}
	
	Now, we are ready to derive the linear convergence result. Note that
	\begin{align*}
		L(\theta(k+1))=&\frac{1}{2}\Norm{\hat{y}(k+1)-y}^2 \\
		=&\frac{1}{2} \norm{\hat{y}(k+1)-\hat{y}(k) + \hat{y}(k) - y}^2\\
		=&\frac{1}{2}\norm{\hat{y}(k+1)-\hat{y}(k)}^2 + \inn{\hat{y}(k+1)-\hat{y}(k)}{\hat{y}(k)-y}  +L(\theta(k)).
	\end{align*}
	In the rest of the proof, we will bound each term above. Note that
	\begin{align*}
		\norm{\hat{y}(k+1)-\hat{y}(k)}
		=&\norm{Z(k+1)b(k+1)-Z(k)b(k)}\\
		\leq &\norm{Z(k+1)b(k+1)-Z(k)b(k+1)} + \norm{Z(k)b(k+1)-Z(k)b(k)}\\
		\leq &\norm{Z(k+1)-Z(k)}\norm{b(k+1)} + \norm{Z(k)}\norm{b(k+1)-b(k)}\\
		\leq & \left[1 + \frac{\gamma_0^2}{(1-\gamma_0)^2}\frac{\lambda_1^2}{\lambda_2^2}\right]\eta\frac{\lambda_3}{(1-\gamma_0)^2}\norm{X}_F^2\norm{\hat{y}(k)-y}\cdot \lambda_3
		+\frac{1}{1-\gamma_0}\Norm{\Phi(k)}_F \cdot \eta\norm{(\partial L/\partial b)(k)}\\
		\leq&\left[1 + \frac{\gamma_0^2}{(1-\gamma_0)^2}\frac{\lambda_1^2}{\lambda_2^2}\right]\eta\frac{\lambda_3^2}{(1-\gamma_0)^2}\norm{X}_F^2\norm{\hat{y}(k)-y}
		+\eta\frac{ \lambda_1^2}{(1-\gamma_0)^2} \norm{X}_F^2 \norm{\hat{y}(k)-y}\\
		\leq &\eta\underbrace{\left(\left[1 + \frac{\gamma_0^2}{(1-\gamma_0)^2}\frac{\lambda_1^2}{\lambda_2^2}\right]\lambda_1^{-2}+\lambda_3^{-2}\right)\frac{1}{(1-\gamma_0)^2}\lambda_1^2\lambda_3^2\norm{X}_F^2}_{:=Q_1}\cdot\norm{\hat{y}(k)-y}.
	\end{align*}
	Let $g:=Z(k)b(k+1)$, then we have
	\begin{align*}
		\inn{\hat{y}(k+1)-\hat{y}(k)}{\hat{y}(k)-y} 
		=\inn{\hat{y}(k+1)-g}{\hat{y}(k)-y} + \inn{g-\hat{y}(k)}{\hat{y}(k)-y} ,
	\end{align*}
	where
	\begin{align*}
		\inn{\hat{y}(k+1)-g}{\hat{y}(k)-y}
		=&\inn{Z(k+1)b(k+1)-Z(k)b(k+1)}{\hat{y}(k)-y}\\
		\leq & \norm{Z(k+1)-Z(k)}\norm{b(k+1)}\norm{\hat{y}(k)-y}\\
		\leq &  	\left[1 + \frac{\gamma_0^2}{(1-\gamma_0)^2}\frac{\lambda_1^2}{\lambda_2^2}\right]\eta\frac{\lambda_3}{(1-\gamma_0)^2}\norm{X}_F^2\norm{\hat{y}(k)-y}\cdot \lambda_3 \cdot \norm{\hat{y}(k)-y}\\
		=& \eta\underbrace{	\left[1 + \frac{\gamma_0^2}{(1-\gamma_0)^2}\frac{\lambda_1^2}{\lambda_2^2}\right]\frac{\lambda_3^2}{(1-\gamma_0)^2}\norm{X}_F^2}_{:=Q_2}\cdot\norm{\hat{y}(k)-y}^2
	\end{align*}
	and
	\begin{align*}
		\inn{g-\hat{y}(k)}{\hat{y}(k)-y} 
		=&\inn{Z(k)b(k+1)-Z(k)b(k)}{\hat{y}(k)-y}\\
		=&-\eta (\hat{y}(k)-y)^T\left[Z(k)Z(k)^T\right](\hat{y}(k)-y)\\
		\leq& -\eta \cdot (\alpha_0^2/4)\norm{\hat{y}(k)-y}^2.
	\end{align*}
	Combining all the results together, we have
	\begin{align*}
		L(\theta(k+1))
		\leq&\left[1 - \eta\frac{\alpha_0^2}{2} + \eta^2Q_1^2 +2\eta Q_2 \right]L(\theta(k))\\
		\leq &\left[1-\eta\left(\frac{\alpha_0^2}{2}-4Q_2\right)\right]L(\theta(k)),\quad\text{By Eq.\eqref{eq:step size}}\\
		\leq &\left[1-\eta\frac{\alpha_0^2}{4}\right]L(\theta(k)),
	\end{align*}
	where the last inequality follows from Eq.\eqref{eq:GD3}.
\end{proof}

\subsection{Proof of Lemma~\ref{lemma: initial}}\label{app: initial}
Note that $\Phi_{ij}\geq 0$ by the definition of Eq.\eqref{eq: def Phi}. Since $Z^0=0$, we have
\begin{align*}
	Z^{1} = \sigma(\gamma Z^0A + \Phi)
	=\sigma(\Phi) = \Phi.
\end{align*}
Since $A_{ij}\geq 0$, we have
\begin{align*}
	Z^{2} = \sigma(\gamma Z^1 A+\Phi)
	=\sigma(\gamma \Phi A + \Phi)
	=\gamma \Phi A + \Phi
	=\Phi\left(\gamma A + I_m\right)
\end{align*}
Repeating the same argument $\ell$ times, we have
\begin{align*}
	Z^{\ell+1} = \Phi\left(\hat{A}^{\ell} + \hat{A}^{\ell-1} + \cdots + \hat{ A} + I_m\right),
\end{align*}
where $\hat{A}:=\gamma A$. Let $\ell\rightarrow\infty$, then Neumann series implies that
\begin{align*}
	Z = \Phi(I_m-\hat{A})^{-1}.
\end{align*}

\clearpage
\section{More experimental results}\label{app: exp}
In this section, we provide more experimental results to justify our theoretical findings.

\begin{figure}[h]
	\centering
	\includegraphics[width=.31\textwidth]{MNIST/train_model_loss}
	\includegraphics[width=.31\textwidth]{MNIST/test_model_loss}
	\includegraphics[width=.31\textwidth]{MNIST/opn}
	\includegraphics[width=.31\textwidth]{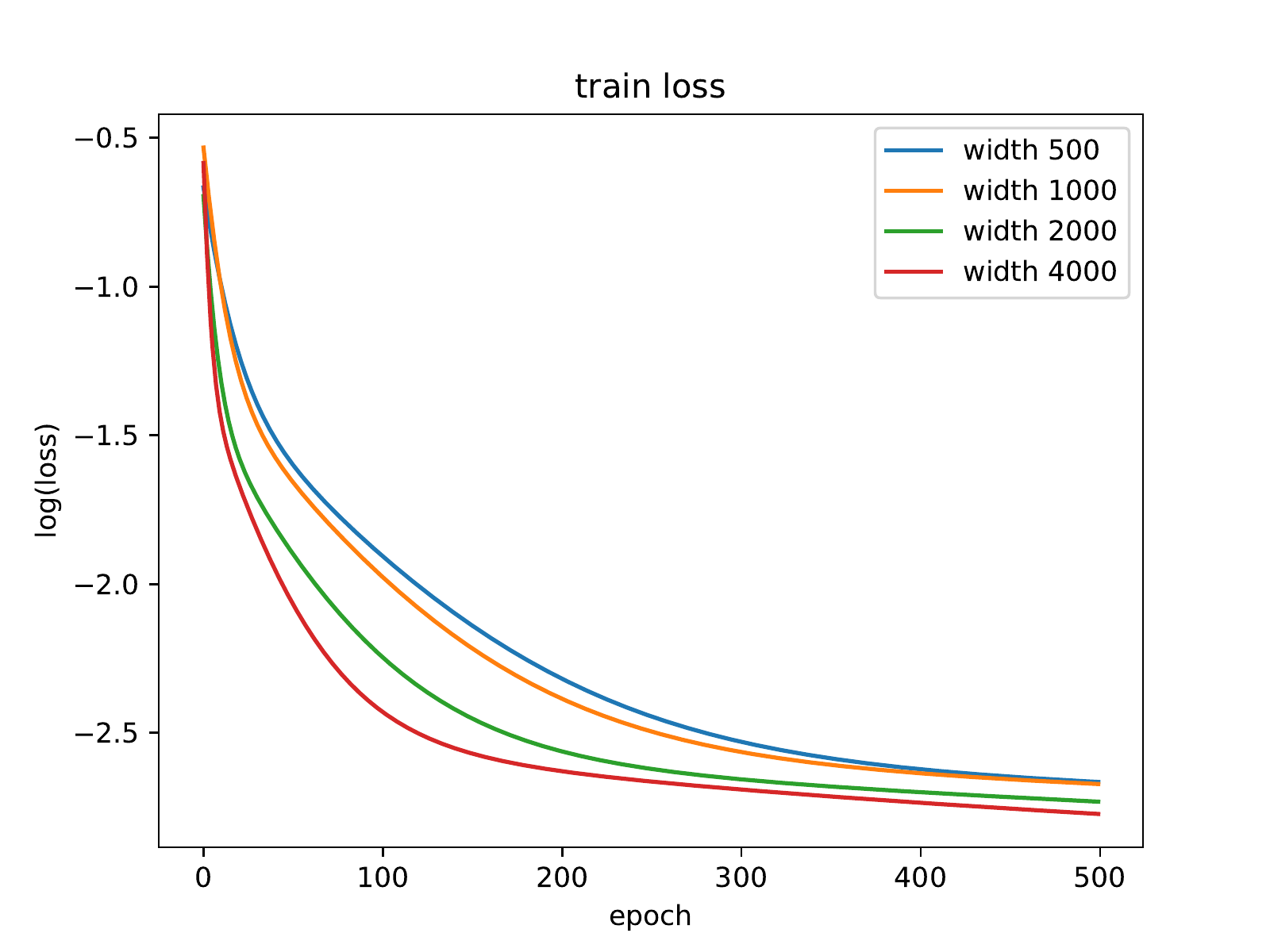}
	\includegraphics[width=.31\textwidth]{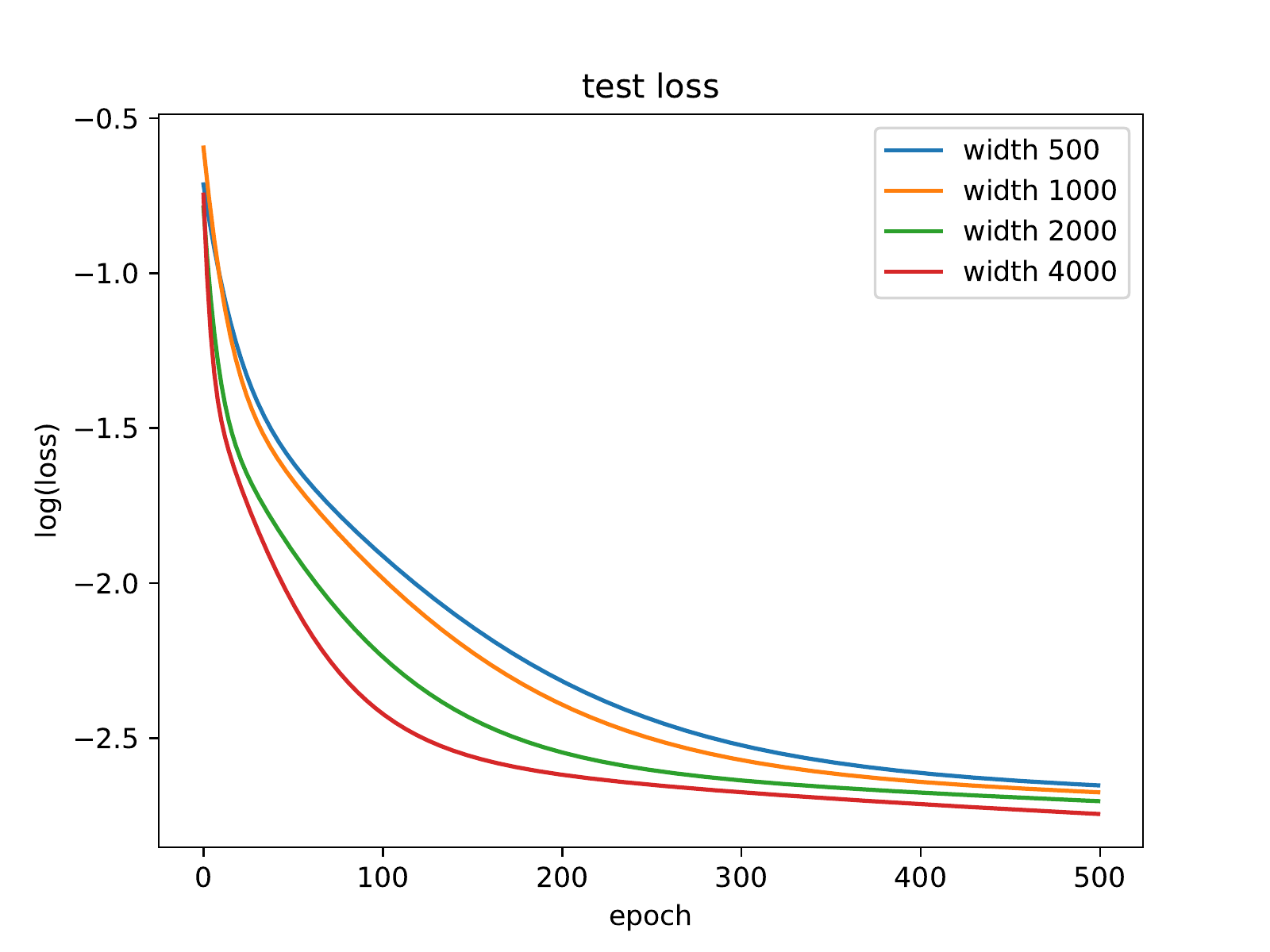}
	\includegraphics[width=.31\textwidth]{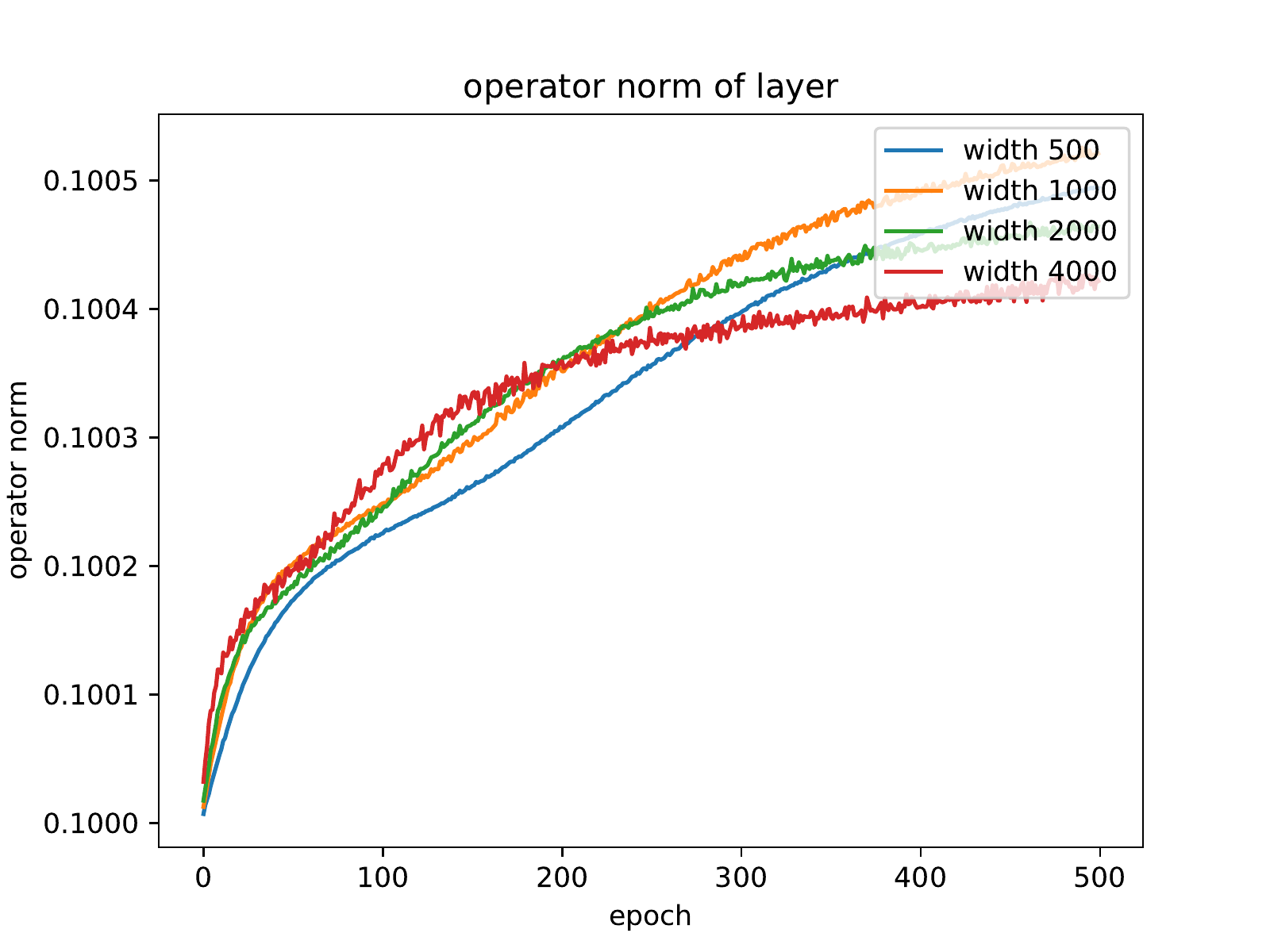}
	\includegraphics[width=.31\textwidth]{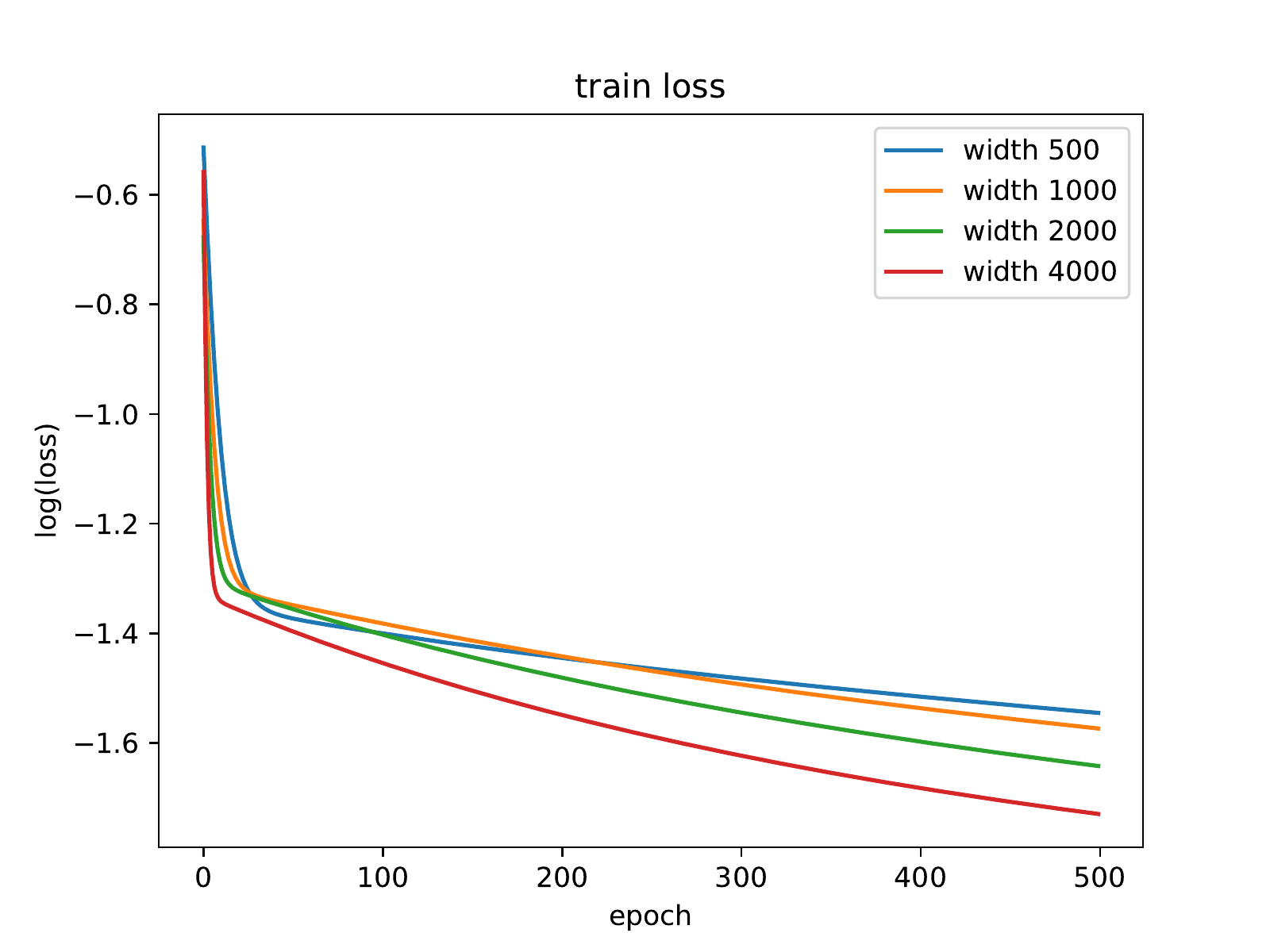}
	\includegraphics[width=.31\textwidth]{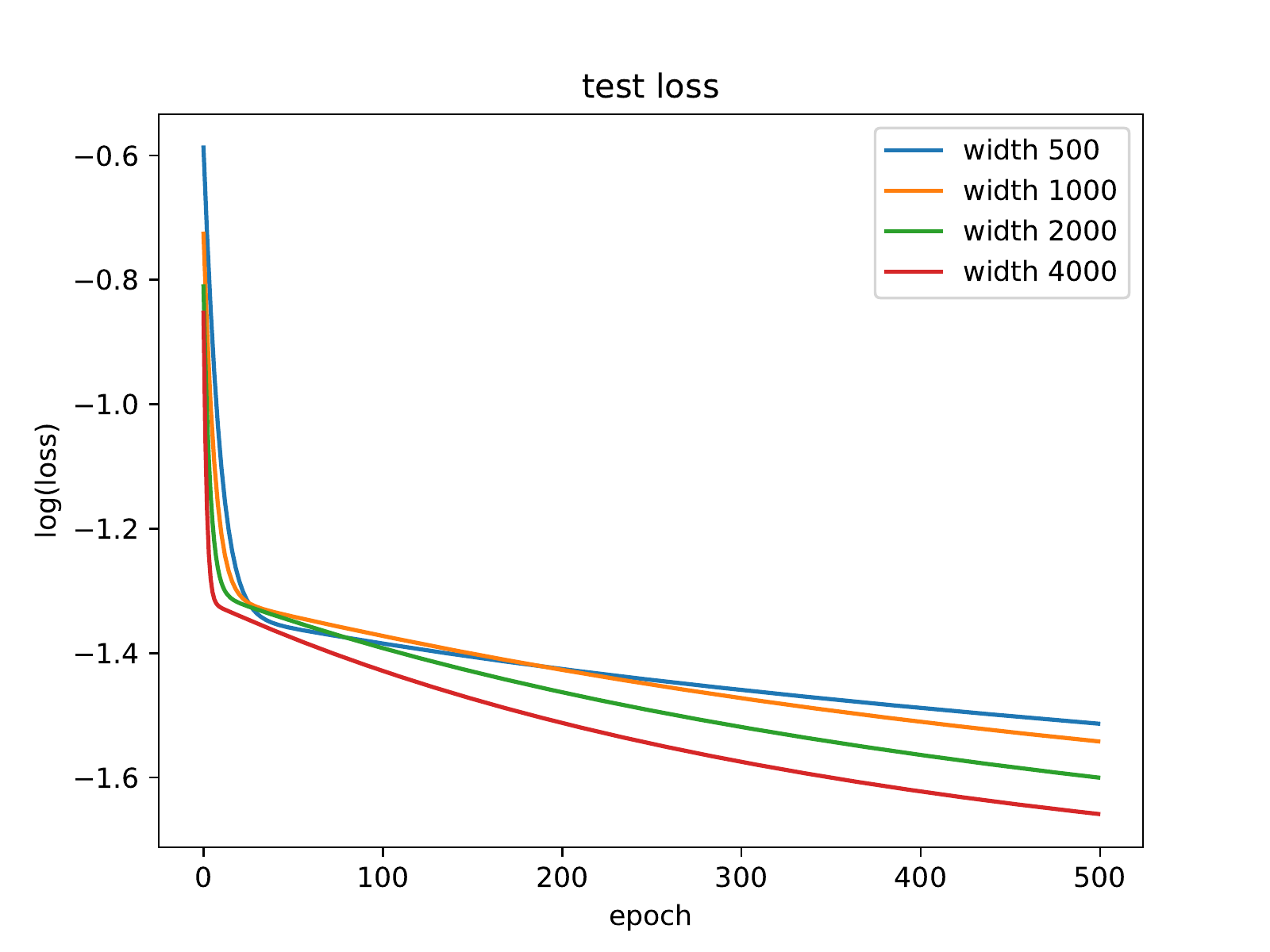}
	\includegraphics[width=.31\textwidth]{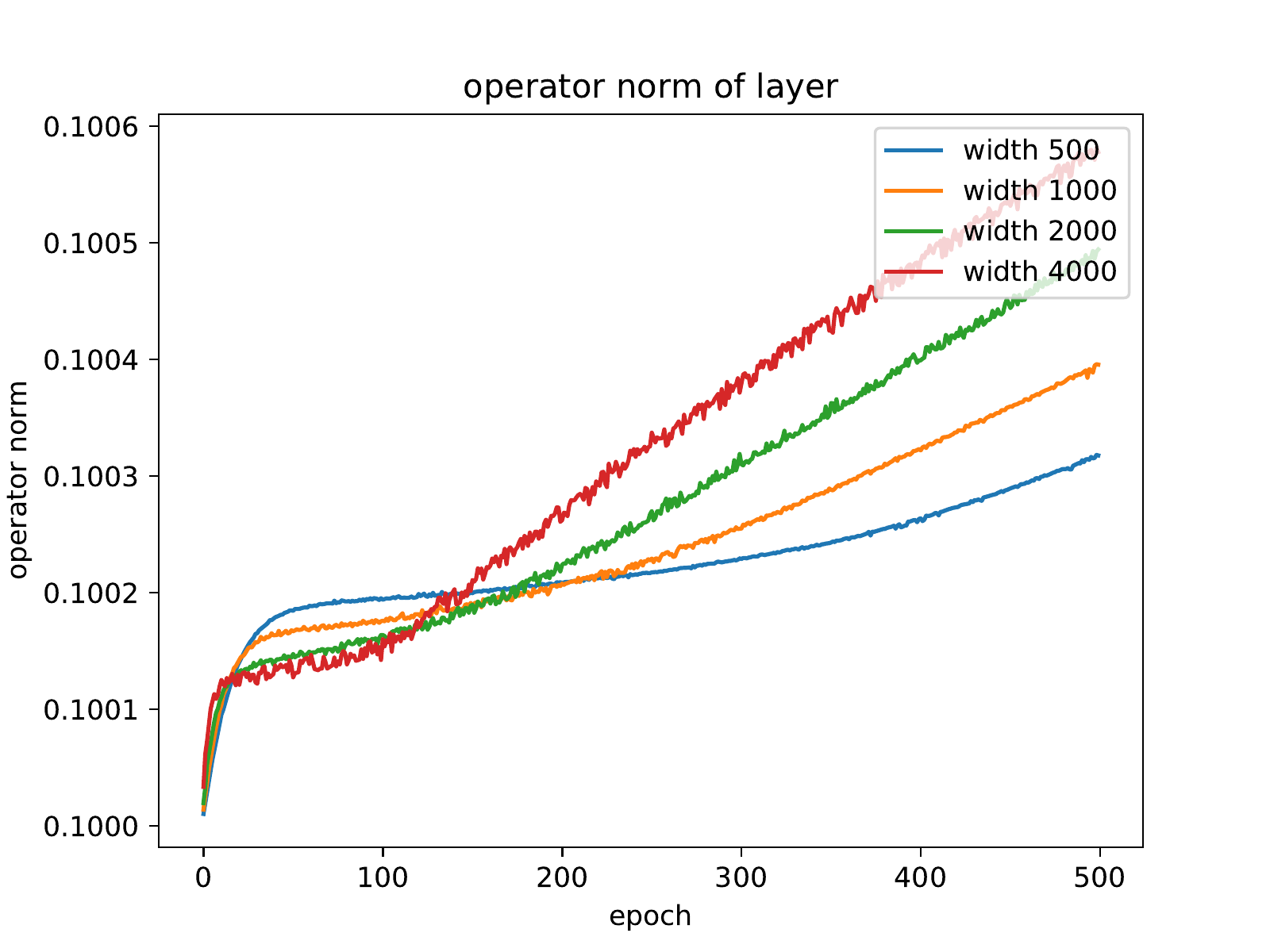}
	\includegraphics[width=.31\textwidth]{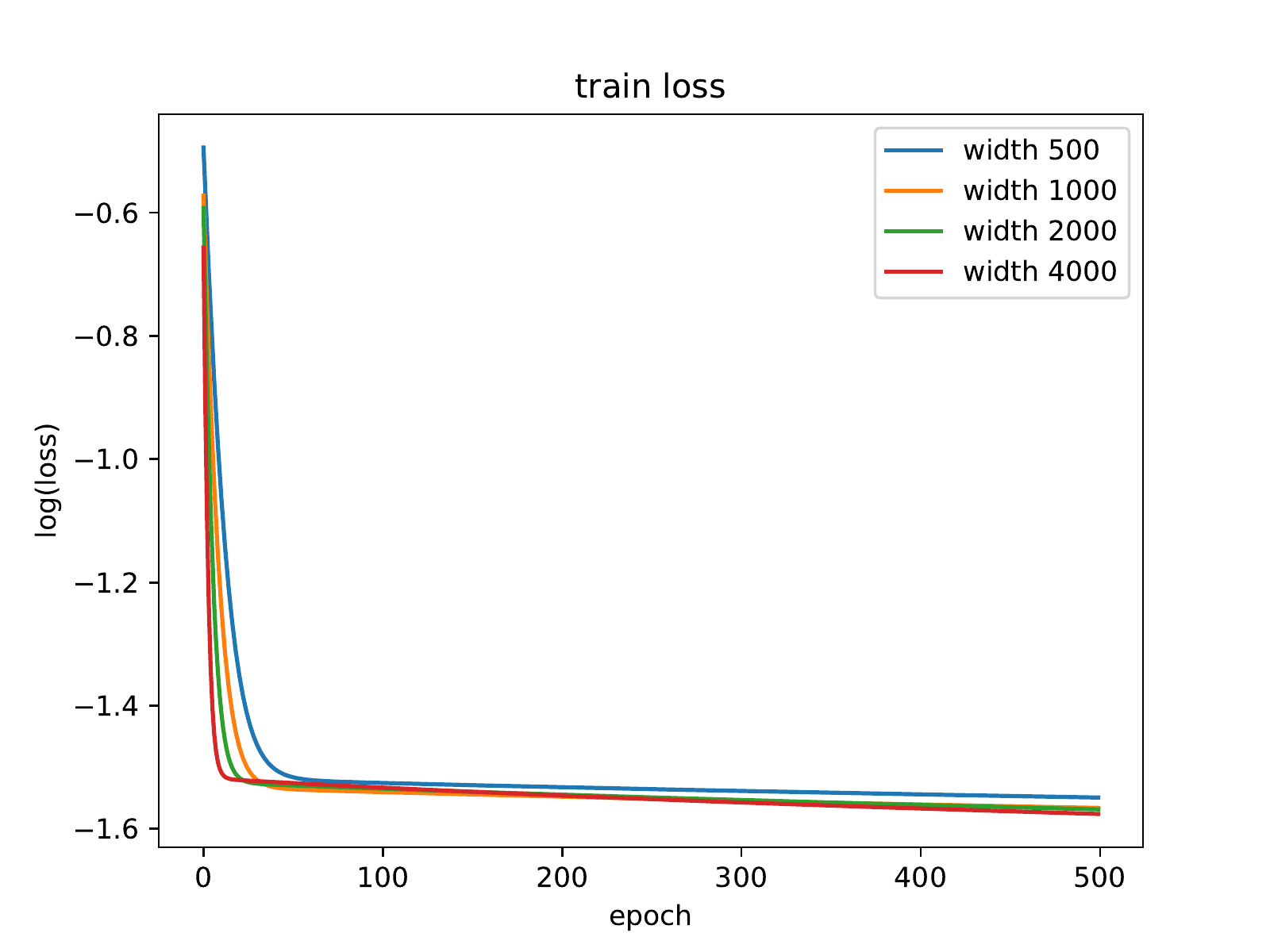}
	\includegraphics[width=.31\textwidth]{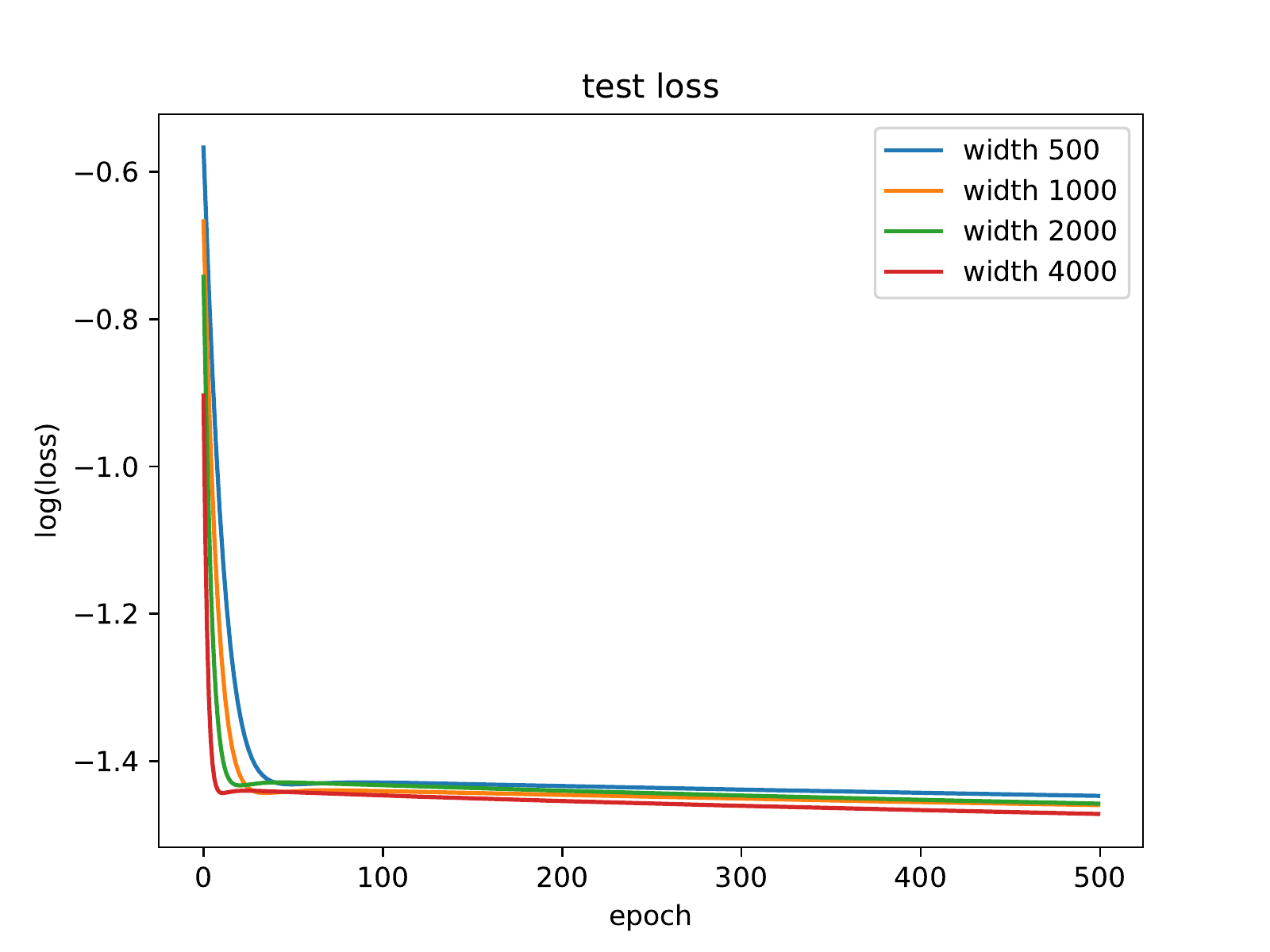}
	\includegraphics[width=.31\textwidth]{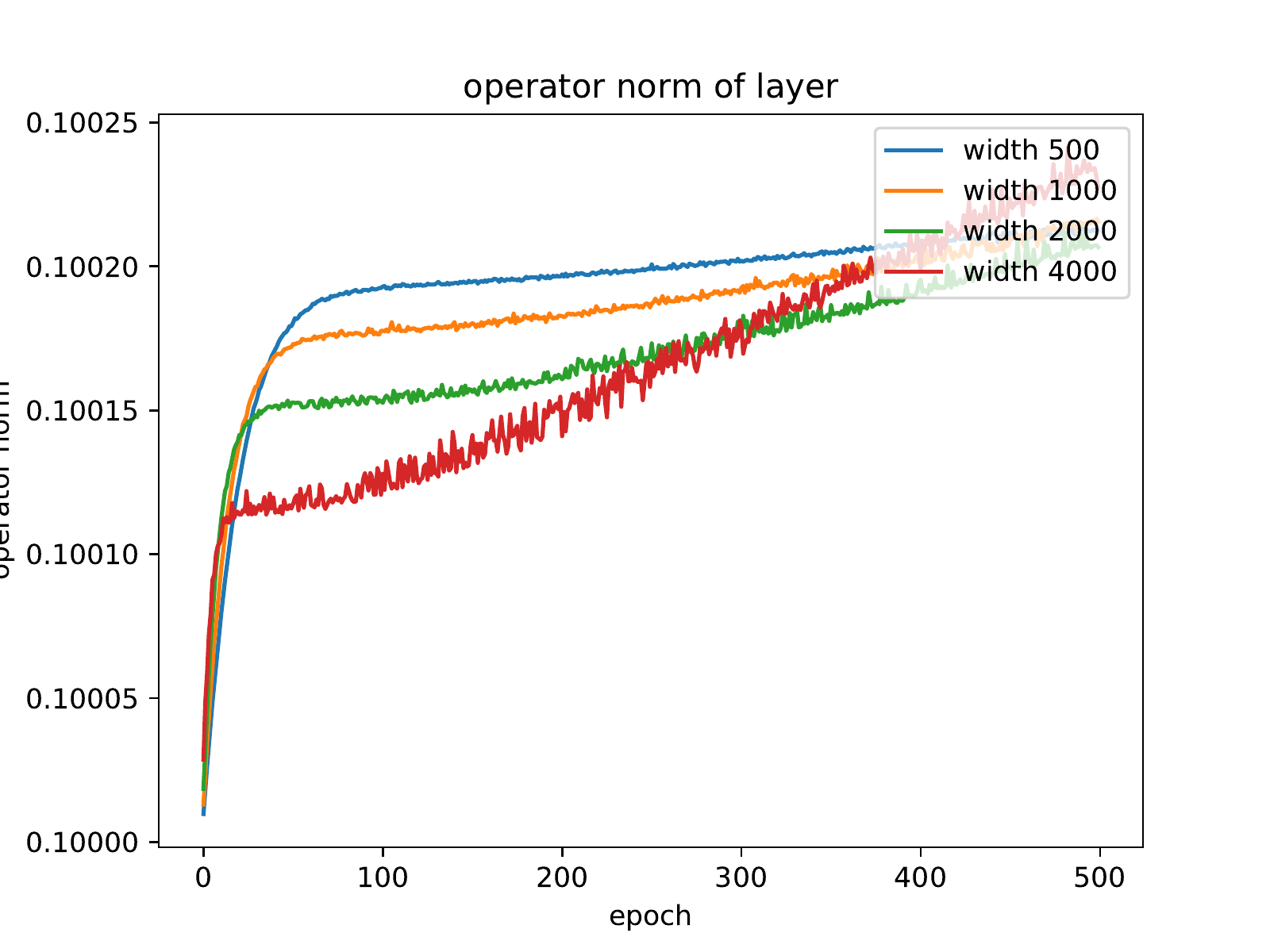}
	\caption{We evaluate the impact of the \textbf{width $m$} on the
		training loss, test loss, and operator norm of the scaled
		matrix $\gamma A(k)$ on the modified dataset of MNIST, FashionMNIST, CIFAR10, and SVHN}
\end{figure}

\begin{figure}
	\centering
	\includegraphics[width=.31\textwidth]{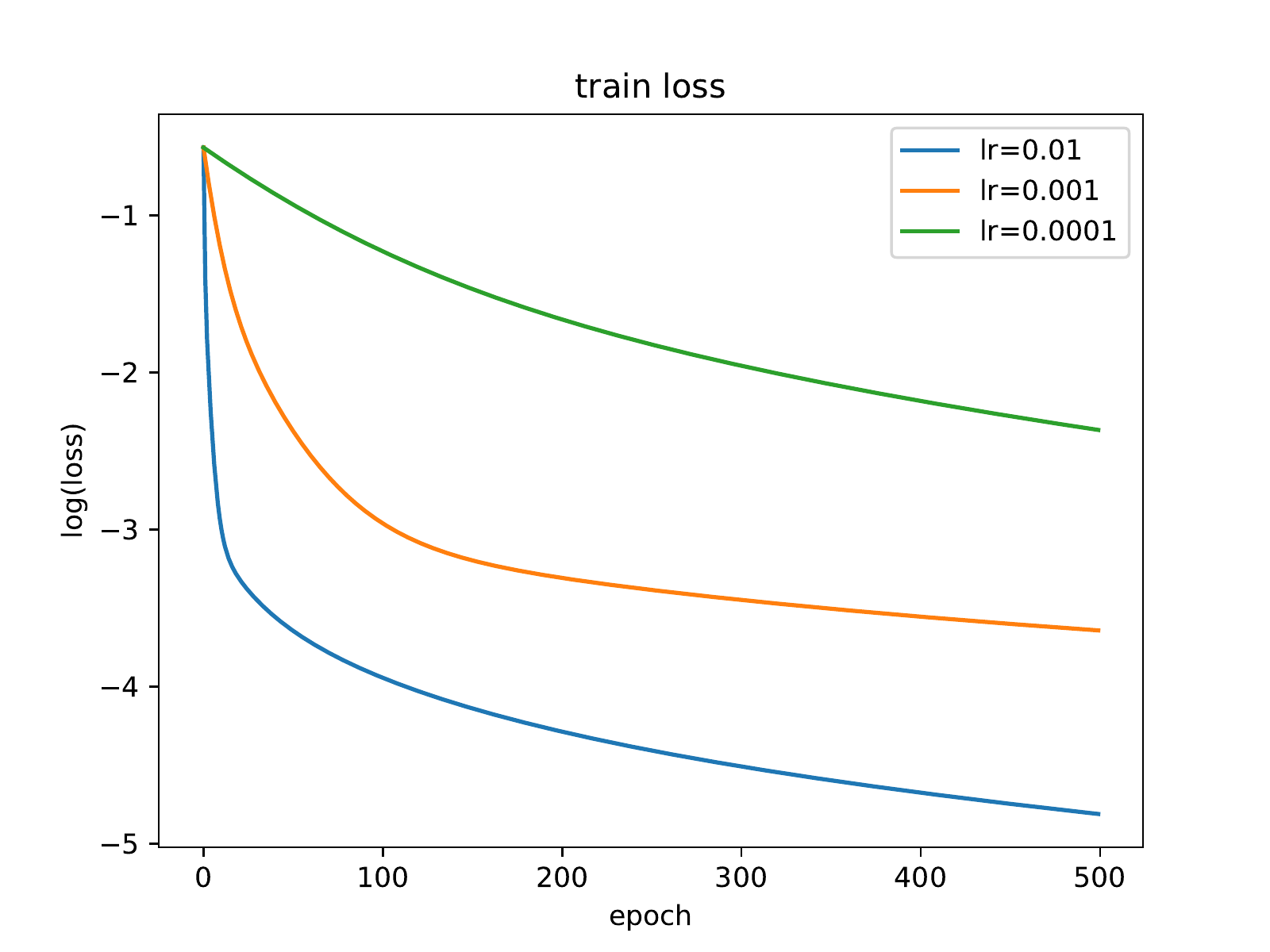}
	\includegraphics[width=.31\textwidth]{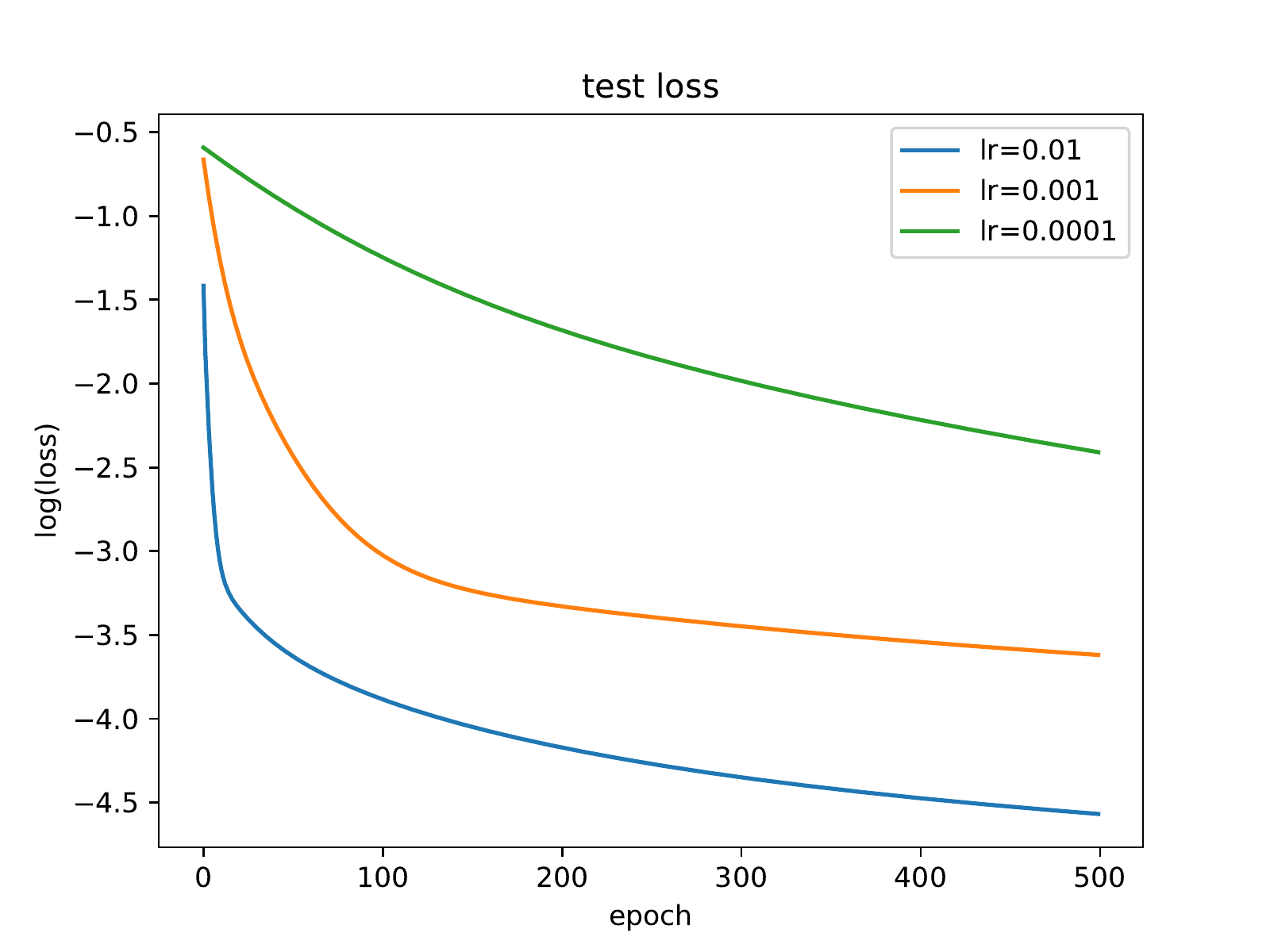}
	\includegraphics[width=.31\textwidth]{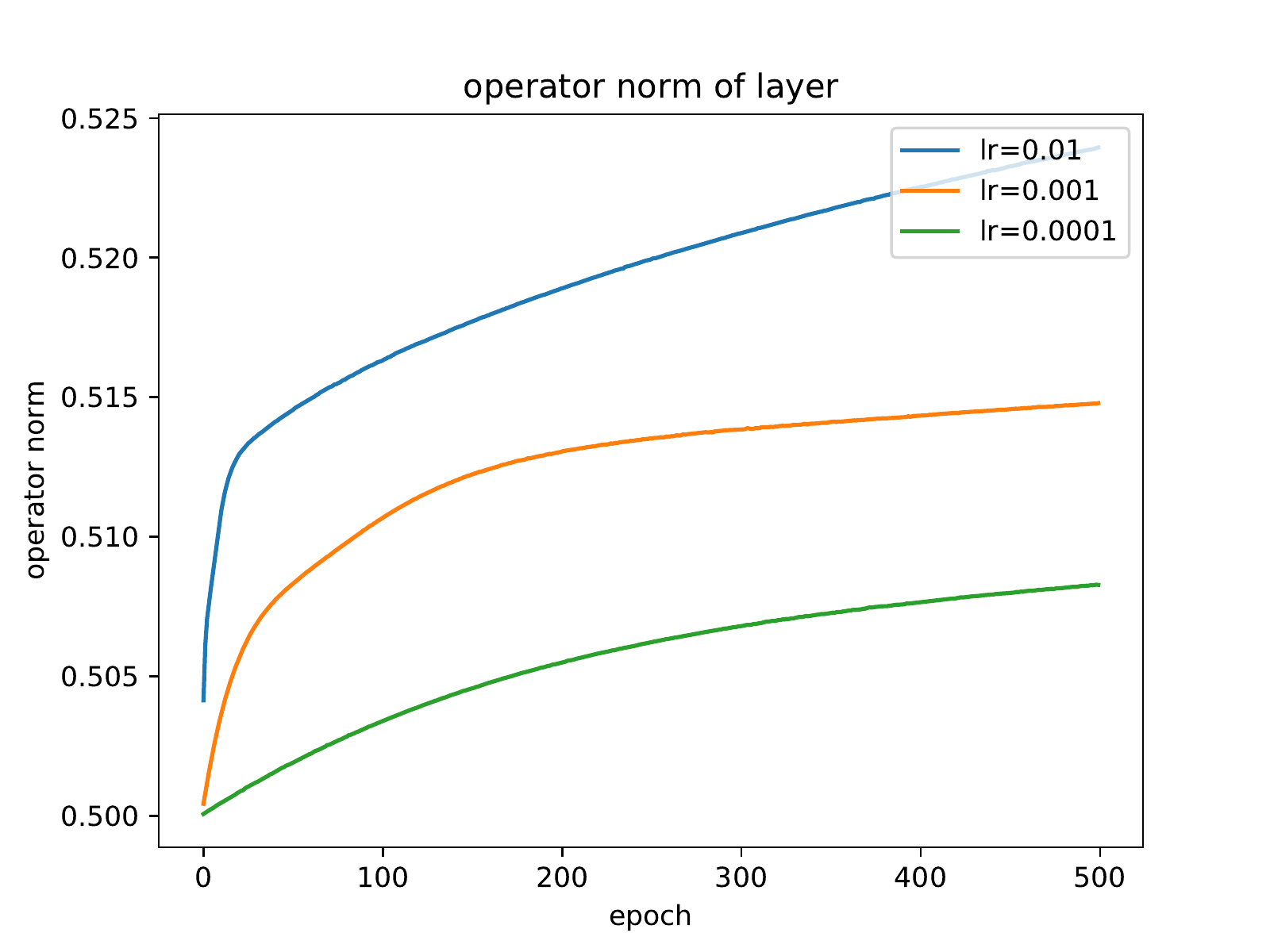}
	\includegraphics[width=.31\textwidth]{lr/FashionMNIST/train_model_loss}
	\includegraphics[width=.31\textwidth]{lr/FashionMNIST/test_model_loss}
	\includegraphics[width=.31\textwidth]{lr/FashionMNIST/opn}
	\includegraphics[width=.31\textwidth]{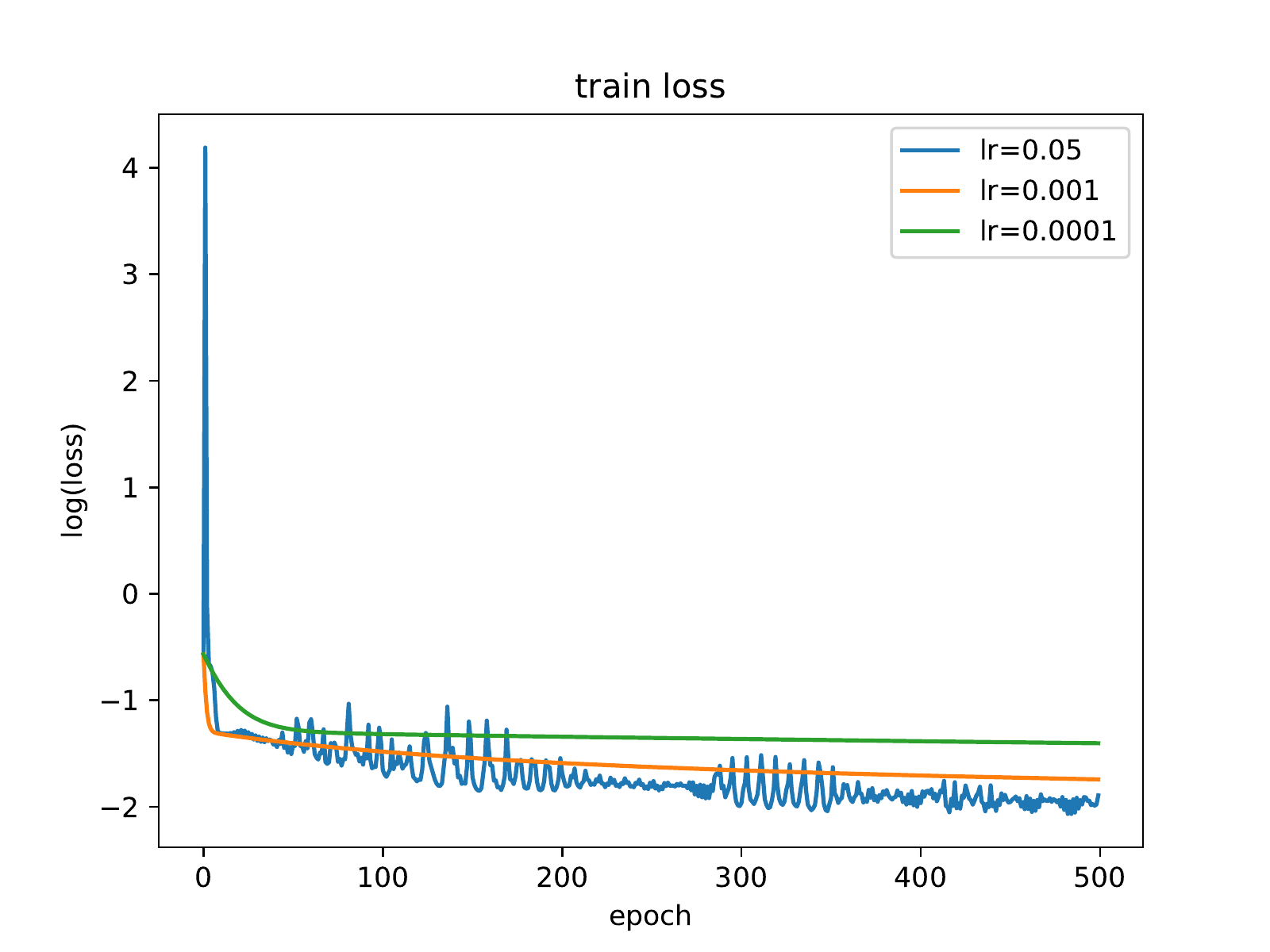}
	\includegraphics[width=.31\textwidth]{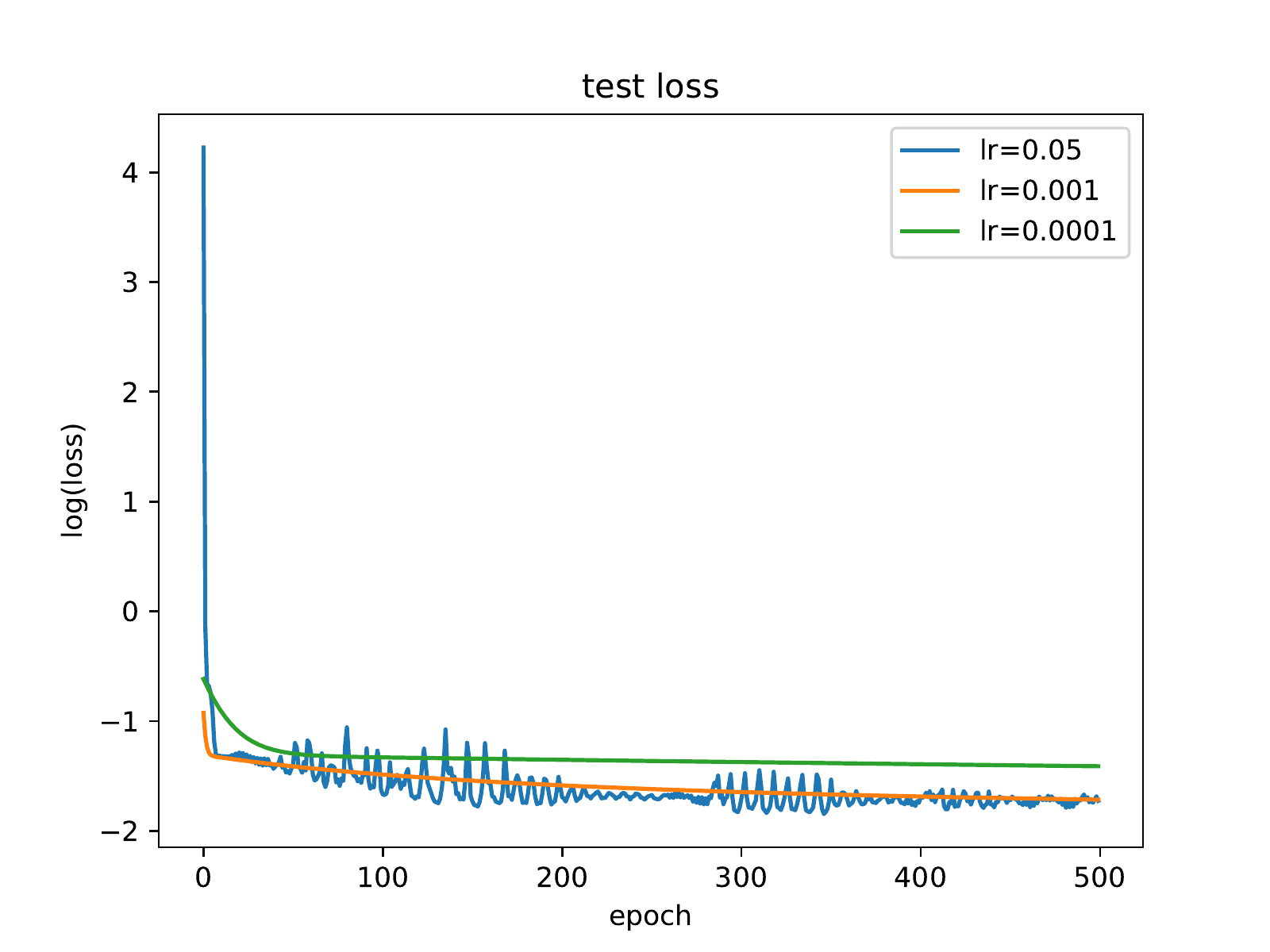}
	\includegraphics[width=.31\textwidth]{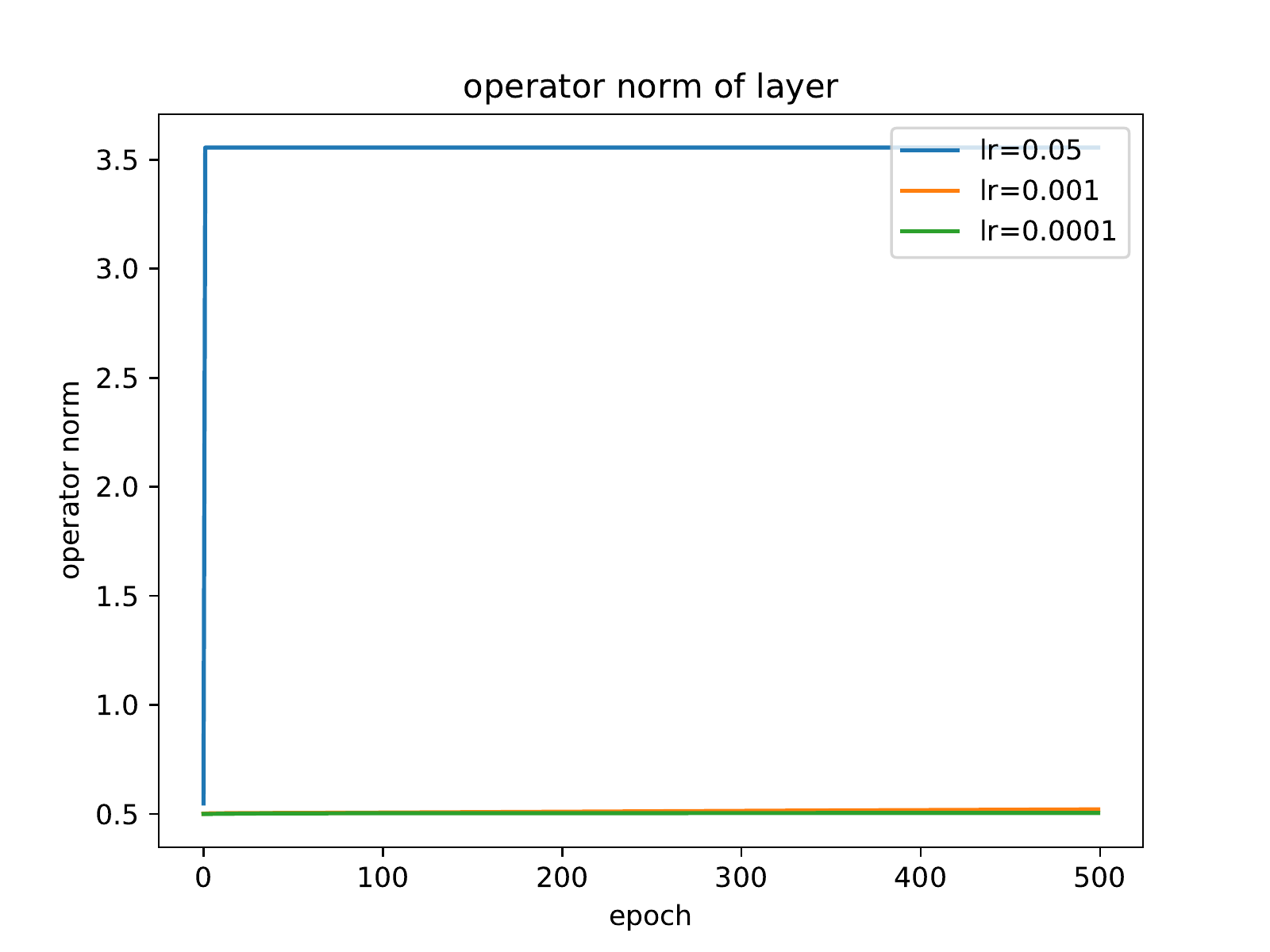}
	\includegraphics[width=.31\textwidth]{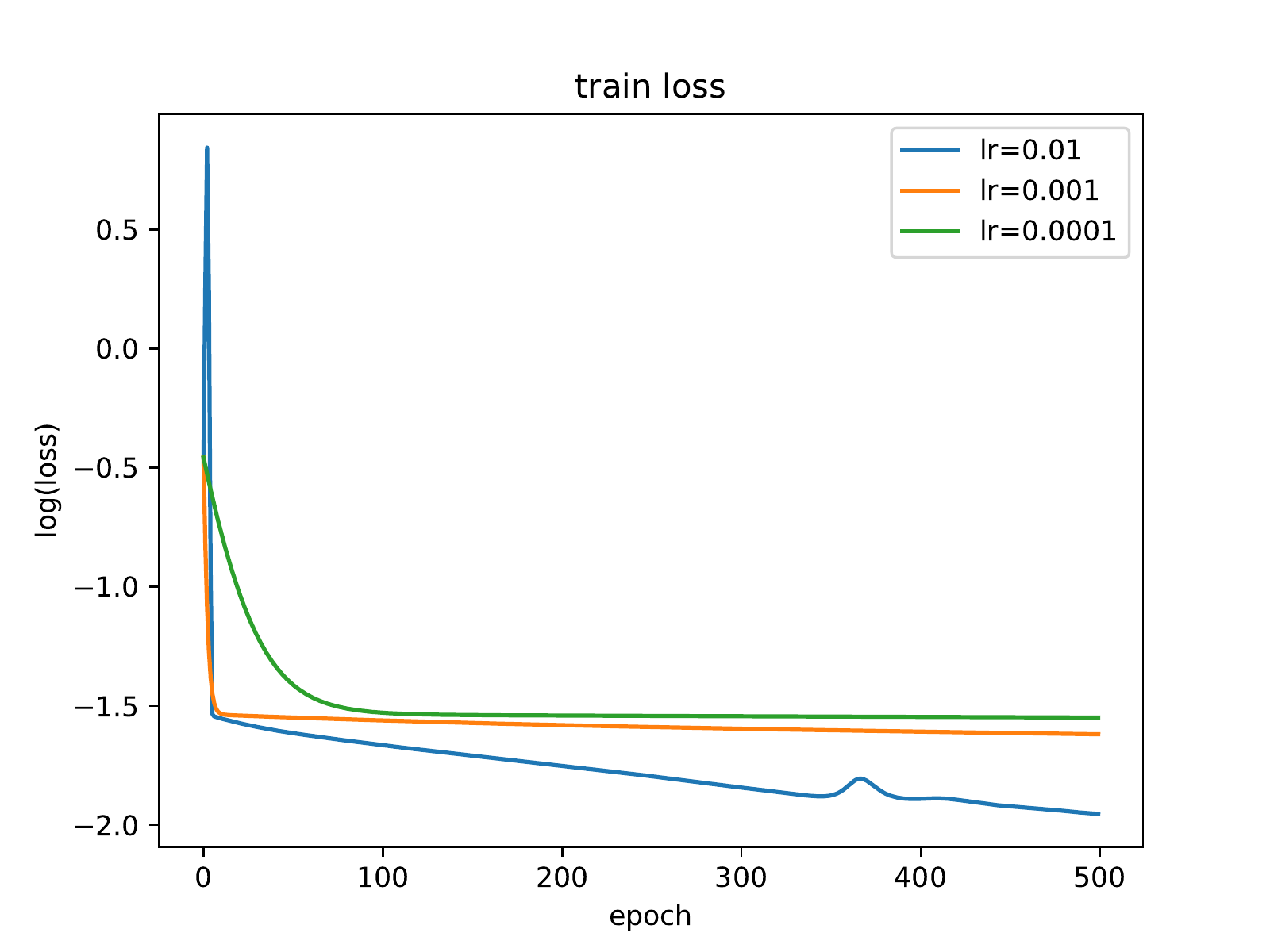}
	\includegraphics[width=.31\textwidth]{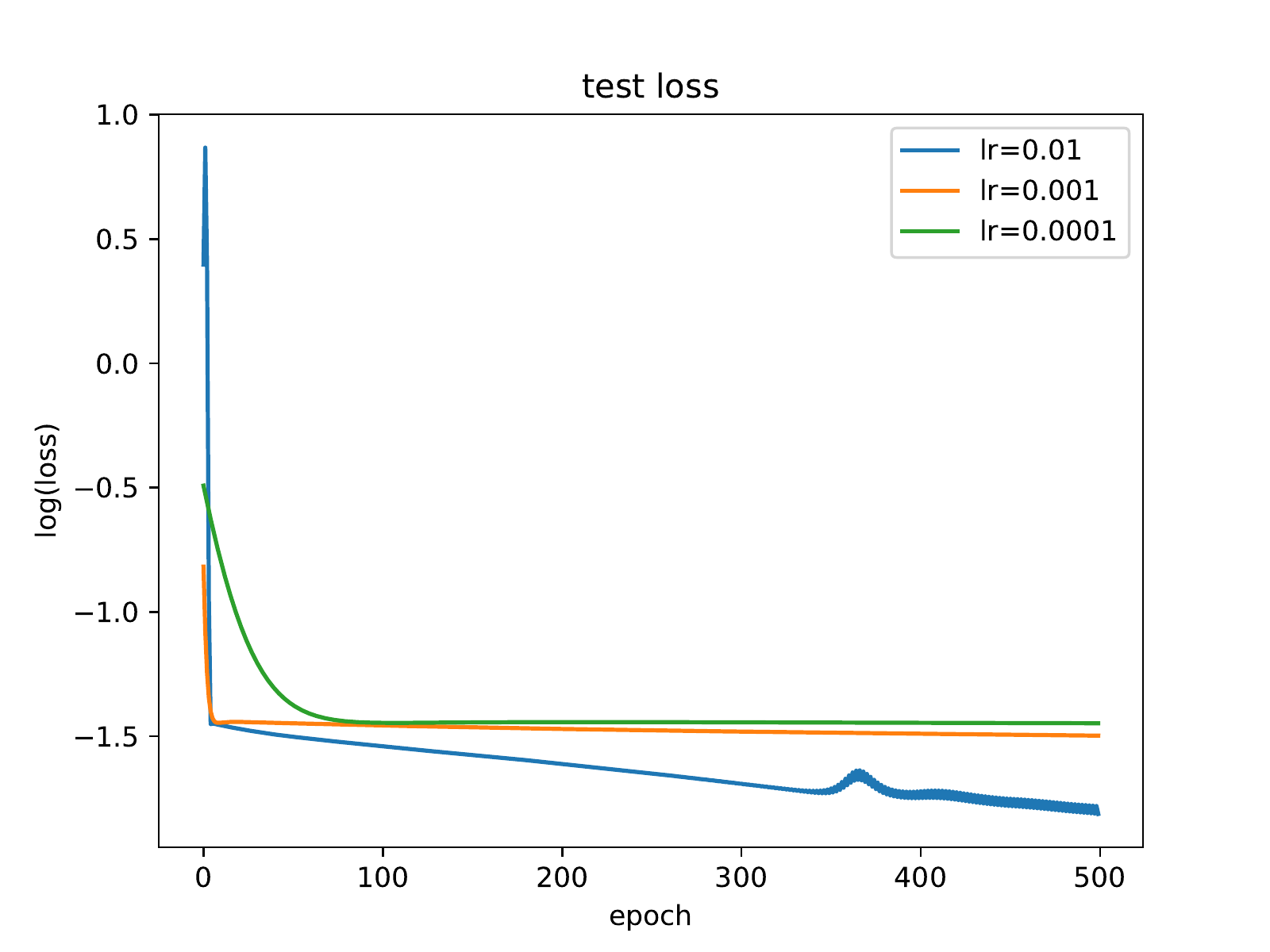}
	\includegraphics[width=.31\textwidth]{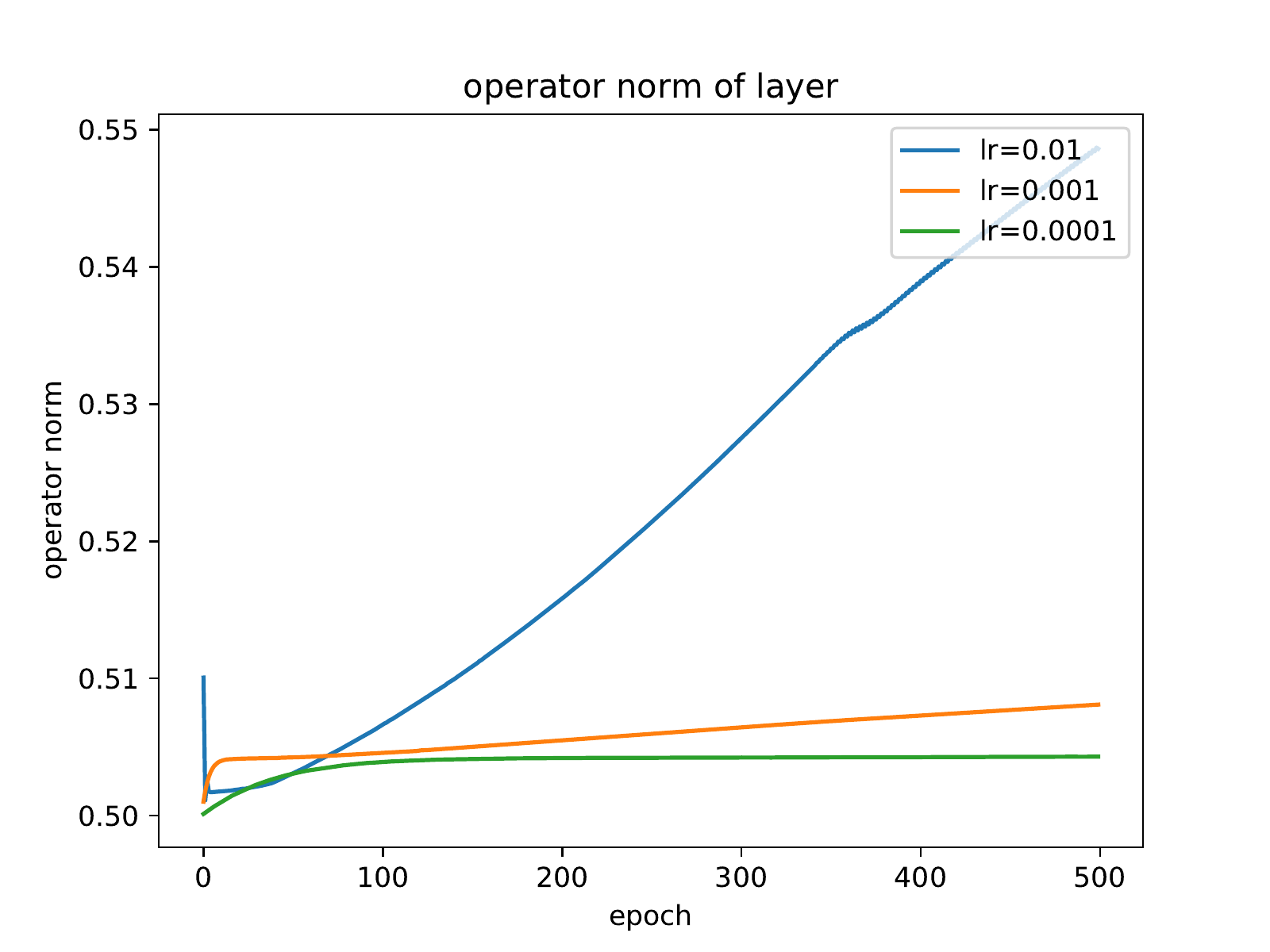}
	\caption{We evaluate the impact of the \textbf{learning rate $\eta$} on the
		training loss, test loss, and operator norm of the scaled
		matrix $\gamma A(k)$ on the modified dataset of MNIST, FashionMNIST, CIFAR10, and SVHN}
\end{figure}

\begin{figure}
	\centering
	\includegraphics[width=.31\textwidth]{gamma/MNIST/train_model_loss}
	\includegraphics[width=.31\textwidth]{gamma/MNIST/test_model_loss}
	\includegraphics[width=.31\textwidth]{gamma/MNIST/opn}
	\includegraphics[width=.31\textwidth]{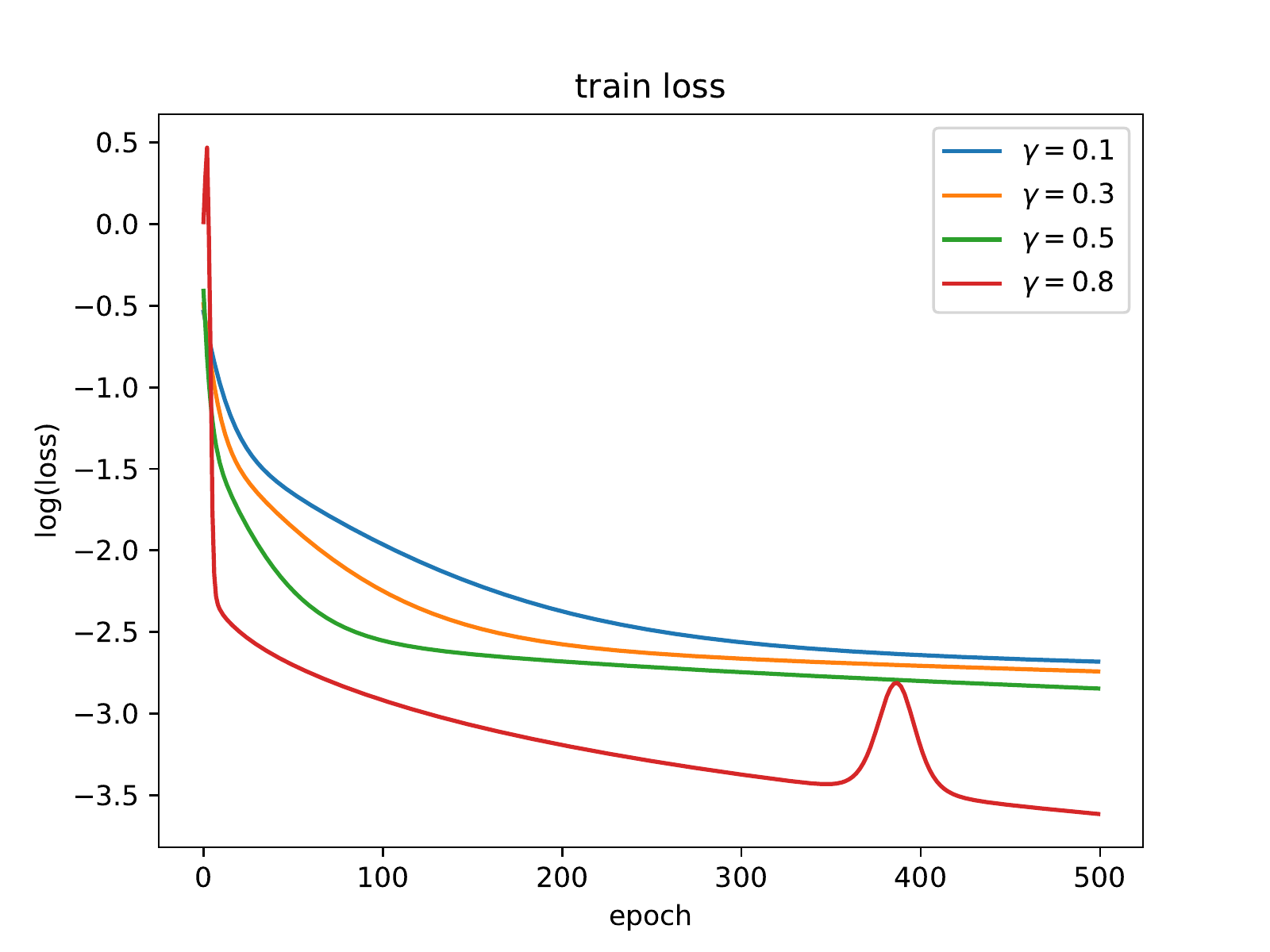}
	\includegraphics[width=.31\textwidth]{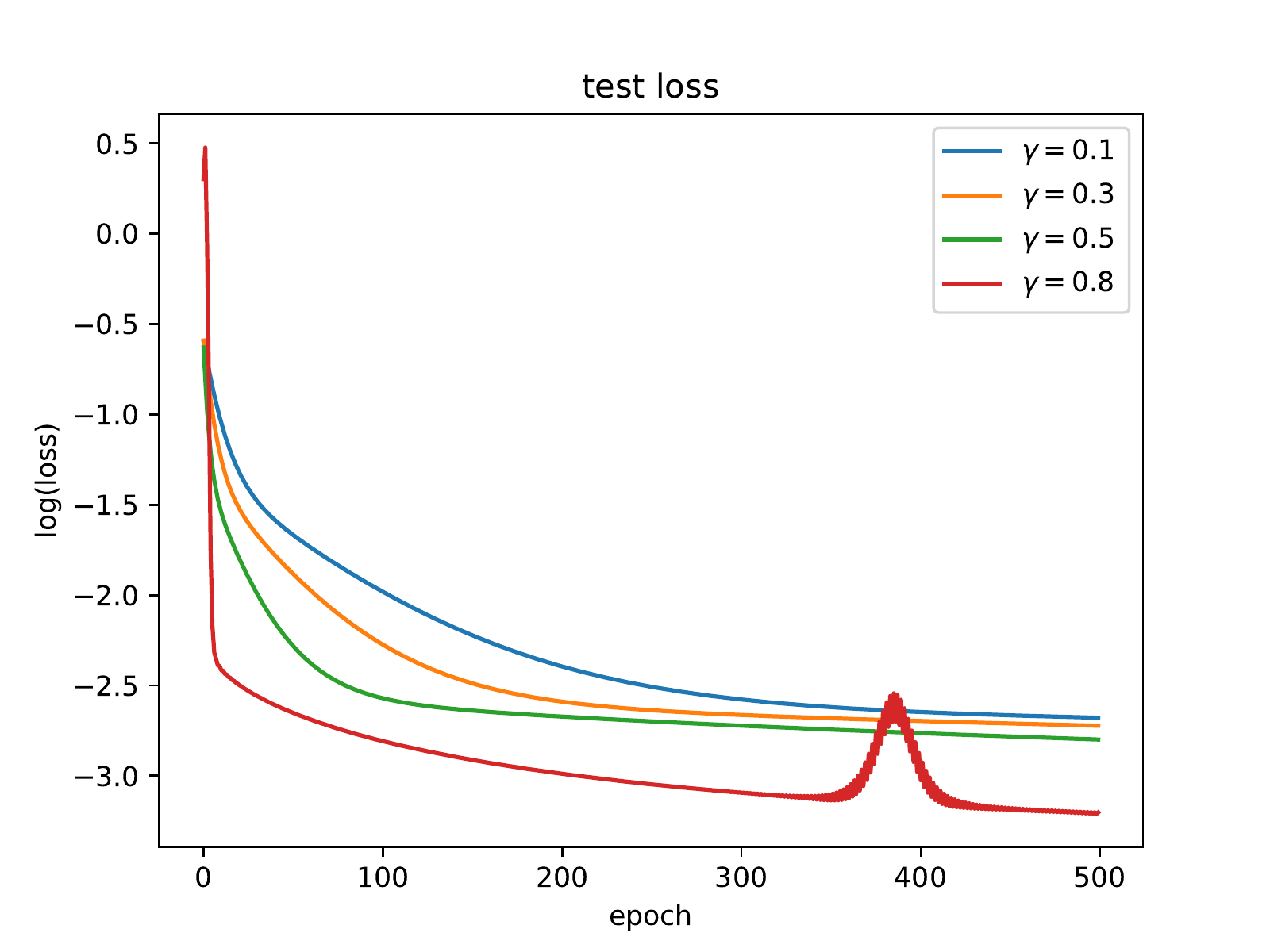}
	\includegraphics[width=.31\textwidth]{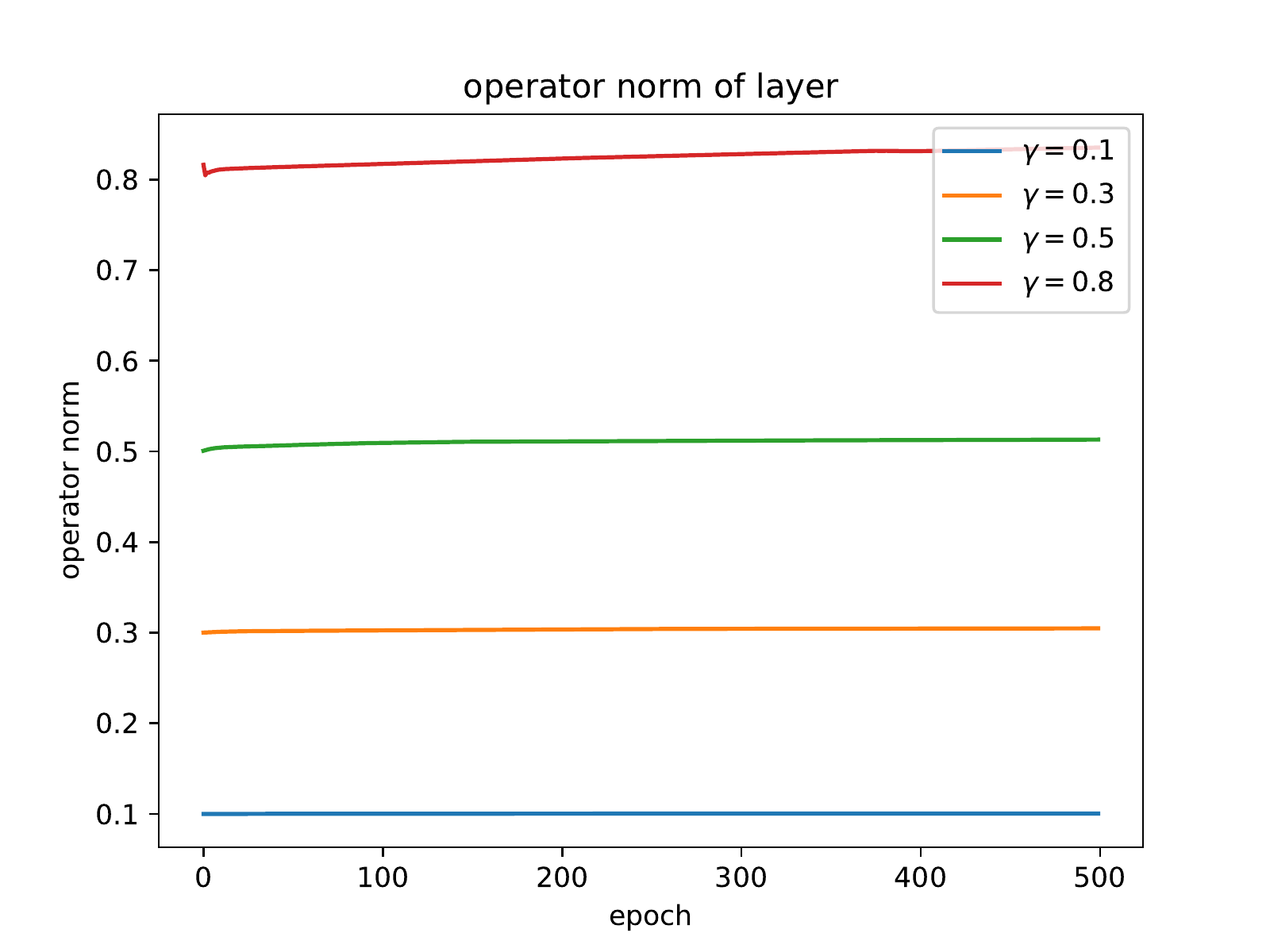}
	\includegraphics[width=.31\textwidth]{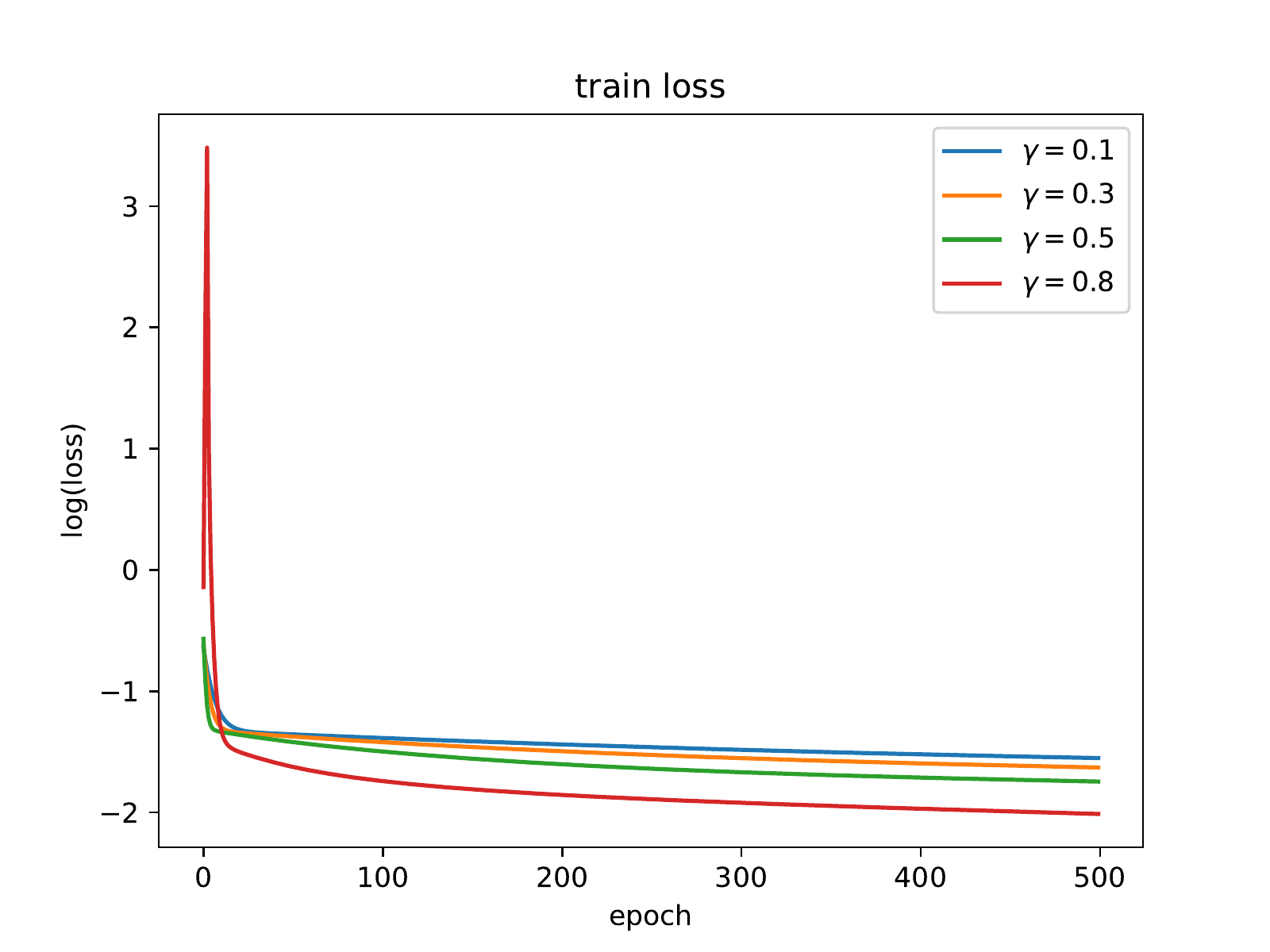}
	\includegraphics[width=.31\textwidth]{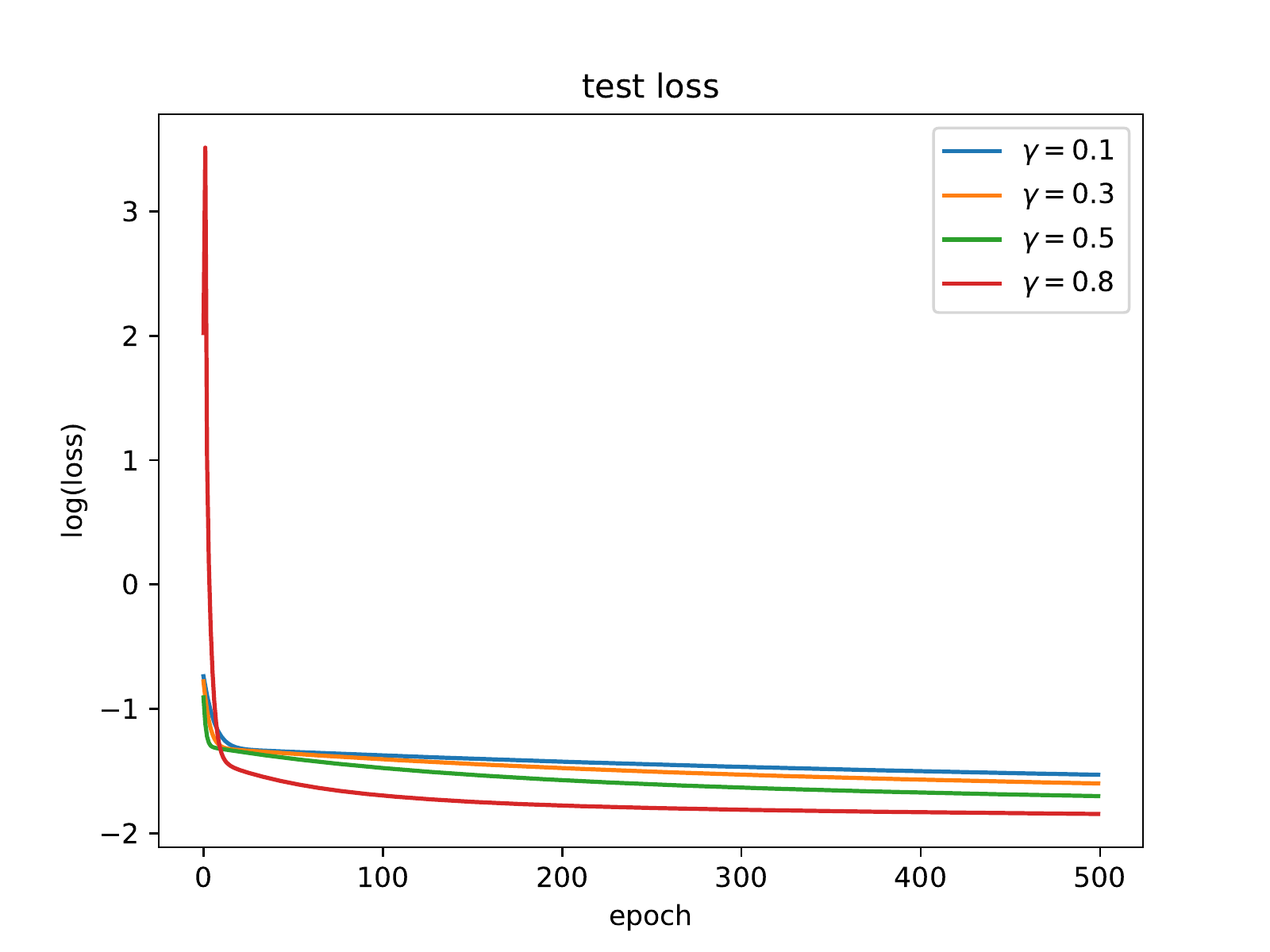}
	\includegraphics[width=.31\textwidth]{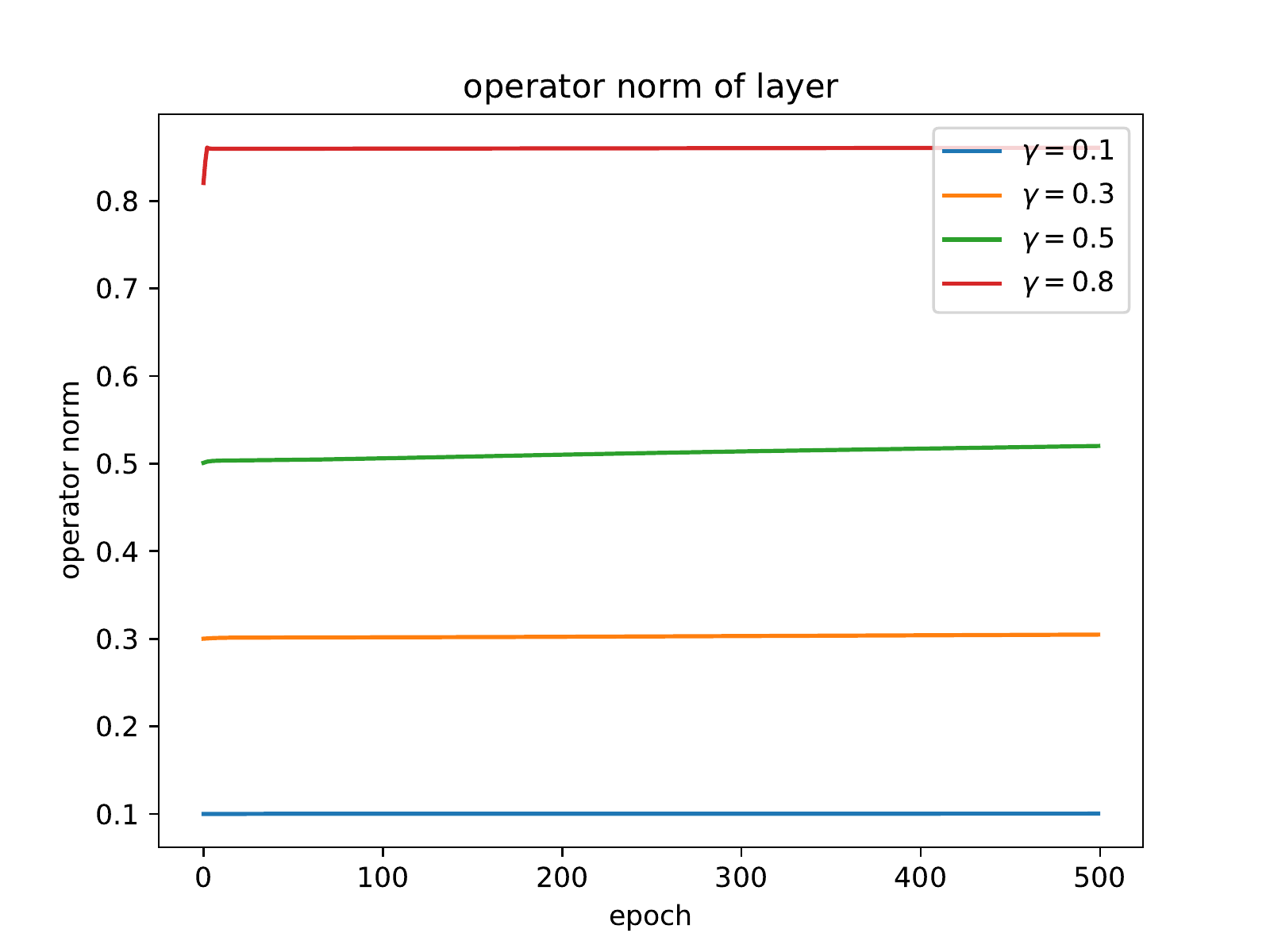}
	\includegraphics[width=.31\textwidth]{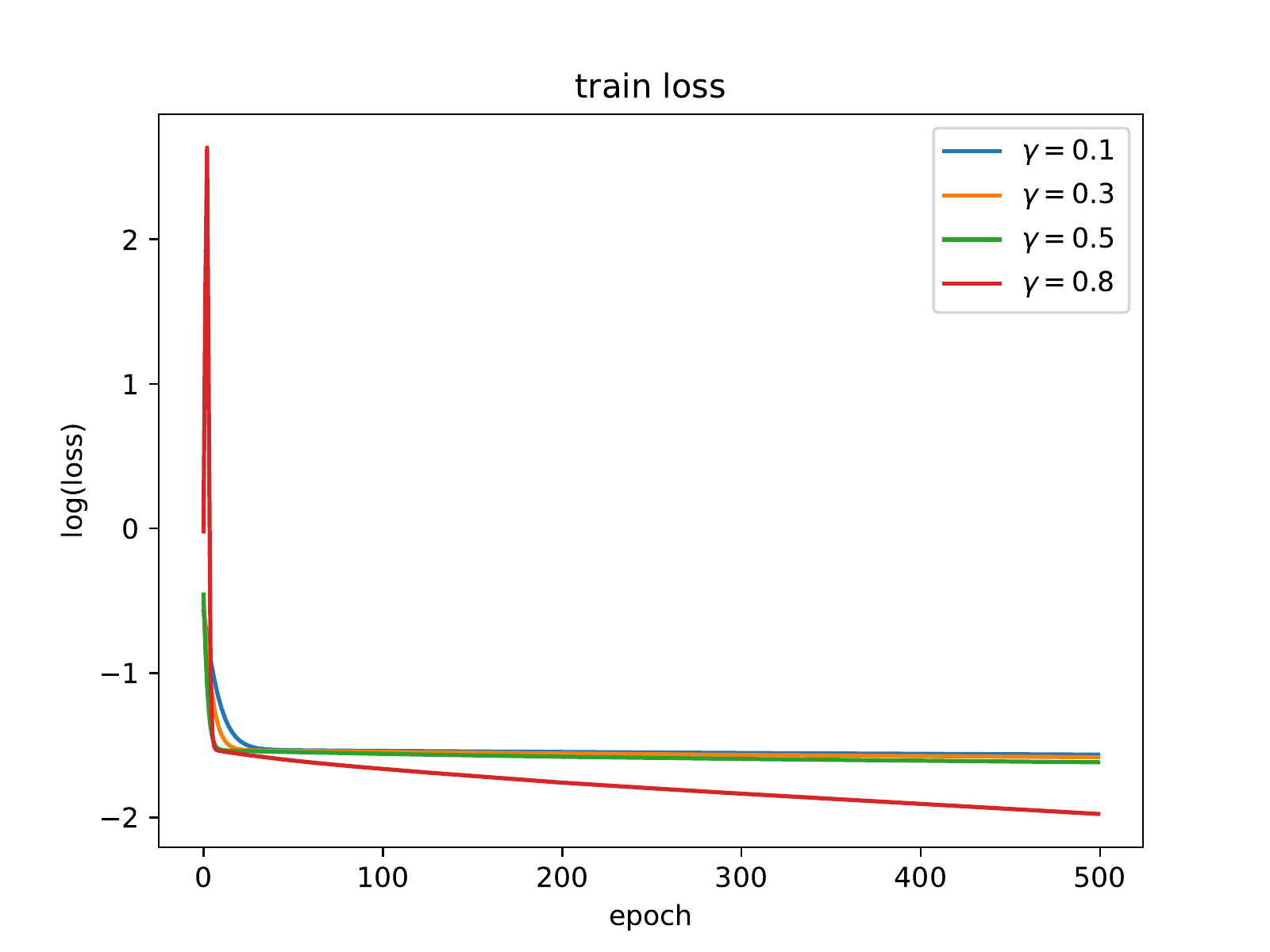}
	\includegraphics[width=.31\textwidth]{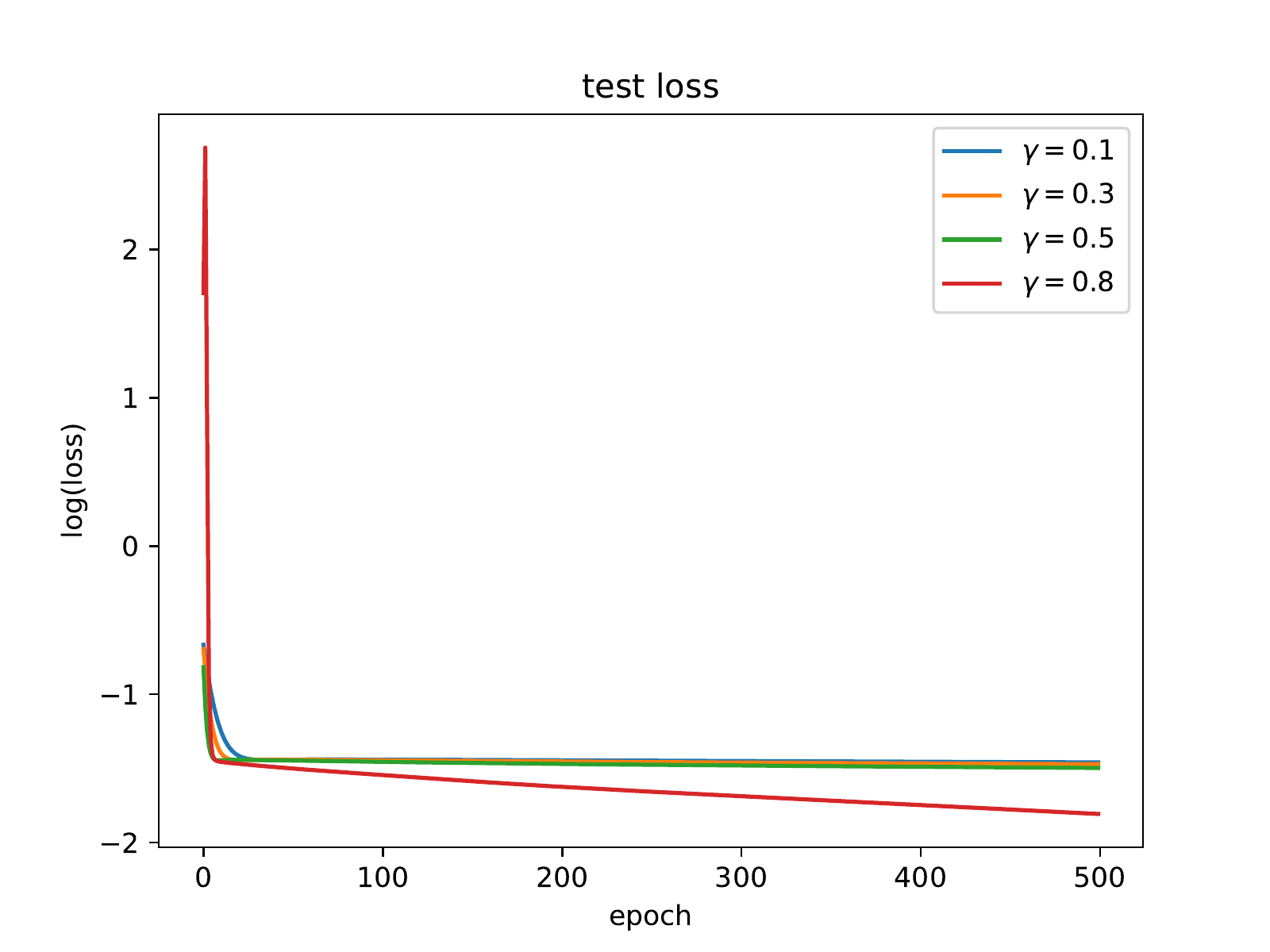}
	\includegraphics[width=.31\textwidth]{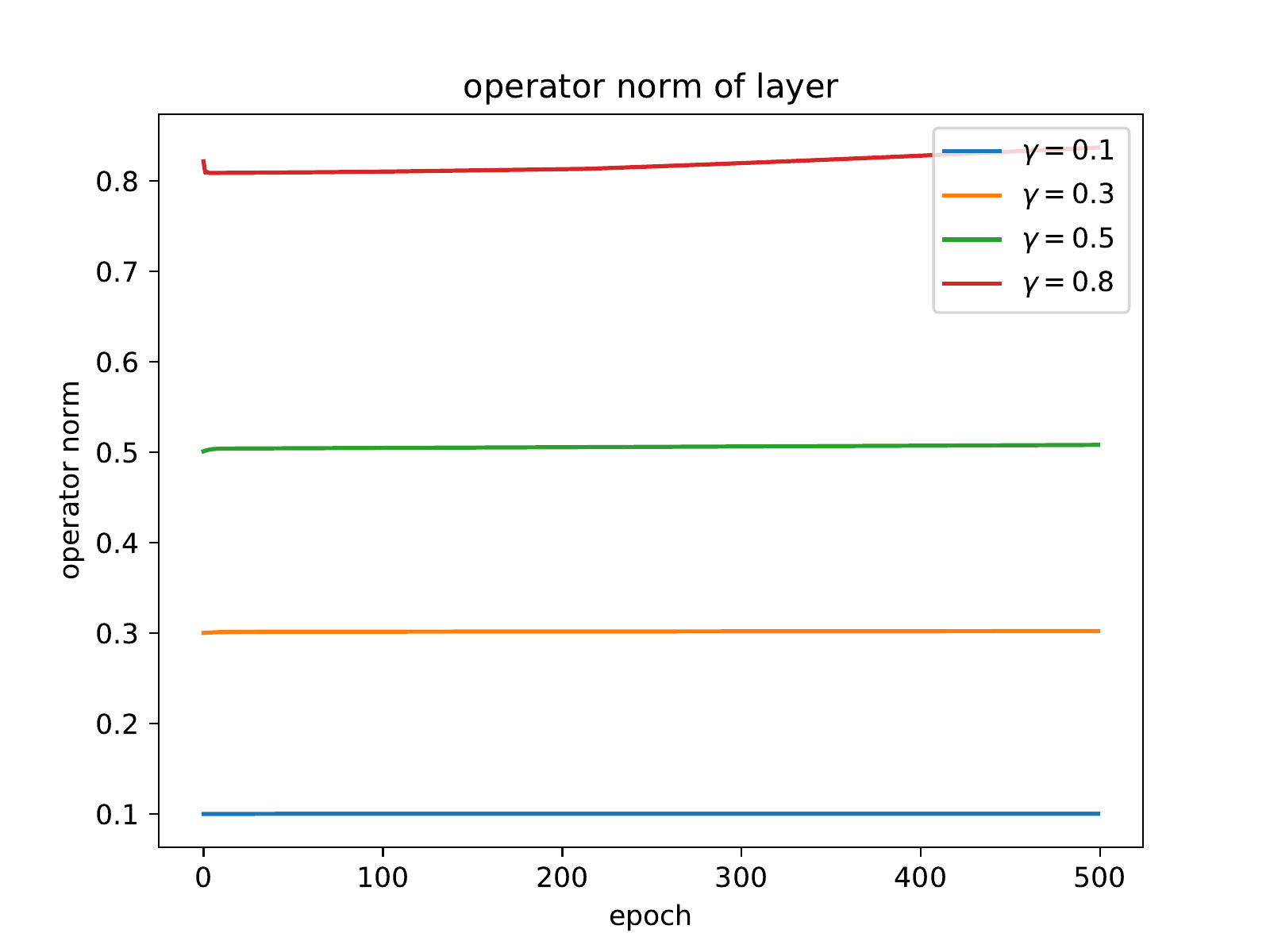}
	\caption{We evaluate the impact of \textbf{choice $\gamma$} on the
		training loss, test loss, and operator norm of the scaled
		matrix $\gamma A(k)$ on the modified dataset of MNIST, FashionMNIST, CIFAR10, and SVHN}
\end{figure}

\clearpage

\end{document}